\documentclass{article}

\usepackage[main, final]{neurips_2026} 

\usepackage[utf8]{inputenc}     %
\usepackage[T1]{fontenc}        %
\usepackage{hyperref}           %
\usepackage{url}                %
\usepackage{booktabs}           %
\usepackage{amsfonts}           %
\usepackage{amsmath}
\usepackage{amssymb}
\usepackage{nicefrac}
\usepackage{microtype}
\usepackage{xcolor}

\usepackage{graphicx}
\usepackage[inline]{enumitem}
\usepackage{placeins}
\usepackage{multirow}
\usepackage{tabularx}
\usepackage{longtable}
\usepackage{algorithm} 
\usepackage{algpseudocode} 

\usepackage[]{color-edits}
\addauthor{ITC}{magenta}
\addauthor{MH}{orange}

\newcommand{\may}[1]{}

\graphicspath{{figures/}}

\title{ZEBRA: Zero-shot Budgeted Resource Allocation\\ for LLM Orchestration}

\author{
  May Hamri$^1$ \quad Inbal Talgam-Cohen$^1$ \\
  \small
  $^1$Tel Aviv University \\
  \texttt{\{mayhamri8,inbaltalgam\}@gmail.com}
}

\usepackage{xcolor}
\newif\ifshowchanges
\showchangesfalse
\ifshowchanges
  \newcommand{\chg}[1]{{\color{red}#1}}
  \newcommand{\chgon}{\color{red}}
\else
  \newcommand{\chg}[1]{#1}
  \newcommand{\chgon}{}
\fi

\begin{document}

\maketitle

\begin{abstract}
As autonomous agents increasingly execute end-to-end tasks under fixed
monetary budgets, the pressing open 
question shifts from whether the budget
is respected, to how to spend it effectively. Existing budget-aware methods typically %
control reasoning step-by-step within a single agent, or learn
resource allocation policies via RL. %
None address
how to split a %
budget across the composing phases of a
multi-agent pipeline at inference time. We propose \textbf{ZEBRA}, a
zero-shot framework that reduces multi-phase budget allocation to a
continuous nonlinear knapsack problem: an LLM controller estimates per-phase
utility curves, and a water-filling search on the Lagrange multiplier
returns the per-phase split. Additive and multiplicative  %
aggregations are unified under the same solver. 
On a $150$-task APPS coding
 benchmark, both ZEBRA variants outperform LLM-direct (budget allocation directly by an LLM) on
every aggregate metric. %
At a budget of
$\alpha = 0.5$ of the unconstrained spend, ZEBRA recovers $94.4\%$ of unconstrained quality,
versus $88.1\%$ for LLM-direct. %
The
advantage is statistically significant and transfers beyond coding: on a $3$-phase HotpotQA pipeline,
ZEBRA beats LLM-direct by \chg{$9.5$}pp, with allocations empirically robust to curve-estimation noise. %
On HotpotQA, ZEBRA arrives at a different budget split (near-balanced) compared to the APPS one (skewed towards a refinement phase), showing adaptation to the
pipeline structure.
More broadly, we show that lightweight
algorithmic guidance at inference time can
improve the economic behavior of autonomous multi-agent systems.
\end{abstract}

\section{Introduction}
\label{sec:introduction}
\begin{figure}[t]
  \centering
  \includegraphics[width=0.85\columnwidth]{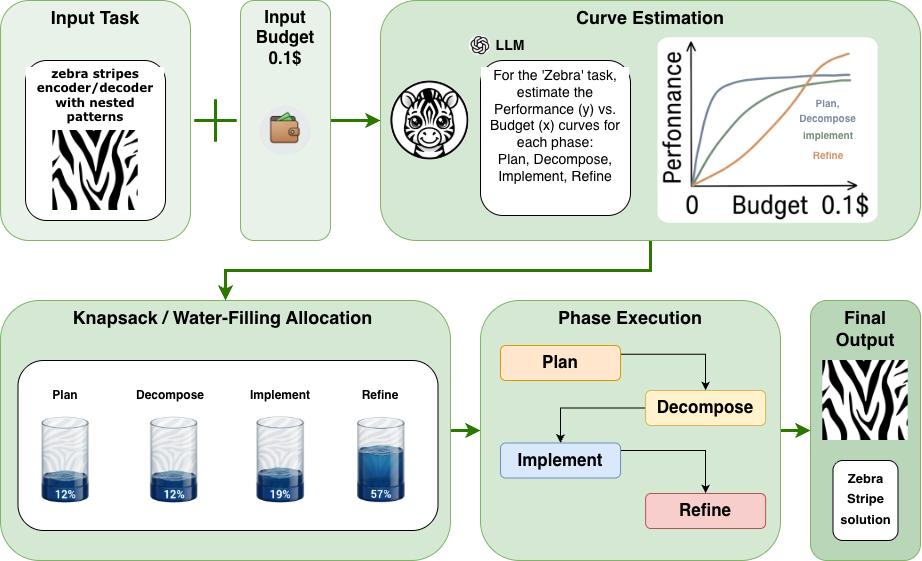}
  \caption{\textbf{ZEBRA overview.} Given an input task and a fixed
  total budget (e.g., \$0.10), ZEBRA adds an allocation agent that
  prompts an LLM to estimate a performance--budget curve for each
  workflow phase (Plan, Decompose, Implement, Refine). ZEBRA solves a
  continuous nonlinear knapsack via water-filling to allocate the
  global budget across phases before execution. The shared
  multi-phase pipeline then runs under these per-phase budget caps.}
  \label{fig:zebra_overview}
\end{figure}

As agents increasingly execute end-to-end tasks, resource
allocation becomes a first-class 
requirement for autonomy. 
Setting limits on resource consumption is crucial -- %
as recently documented, uncontrolled waste of resources emerges in autonomous agentic behavior~\citep{shapira2026agentsofchaos}. 
But users want agents not
only to stay within limits on money, tokens, latency, or compute;
they also need these resources to be used effectively. 
E.g., for a given task, the optimal planning depth, tool usage, and
verification effort differ considerably under a \$10 budget
compared with a \$100 budget. 

The need for effective budget usage is growing in urgency as agent systems become more expensive
and complex. Orchestrators like LangGraph or Gas Town \citep{gastown_github}, which coordinate multiple coding agents, illustrate why economic decision-making is becoming a
core system capability: the orchestrator must allocate resources judiciously across workers, phases, and tools. 
Platforms
such as Replit or Base44 (which autonomously create apps from prompts) enforce %
usage limits and billing
controls in a brute-force manner~\citep[e.g.][]{base44_docs}. They do not currently provide a controller that optimizes the expected
outcome quality under a fixed budget.
Some recent
works have begun to frame budgeting as an optimization problem, but many of these
encode budget reasoning through training-time dynamics (RL
controllers, knapsack exploration) rather than inference-time
allocation. %

We take a complementary route. \textbf{ZEBRA} (\textbf{ZE}ro-shot \textbf{B}udgeted \textbf{R}esource \textbf{A}llocation) adds an allocation agent to a multi-agent pipeline that, instead of directly outputting per-phase budgets, estimates phase utility curves (concretely,
saturating-exponential functions). 
This results in a continuous nonlinear knapsack problem, which the system solves via bisection on the dual Lagrange multiplier (water-filling).
Taken together, we get a zero-shot, inference-time allocation mechanism with no fine-tuning or RL, allowing budgeted execution to approach unconstrained quality at much lower spend.
We study two quality aggregation forms over the phases -- an additive function and a multiplicative one (with offset curves that still enable phase starvation). Despite their fundamental difference in how overall quality depends on per-phase quality, we unify the multiplicative objective with the additive one under the same solver %
via a logarithmic transformation.

We summarize our primary contributions: \emph{(i) Method.} We develop a zero-shot, inference-time
controller, which allocates a fixed budget across pipeline phases, the quality of which aggregates additively or multiplicatively. This is done by
prompting an LLM to estimate phase utility curves, and solving a
continuous nonlinear knapsack problem -- no fine-tuning or RL required. %
\emph{(ii) Experiments.} On a %
$150$-task APPS coding interview
benchmark~\citep{hendrycks2021apps}, and across budget regimes $\alpha \in \{0.3, 0.5, 0.8\}$, both
ZEBRA objectives outperform direct LLM allocation on every aggregate
metric (paired Wilcoxon, BH-adjusted $p < 0.01$), with gains that
scale with budget pressure and concentrate on medium and hard tasks (Section~\ref{sec:results_main}).
The same effect reproduces on two additional coding benchmarks:
HumanEval~\citep{chen2021codex} and
CodeContests~\citep{li2022alphacode} at $\alpha = 0.5$
(Section~\ref{sec:results_transfer}). Importantly, it also generalizes 
\emph{beyond coding}: on the $3$-phase 
question-answering %
HotpotQA pipeline~\citep{yang2018hotpotqa}, ZEBRA beats LLM-direct
by \chg{$+9.5$}pp NB retention (\chg{$p = 0.001$}) (Section~\ref{sec:results_hotpotqa}). ZEBRA also recovers a
near-balanced allocation that matches the (here near-optimal)
uniform baseline -- a distinct allocation shape from the 
strongly-skewed split it learns on APPS. %
This is evidence that ZEBRA
adapts to, rather than memorizes, the task type. %
\emph{(iii) Implication.} More broadly, we show that lightweight
algorithmic guidance -- applied at inference time -- can meaningfully
improve the economic behavior of autonomous multi-agent systems.

\section{Related Work}
\label{sec:related_work}

\noindent
ZEBRA builds on two distinct bodies of prior work. First, \emph{cost-aware
LLM research}, including: (i) the AI provider's cost-saving problem (an inference
operator redistributing a shared compute pool across independent user
queries), and (ii) multi-agent cost management (selecting, routing, and
coordinating agents under a budget). Second, classic \emph{resource
allocation theory}, which provides a principled framework for
distributing a fixed budget across subprocesses with concave utility
curves. ZEBRA combines these by applying the theoretical foundation
to a problem that neither line has addressed: allocating a monetary
budget by an orchestrator across the phases of a multi-agent pipeline, the outputs of which
jointly compose into a single task outcome, %
without
training data or model modification. Table~\ref{tab:rw_comparison} in Appendix~\ref{app:related_work} gives a comparison of related methods along four axes: provider vs.~orchestrator budget, discrete vs.~continuous spending decisions, zero-shot or not, and dependency of items sharing the budget. Here we overview the main approaches. %

{\bf Inference-time token budget allocation.}
Token-budget methods for independent user queries
\citep{han2024tokenbudgetaware,brown2025predictivescheduling,zhao2026roireasoning,hbpo2025}
take the \emph{provider} perspective, operate on \emph{independent}
queries, and pick \emph{discrete} per-query token budgets; most
require task-specific training data or fine-tuning. ZEBRA differs on
every axis (Table~\ref{tab:rw_comparison}): it is an
\emph{orchestration}-side controller that splits a \emph{continuous}
monetary budget across the \emph{dependent} phases of a single
multi-agent pipeline, zero-shot.

{\bf Multi-agent cost management and routing.}
A parallel line of work reduces cost through discrete model selection or agent
routing rather than continuous budget allocation: LLM
cascades~\citep{chen2024frugalgpt}, RL-trained per-query model
selectors~\citep{jin2025corl}, ILP-based agent assembly with RL-fixed
topology~\citep{yang2025bamas}, VAE-driven workflow
synthesis~\citep{su2026daao}, enterprise pipeline
design~\citep{kandogan2025orchestrating}, and upfront vs.\ reactive
selection~\citep{amayuelas2025selfresource}. In every case the decision
variable is \emph{discrete and combinatorial} and the controller is
trained.

{\bf Budget-aware agent control.}
A concurrent line studies budget-aware control at inference time
\emph{inside} a single agent: BATS~\citep{bats2025} pairs a Budget
Tracker plug-in with a per-step ``dig deeper vs.\ pivot'' binary decision;
INTENT~\citep{liu2026intent} trains an intention-aware hierarchical
world model for online tool-call risk calibration; and
BAVT~\citep{li2026bavt} scales multi-hop reasoning value-tree nodes by
the remaining-resource ratio. All three are \emph{step-by-step}
controllers inside a single running agent. ZEBRA targets the
complementary, higher-level problem -- it determines how much of the total budget is allocated to each agent (rather than how the agent spends its share of the budget).  %

{\bf Resource allocation theory and knapsack methods.}
Knapsack and its variants provide the theoretical backbone for budget
allocation~\citep{katohIbarakiHandbook,kellerer2004knapsack,boydVandenberghe}.
\citet{boutilier2016budgetmdps} use multi-choice knapsack for allocating budget among %
independent MDPs (predating LLMs);
recent LLM uses of knapsack
formulations~\citep{zhao2026roireasoning,brown2025predictivescheduling,li2025knapsackrl}
all operate on \emph{independent} queries or training batches rather
than the composing phases of a single running pipeline. %
In the context of question answering, 
SEER~\citep{tonglet2023seer} packages an LLM system with a discrete knapsack solver for selection of exemplars (pairs of training instances and answers, provided as part of the prompt to assist the LLM in answering questions correctly). The purpose of knapsack here is to control the total prompt size, and the aim is to choose exemplars that are simultaneously diverse and similar to the prompt. In comparison, ZEBRA applies a continuous knapsack solver, and uses it for budget allocation across a pipeline with the aim of maximizing aggregate quality, after informing the solver with learned performance curves. 
Delegating the allocation step to
an external knapsack solver (rather than an LLM) is further motivated by
\citet{duchnowski2025knapsack}, who document systematic LLM failure on knapsack problems.

\section{Problem Formulation}
\label{sec:background}

Autonomous agent systems increasingly execute end-to-end tasks under explicit
resource constraints (e.g., monetary cost, API credits, token usage, latency,
GPU time, or memory). A central challenge is planning actions so that the
achieved task quality is maximized subject to a fixed budget, motivating an
explicit optimization formulation rather than a behavior learned implicitly
by fine-tuning or RL.

\subsection{Continuous resource allocation as separable utility maximization}
\label{sec:cnk}

We model an agent or multi-agent pipeline as executing $n$ phases,
indexed by $i \in \{1,\dots,n\}$. Let $B>0$ denote the total available budget
and $x_i \ge 0$ the budget allocated to phase $i$. Each phase has an
associated performance--budget curve $f_i : \mathbb{R}_{\ge 0} \rightarrow
\mathbb{R}$, where $f_i(x_i)$ quantifies the expected contribution to overall
task outcome from spending $x_i$ on phase $i$. The resulting optimization
problem:
\begin{equation}
  \max_{\mathbf{x}} \;\sum_{i=1}^n f_i(x_i) \quad \text{s.t.} \quad
  \sum_{i=1}^n x_i \le B, \;\; x_i \ge 0 \;\forall i.
  \label{eq:primal}
\end{equation}
Problem~\eqref{eq:primal} is a classic \emph{single-resource allocation}
problem with \emph{separable utilities}, equivalent to the continuous
nonlinear knapsack family with ``water-filling''-type allocations under
appropriate structure assumptions
\citep{katohIbarakiHandbook,katohIbarakiMIT,boydVandenberghe}. We assume each
$f_i$ is non-decreasing, concave, and differentiable, so~\eqref{eq:primal}
is a convex program with optimality fully characterized by KKT
conditions~\citep{boydVandenberghe} (full assumptions, optional per-phase
upper bounds $u_i$, and the non-convex case are deferred to
Appendix~\ref{app:opt_details}).

\subsection{KKT conditions and dual-search algorithm}
\label{sec:dual}

Introducing a Lagrange multiplier $\lambda \ge 0$ for the budget constraint
yields the well-known \emph{marginal equalization} rule
$f_i'(x_i^\star) = \lambda$ at every active phase, with the natural
boundary modifications when $x_i^\star$ hits $0$ or $u_i$. Economically,
$\lambda$ is the shadow price of budget and matches the water-filling
solution from communication systems~\citep{boydVandenberghe}. The total
allocation $S(\lambda) = \sum_i x_i(\lambda)$ is non-increasing in
$\lambda$, so $\lambda^\star$ can be found by bisection in
$O(n \log\tfrac{1}{\varepsilon})$ time, matching the classical
resource-allocation result~\citep{katohIbarakiHandbook,katohIbarakiMIT} (full derivation in Appendix~\ref{app:opt_details}). ZEBRA also handles
multiplicative objectives via a $\log$ transformation that reduces a
weighted product $\prod_i h_i(x_i)^{w_i}$ to $\sum_i w_i \log h_i(x_i)$,
solvable by the same dual-search (Appendix~\ref{app:log_transform}).

\section{Method}
\label{sec:method}

\begin{figure}[t]
  \centering
  \includegraphics[width=0.95\columnwidth]{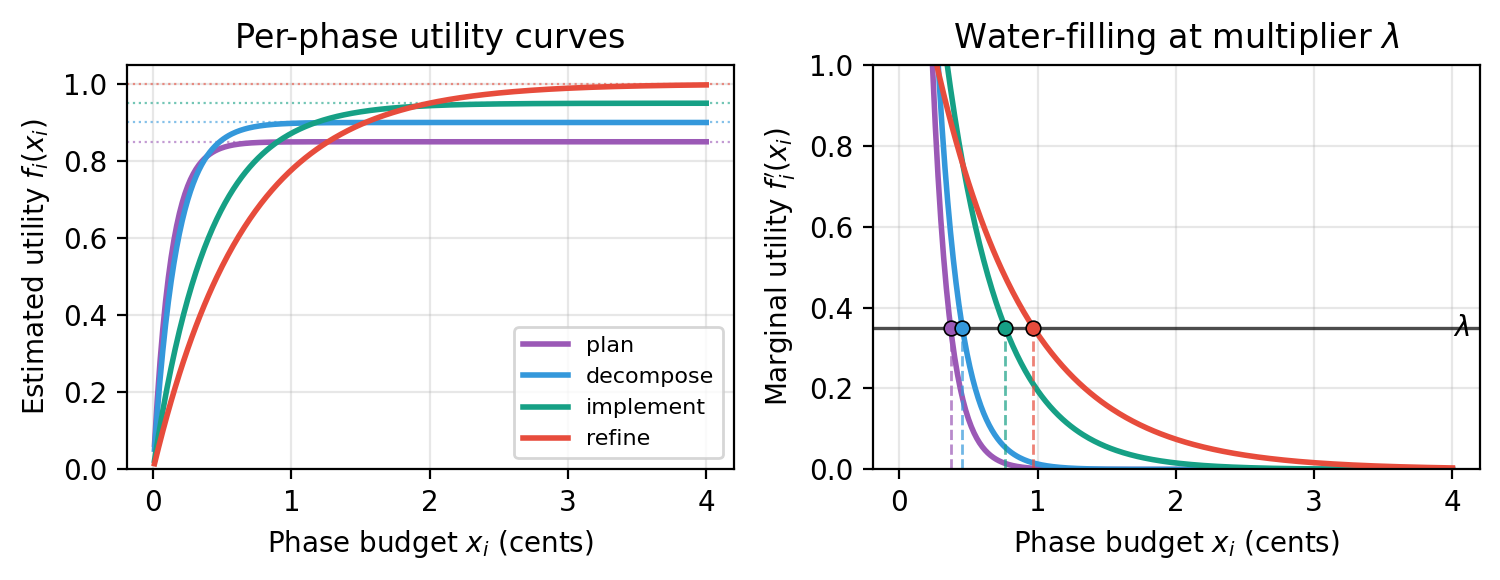}
  \caption{\textbf{Per-phase utility curves and water-filling.}
  \emph{Left:} the saturating exponential
  $f_i(x) = a_i(1 - e^{-b_i x})$ for the four pipeline phases, with the
  quality ceiling $a_i$ shown as a dotted line. Phases differ in how
  quickly they saturate (the $b_i$ parameter): \texttt{plan} reaches its ceiling
  fastest, \texttt{refine} slowest. \emph{Right:} the corresponding
  marginal-utility curves $f_i'(x) = a_i b_i e^{-b_i x}$. ZEBRA's
  knapsack solution is precisely water-filling: pick a single Lagrange
  multiplier $\lambda$ (horizontal line), and read off each phase's budget
  $x_i$  where its marginal-utility curve crosses $\lambda$. The
  total budget constraint $\sum_i x_i \le B$ is enforced by binary
  search on $\lambda$.}
  \label{fig:phase_curves}
\end{figure}

Figure~\ref{fig:zebra_overview} shows how ZEBRA augments a
multi-agent pipeline of $n$ phases with an \emph{allocation agent}
(controller). Before execution, it assigns each phase a budget
$x_i \ge 0$ with $\sum_{i=1}^n x_i \le B$. Our method is
inference-time \emph{algorithmic guidance}: (1) estimate curves; (2)
solve for allocations; (3) execute phases under fixed budgets. Figure~\ref{fig:phase_curves} schematically shows the curves and water-filling solution\chg{, and Algorithm~\ref{alg:zebra} (Appendix~\ref{app:zebra_algorithm}) states all three steps end-to-end in a single self-contained procedure.
}

\subsection{Parametric phase utility curves}
\label{sec:curves}
Let $f_i(x)$ denote the expected utility of phase $i$ when spending budget $x$. To preserve diminishing returns and enable fast optimization, we fit each phase with a concave, saturating exponential:
\begin{equation}
f_i(x) = a_i \left(1 - e^{-b_i x}\right),
\label{eq:sat_exp}
\end{equation}
where $a_i \in (0,1]$ is the quality ceiling and $b_i > 0$ controls how quickly the phase saturates. This form is monotone and concave for $a_i,b_i > 0$ (derivatives in Appendix~\ref{app:method_details}).

{\bf Curve-parameter estimation via two-point fitting.}
Rather than asking the controller for abstract curve parameters, we elicit two intuitive operating points per phase: $n_{\text{basic},i}$, the total output tokens this phase needs to reach roughly $50\%$ of its potential quality, and $n_{\text{great},i}$, the total output tokens to reach roughly $90\%$ (named \texttt{tokens\_basic} and \texttt{tokens\_great} in the controller prompt; see Appendix~\ref{app:example:prompts}). We fit $b_i$ as the geometric mean of the two implied rates and ask the controller to estimate the ceiling $a_i \in (0,1]$ alongside $b_i$ (full equations and the propagation-weighted variant in Appendices~\ref{app:method_details} and~\ref{app:prop_offset}). Token operating points are converted to USD via per-phase model pricing before $b_i$ is computed, so $b_i$ has units of inverse-USD and matches the monetary budget $B$\chg{. Appendix~\ref{app:curve_validation} collects the validation of these curves in one place: shape calibration against measured phase response, pipeline monotonicity, robustness to independent and correlated estimation error, and shuffle controls establishing that the estimates carry per-task and per-phase signal}.

\subsection{Allocation objectives}
\label{sec:objectives}

Given the estimated curves $\{(a_i, b_i)\}_{i=1}^n$, we consider two ways to aggregate per-phase utilities into a pipeline-level objective: an independent sum and a multiplicative product. Both reduce to a separable concave program with a single budget constraint, so both are solved by the same Lagrangian dual-search of Section~\ref{sec:dual}: bisect on a single shadow price $\lambda$ until the per-phase responses sum to $B$. The two objectives differ only in the closed form of $x_i(\lambda)$.
\subsubsection{Additive objective}
\label{sec:additive}
The additive formulation assumes phase contributions are independent and combine as a sum:
\begin{equation}
  \max_{\mathbf{x}} \;\sum_{i=1}^n a_i\bigl(1 - e^{-b_i x_i}\bigr)
  \quad \text{s.t.} \quad \sum_{i=1}^n x_i \le B, \;\; x_i \ge 0 \;\forall i.
  \label{eq:zebra_additive}
\end{equation}
Setting $f_i'(x_i) = \lambda$ at every active phase and solving gives the closed-form per-phase response
$x_i(\lambda) = \max\!\bigl(0,\, \tfrac{1}{b_i}\ln\tfrac{a_i b_i}{\lambda}\bigr)$,
with $\lambda$ as the shadow price of the budget. Phase $i$ is starved exactly when its \emph{marginal at zero} $f_i'(0) = a_i b_i$ falls below this price -- i.e., the first dollar spent on phase $i$ buys less pipeline quality than the same dollar spent elsewhere.

\subsubsection{Multiplicative offset objective}
\label{sec:mult_offset}
A multiplicative aggregator captures the intuition that the pipeline behaves as a chain: a failure in any phase degrades the whole task. The na\"ive choice $\prod_i f_i(x_i)$ has a problematic corner  $f_i(0) = 0$, so any unfunded phase forces the product to zero, and the optimizer is forced to fund every phase. We restore the ability to starve phases by replacing $f_i$ with the \emph{offset} curve
\begin{equation}
g_i(x) \;=\; (1 - a_i) \;+\; a_i\bigl(1 - e^{-b_i x}\bigr) \;=\; 1 - a_i\, e^{-b_i x},
\label{eq:offset}
\end{equation}
i.e., $f_i(x)$ shifted up by a baseline $1 - a_i$. Concretely, $g_i$ ranges from $g_i(0) = 1 - a_i$ to $g_i(\infty) = 1$, so it can be read as a per-phase \emph{pass-through quality}: a phase that gets no budget still contributes the baseline $1 - a_i$, and approaches a perfect $1$ as more budget is spent. The pipeline-level objective is 
\begin{equation}
  \max_{\mathbf{x}} \;\prod_{i=1}^n g_i(x_i)
  \;=\; \prod_{i=1}^n \bigl(1 - a_i\, e^{-b_i x_i}\bigr)
  \quad \text{s.t.} \quad \sum_i x_i \le B,\;\; x_i \ge 0,
  \label{eq:mult_offset}
\end{equation}
i.e., the product of these pass-through qualities, which lies in $[\prod_i (1-a_i),\, 1]$. Phases with high ceilings (large $a_i$) hurt the product most when starved and therefore attract budget; phases with $a_i$ near zero already have $g_i(0)$ near $1$ and can be skipped at almost no cost. Taking logs (see Appendix~\ref{app:log_transform}) reduces \eqref{eq:mult_offset} to the separable concave program $\max \sum_i \log g_i(x_i)$, solvable by the same dual-search; the closed-form per-phase response and starvation threshold are derived in Appendix~\ref{app:method_details}.

{\bf Sum vs.\ product: different starvation criteria.}
The two objectives differ in \emph{which} phases they starve when the budget binds. The additive objective starves phases with low marginal at zero (small $a_i b_i$), due to ``low return on the first dollar.'' The multiplicative offset objective starves phases with low ceiling (small $a_i$), since ``baseline pass-through is already near $1$, so the product is fine without budget.'' These two policies thus behave differently when the budget binds (Section~\ref{sec:results_main}). A propagation-weighted extension is in Appendix~\ref{app:prop_offset}.

\section{Experiments}
\label{sec:experiments}

\subsection{Setup}
\label{sec:exp_setup}

{\bf Question.}
Given a fixed multi-phase coding workflow and total budget, does
estimating phase utility curves and solving for allocations beat
asking an LLM to split the budget directly? We hold the pipeline,
models, and budget constant and vary only the \texttt{allocate}
component.

{\bf Benchmark.}
We evaluate on the APPS benchmark~\citep{hendrycks2021apps} at the
\emph{interview} difficulty level. We construct a benchmark of
$150$ tasks balanced across difficulty tiers ($50$ each), where
difficulty is defined by the unconstrained pipeline's mean score:
\emph{easy} ($\bar{s} \ge 0.8$), \emph{medium}
($0.4 \le \bar{s} < 0.8$), \emph{hard} ($\bar{s} < 0.4$). Tasks
are screened for cost stability
($\mathrm{CV}_{\text{cost}} < 0.35$ over $30$ unconstrained runs)
so that per-task budget references are meaningful. We additionally
run two transfer benchmarks at $\alpha = 0.5$ --
HumanEval~\citep{chen2021codex} and
CodeContests~\citep{li2022alphacode} -- with the same protocol
(Section~\ref{sec:results_transfer}). Construction details are in
Appendix~\ref{app:additional_results}.

{\bf Budget levels.}
For each task we set the total budget relative to its mean
unconstrained cost $\bar{c}$, with $B = \alpha\,\bar{c}$. We
evaluate every strategy at $\alpha \in \{0.5, 0.8\}$ -- the
\emph{tight} and \emph{moderate} regimes -- and additionally run a
\emph{very-tight} regime $\alpha = 0.3$ on the $50$ easy tasks.

{\bf Strategies.}
We compare the two ZEBRA objectives -- \texttt{additive}
(Sec.~\ref{sec:additive}) and \texttt{mult\_offset}
(Sec.~\ref{sec:mult_offset}) -- against \textbf{LLM-direct}: the
controller, given the same pipeline description and total budget,
outputs a per-phase split. This baseline asks whether an LLM can
self-allocate budget via internal reasoning instead of an explicit
solver -- a non-trivial question given documented LLM weakness on
NP-hard combinatorial problems including
knapsack~\citep{duchnowski2025knapsack}, which had not been
measured in the orchestration setting (where the model can
introspect over its own pipeline structure). All controller-based
strategies share the controller model and prompt scaffold and
differ only in how the allocation is produced. \textbf{no\_budget},
the unconstrained pipeline, is the oracle ceiling.

{\bf Pipeline and models.}
The workflow is a LangGraph
\texttt{StateGraph}~\citep{langgraph_graph_api,langgraph_stategraph_ref}:
\texttt{allocate} $\to$ \texttt{plan} $\to$ \texttt{decompose} $\to$
\texttt{implement} $\to$ \texttt{refine} $\to$ \texttt{END}. The
graph is identical across strategies; only \texttt{allocate}
differs. Base phases use \texttt{gpt-4o-mini}; \texttt{refine} (a
review$\to$revise loop, up to $3$ iterations) and the controller
use \texttt{gpt-4o}. Budget enforcement uses a budget-capped LLM
interface\chg{.
}\emph{\chg{Generation is test-blind}}\chg{: no test case, test outcome, or score is ever
exposed to any phase, and }\texttt{\chg{refine}}\chg{'s review$\to$revise loop is driven
by an LLM reviewer that critiques the candidate against the problem
specification - not by executing tests. Tests are used only to select among candidates a run has already generated, after all generation for that run is complete. At that point we score }the best candidate
\chg{the run produced rather than the last one, which keeps the pipeline monotone
in iterations, as the utility model of Section~\ref{sec:cnk} assumes. This
best-candidate rule is a property of the measurement, not of the pipeline: it
is applied identically to every allocator, including LLM-direct, so it cannot
favour one strategy over another}. See
Appendix~\ref{app:phases} and \ref{app:model_snapshots}.

{\bf Evaluation.}
For each $(\text{task}, \alpha, \text{strategy})$ cell we perform
$15$ independent runs and report mean and median per-task score,
task-averaged success rate ($\text{score} > 0.5$), realized cost,
and \emph{NB retention} (mean score divided by the no-budget mean
on the same tasks). Significance vs.\ LLM-direct uses paired
Wilcoxon signed-rank tests on per-task mean scores. Across the
$18$ ZEBRA-vs-LLM-direct comparisons of the main APPS family
(App.~\ref{app:wilcoxon}, Table~\ref{tab:wilcoxon}) we control FDR
at $q = 0.05$ via Benjamini--Hochberg
\citep{benjamini1995controlling}; significance markers reflect
BH-adjusted $p$-values.

\subsection{Main results}
\label{sec:results_main}

\begin{table*}[t]
\centering
\small
\setlength{\tabcolsep}{5pt}
\renewcommand{\arraystretch}{1.15}
\begin{tabularx}{\textwidth}{@{} l l
  >{\centering\arraybackslash}X
  >{\centering\arraybackslash}X
  >{\centering\arraybackslash}X
  >{\centering\arraybackslash}X
  >{\centering\arraybackslash}X @{}}
\toprule
$\boldsymbol{\alpha}$ & \textbf{Strategy} &
\textbf{Mean score} &
\textbf{Median} &
\textbf{Success rate} &
\textbf{Cost (\$)} &
\textbf{NB retention} \\
\midrule

\multirow{4}{*}{$0.5$}
& zebra-additive        & $0.5258$           & $0.5333$           & $52.7\%$           & $0.0134$ & $92.0\%^{**}$  \\
& zebra-mult\_offset    & $\mathbf{0.5397}$  & $\mathbf{0.5362}$  & $\mathbf{54.1\%}$  & $0.0110$ & $\mathbf{94.4\%}^{***}$ \\
& llm                   & $0.5038$           & $0.5026$           & $49.7\%$           & $0.0109$ & $88.1\%$       \\
& no\_budget (oracle)   & $0.5717$           & ---                & ---                & ---      & $100.0\%$      \\
\midrule

\multirow{4}{*}{$0.8$}
& zebra-additive        & $\mathbf{0.5605}$  & $\mathbf{0.5972}$  & $\mathbf{56.2\%}$  & $0.0192$ & $\mathbf{98.1\%}^{***}$ \\
& zebra-mult\_offset    & $0.5576$           & $0.5957$           & $55.2\%$           & $0.0167$ & $97.5\%^{**}$  \\
& llm                   & $0.5389$           & $0.5323$           & $54.0\%$           & $0.0140$ & $94.3\%$       \\
& no\_budget (oracle)   & $0.5717$           & ---                & ---                & ---      & $100.0\%$      \\

\bottomrule
\end{tabularx}
\caption{\textbf{Main results on $150$ APPS interview-level tasks}
($15$ runs per cell). Mean and median are taken over per-task mean
scores; success rate is the task-averaged fraction of runs with
score $> 0.5$; NB retention is computed as a ratio of means, $\overline{s}/\overline{s}_{\text{NB}}$ (the strategy's mean score on the task set divided by the no-budget mean score on the same set). The two ZEBRA variants differ
in the allocation objective (sum vs.\ product).
Significance markers on NB retention come from a paired Wilcoxon
signed-rank test on per-task mean score against LLM-direct, with
$p$-values BH-adjusted within the $18$-test APPS main family
(Appendix~\ref{app:wilcoxon}, Table~\ref{tab:wilcoxon};
$^{*}p_{\text{BH}}<0.05$, $^{**}p_{\text{BH}}<0.01$,
$^{***}p_{\text{BH}}<0.001$). Bold marks the best
ZEBRA variant per column within each $\alpha$ block. \textbf{Both
ZEBRA variants outperform LLM-direct at both budget levels on every
aggregate metric.}}
\label{tab:main_overall}
\end{table*}

Table~\ref{tab:main_overall} reports the overall comparison on all
$150$ tasks. \emph{Both ZEBRA objectives outperform LLM-direct at
both budget levels on every aggregate metric we track.} At the tight
budget ($\alpha = 0.5$), \texttt{mult\_offset}
recovers $94.4\%$ of the unconstrained score, versus $88.1\%$ for
LLM-direct -- a $6.3$-point retention gap at essentially identical
spend ($\$0.011$ per task). At the moderate budget ($\alpha = 0.8$),
\texttt{additive} recovers $98.1\%$
versus $94.3\%$ for LLM-direct. Both ZEBRA variants are significant
against LLM-direct at $p < 0.01$ (paired Wilcoxon, $n = 150$); the
strongest variant at each budget level achieves $p < 0.001$.

\subsection{The advantage concentrates on medium and hard tasks}
\label{sec:results_by_tier}

\begin{table*}[t]
\centering
\small
\setlength{\tabcolsep}{4pt}
\renewcommand{\arraystretch}{1.1}
\begin{tabularx}{\textwidth}{@{} l l
  >{\centering\arraybackslash}X
  >{\centering\arraybackslash}X
  >{\centering\arraybackslash}X
  >{\centering\arraybackslash}X
  >{\centering\arraybackslash}X
  >{\centering\arraybackslash}X @{}}
\toprule
\textbf{Tier} & \textbf{Strategy} &
\multicolumn{3}{c}{$\boldsymbol{\alpha = 0.5}$} &
\multicolumn{3}{c}{$\boldsymbol{\alpha = 0.8}$} \\
\cmidrule(lr){3-5} \cmidrule(lr){6-8}
 & & \textbf{Mean} & \textbf{Succ\%} & \textbf{$p$ vs LLM}
   & \textbf{Mean} & \textbf{Succ\%} & \textbf{$p$ vs LLM} \\
\midrule

\multirow{3}{*}{\textbf{Easy} ($n=50$)}
& zebra-additive     & $0.9216$           & $94.3\%$           & $0.19$~\textsuperscript{ns}  & $0.9556$          & $\mathbf{98.1\%}$ & $0.61$~\textsuperscript{ns}  \\
& zebra-mult\_offset & $\mathbf{0.9385}$  & $\mathbf{95.9\%}$  & $0.18$~\textsuperscript{ns}  & $\mathbf{0.9582}$ & $97.9\%$          & $0.27$~\textsuperscript{ns}  \\
& llm                & $0.9325$           & $94.8\%$           & ---                          & $0.9505$          & $97.5\%$          & ---                          \\
\midrule

\multirow{3}{*}{\textbf{Medium} ($n=50$)}
& zebra-additive     & $0.5324$           & $59.6\%$           & $0.0025$\textsuperscript{**}     & $\mathbf{0.5847}$ & $\mathbf{65.5\%}$ & $0.0043$\textsuperscript{**}     \\
& zebra-mult\_offset & $\mathbf{0.5506}$  & $\mathbf{62.7\%}$  & $0.0000$\textsuperscript{***}    & $0.5772$          & $64.3\%$          & $0.022$\textsuperscript{*}       \\
& llm                & $0.4833$           & $52.7\%$           & ---                              & $0.5428$          & $61.1\%$          & ---                              \\
\midrule

\multirow{3}{*}{\textbf{Hard} ($n=50$)}
& zebra-additive     & $0.1234$           & $\mathbf{4.3\%}$   & $0.0042$\textsuperscript{**}     & $\mathbf{0.1412}$ & $\mathbf{5.1\%}$  & $0.0041$\textsuperscript{**}     \\
& zebra-mult\_offset & $\mathbf{0.1300}$  & $3.9\%$            & $0.0044$\textsuperscript{**}     & $0.1376$          & $3.6\%$           & $0.019$\textsuperscript{*}       \\
& llm                & $0.0956$           & $1.7\%$            & ---                              & $0.1233$          & $3.6\%$           & ---                              \\

\bottomrule
\end{tabularx}
\caption{\textbf{Results by difficulty tier.} Mean score is the
per-task mean score averaged over the tier; success rate is the
fraction of runs with score $> 0.5$, task-averaged. The
\textbf{$p$ vs LLM} column reports the two-sided paired Wilcoxon
signed-rank $p$-value on per-task mean score against LLM-direct
($n = 50$ per tier; $^{*}p<0.05$, $^{**}p<0.01$, $^{***}p<0.001$,
ns~=~not significant). Differences are non-significant on easy tasks
(both methods saturate within $\pm 0.6$ points of each other) and
become significant for both ZEBRA variants on the medium and hard
tiers. The largest jumps appear on \emph{medium} tasks at $\alpha =
0.5$, where \texttt{mult\_offset} lifts mean score by $+6.7$ points
($+10.0$~pp success) over LLM-direct. Bold marks the best strategy per
column within each tier $\times$ $\alpha$ cell.}
\label{tab:main_by_tier}
\end{table*}

Table~\ref{tab:main_by_tier} breaks the same results down by
difficulty tier. \textbf{Easy tasks tie} (all strategies recover
$\approx 93$--$96\%$ of unconstrained quality; no ZEBRA-vs-LLM
difference is significant); \textbf{the gap opens on medium and
hard tasks}: on medium tasks at $\alpha = 0.5$ the best ZEBRA
variant lifts mean score by $+6.7$ points ($+10.0$~pp success rate)
over LLM-direct ($p_{\text{BH}} = 3.7\!\times\!10^{-4}$); on hard
tasks ZEBRA wins by $+3.4$ points ($p_{\text{BH}} < 0.01$). The
same pattern holds at $\alpha = 0.8$, with significance preserved
on every medium/hard cell. Allocation quality matters only when
the budget binds; a per-task scatter view
(Fig.~\ref{fig:per_task_scatter}, App.~\ref{app:per_task_scatter})
confirms easy tasks cluster on the diagonal while harder tasks
drift above it.

\subsection{Budget pressure: ZEBRA's gap scales with tightness}
\label{sec:results_pressure}

To test whether the easy-tier tie is about the tasks being easy or
the budget being generous, we rerun the $50$ easy tasks at a
very-tight budget $\alpha = 0.3$. The picture flips:
\texttt{zebra-additive} now beats LLM-direct on per-task mean
($p_{\text{BH}} = 0.030$; full numbers in
App.~\ref{app:budget_pressure}). Combined with the tier breakdown
this gives a simple rule of thumb: \emph{ZEBRA's advantage over LLM
allocation scales with budget tightness} -- whether tightness comes
from a harder task or a smaller $\alpha$
(Fig.~\ref{fig:pressure_retention}, App.~\ref{app:budget_pressure}).

\subsection{Why ZEBRA wins: it adapts the spend}
\label{sec:results_allocations}

ZEBRA spends \emph{differently}:
Figure~\ref{fig:alloc_bars} and
Table~\ref{tab:allocation_distribution}
(Appendix~\ref{app:allocation_distribution}) show the average fraction
of the budget spent on each phase by tier.
\textbf{ZEBRA adapts to task difficulty; LLM-direct does not.} On
easy tasks at $\alpha = 0.5$, ZEBRA assigns \texttt{refine} only
$30$--$45\%$ of the budget; on medium and hard tasks, both variants
pull strongly toward \texttt{refine} ($60$--$66\%$). LLM-direct, by
contrast, spends an essentially fixed $\sim$$25\%$ on \texttt{implement}
and $\sim$$50\%$ on \texttt{refine} regardless of tier or budget. At
$\alpha = 0.3$ on easy tasks, ZEBRA variants compress \texttt{refine}
still further ($10.7\%$ for \texttt{additive}, $19.3\%$ for
\texttt{mult\_offset}), while LLM-direct keeps it near $47\%$
(Table~\ref{tab:alloc_alpha03}, Appendix~\ref{app:budget_pressure}). This adaptation
is the likely cause of the score gaps in
Table~\ref{tab:main_by_tier} and the retention scaling in
Figure~\ref{fig:pressure_retention}.

\paragraph{Additive vs.\ mult\_offset.}
\texttt{additive} is the per-task win-count leader at every
aggregate budget ($64/150$ at $\alpha = 0.5$;
Table~\ref{tab:variant_wins}). \texttt{mult\_offset} wins where its
product form matters: on hard tasks at $\alpha = 0.5$ it reaches
$0.130$ mean (vs.\ $0.123$ \texttt{additive}, $0.096$ LLM-direct)
and takes $21/50$ wins. \texttt{mult\_offset}'s product form drags the objective down whenever any single phase is under-funded, so the optimizer concentrates budget on the phase that would otherwise be the weakest link. This bias only pays off when the budget is too tight to fund every phase well \emph{and} the task can't tolerate a weak phase, which is precisely the hard-tier, $\alpha = 0.5$
cell.

\subsection{Ablations}
\label{sec:results_ablations}

We run \chg{five }ablations against \texttt{zebra-mult\_offset} on the
same $150$ APPS tasks (full numbers in App.~\ref{app:ablations}).
\textbf{(1) ZEBRA-LLM} -- swap the knapsack solver for the
controller given the \emph{same} fitted curves: $-4.20$ to $-4.31$
points across budgets. \textbf{(2) Uniform} ($25/25/25/25$) -- drop
both curves and solver: $-5.14$ to $-5.92$ points (both
$p<10^{-4}$, $n=150$). \textbf{(3) LLM-CoT} rules out a prompting
confound: a CoT scaffold on LLM-direct at $\alpha = 0.5$ scores
$0.0070$ worse than plain LLM-direct ($p_{\text{raw}} = 0.054$,
n.s.), and ZEBRA still beats CoT by $+0.0429$ ($p_{\text{raw}}
< 10^{-4}$). \textbf{(4) Fixed-Avg} bakes in ZEBRA's global mean
per-phase share ($11.3/14.0/24.1/50.6\%$) and applies it to every
task, isolating per-task adaptation: it matches dynamic ZEBRA on
easy tasks but loses $+9$ to $+22$ pp NB-retention on medium/hard,
with the gap statistically significant in $3$ of the $4$
medium/hard cells ($p < 0.01$); the exception is \texttt{additive}
vs.\ Fixed-Avg on hard, $p = 0.17$\chg{. The same split holds at a much tighter
budget: at $\alpha = 0.3$ ($100$ tasks) the two tie on easy
($0.9217$ vs.\ $0.9242$) while }\texttt{\chg{additive}} \chg{leads on hard
($0.1020$ vs.\ $0.1115$, $+9.3\%$ relative)
}(App.~\ref{app:fixed_avg}).
\textbf{\chg{(5) Cost-Prop}} \chg{drops the allocator entirely and splits the budget by
each phase's share of unconstrained spend ($9.0/5.4/14.8/70.8\%$, measured
over no-budget runs). It is the strongest baseline we tested: it
beats LLM-direct by $+0.0254$ ($p = 4\!\times\!10^{-4}$) and ties
}\texttt{\chg{additive}} \chg{($-0.0034$, n.s.), while }\texttt{\chg{mult\_offset}} \chg{still wins
($+0.0105$, $p = 0.049$). Note that the ratios must be profiled per pipeline, whereas ZEBRA's are estimated zero-shot (App.~\ref{app:cost_prop}).
Ablations }confirm that
curves, solver, allocator, \emph{and} per-task adaptation are each
required.

\subsection{Robustness, overhead, and assumptions}
\label{sec:results_robustness}

Methodology and full
details are in Appendices~\ref{app:sensitivity}-\ref{app:controller_compare},
\ref{app:overhead}-\ref{app:monotonicity}\chg{; Appendix~\ref{app:curve_validation}
collects the curve-validation evidence in one place.
}\textbf{(i) Curve-estimation noise \chg{and curve validation}.}
Multiplicative Gaussian noise on $(a_i, b_i)$ at
$\sigma \in \{0, 10, 50\}\%$ leaves performance close to the live
controller. At $\sigma = 50\%$ (parameters halved/doubled), the mean
score drops only $-1.4$pp ($p = 0.08$ vs.\ live, n.s.) and still
beats Uniform ($p = 0.03$). Swapping the controller LLM
(\texttt{gpt-4o} $\leftrightarrow$ \texttt{gpt-4o-mini}) leaves
per-task fractional allocations highly correlated
(per-phase-averaged Pearson $0.95$ on \texttt{additive}, $0.92$ on
\texttt{mult\_offset}; App.~\ref{app:controller_compare}).
\chg{Replacing the
controller with an }\emph{\chg{open-weight}} \chg{model (Qwen3.5-9B, DeepSeek-V4-Pro)
preserves the ordering: both ZEBRA variants beat LLM-direct in every
controller column, on APPS and on HotpotQA, so the advantage is not an
artifact of one model family (App.~\ref{app:open_weight}).
}A post-hoc calibration check
of the controller's curve estimates against the per-call quality
gains actually observed in run logs shows the predictions track
the measured curve shape well
(App.~\ref{app:curve_calibration}, Fig.~\ref{fig:refine_calibration}).
\chg{Two further checks isolate the curves themselves
(App.~\ref{app:curve_validation}). Under }\emph{\chg{correlated}} \chg{estimation error
(shared latent, $\rho = 0.7$) ZEBRA degrades gracefully: at $\sigma = 0.10$ the
drop is negligible ($-0.005$, $p = 0.34$) and it still beats LLM-direct
significantly ($+0.023$, $p = 0.012$); only at $\sigma = 0.50$ does it fall to
LLM-direct's level. Conversely, replacing a task's curves with another task's,
or permuting its curves across phases, each costs a significant amount
($-0.016$, $p = 0.018$; $-0.026$, $p = 2 \times 10^{-4}$) and erases the
significant margin over LLM-direct - so the estimates carry per-task and
per-phase signal, not merely a correct structural prior.
}\textbf{(ii) Controller overhead.} Curve estimation costs
$\sim$$33\%$ of the budget at $\alpha = 0.5$ for both
\texttt{additive} ($29.1\%$ of total spend) and \texttt{mult\_offset}
($29.2\%$); full per-strategy table in App.~\ref{app:overhead}. To
check that the allocation gain justifies this overhead we compare
ZEBRA at $\alpha = 0.5$ against \emph{Uniform at $\alpha = 0.8$}
(60\% more budget, zero overhead); both ZEBRA variants still win
significantly ($+0.027$, $p = 1.6\!\times\!10^{-4}$ for
\texttt{additive}; $+0.041$, $p < 10^{-6}$ for
\texttt{mult\_offset}).
\textbf{(iii) \chg{Candidate selection}.} \chg{To rule out that the advantage comes from
the scoring method, we re-score every run by the last candidate
the pipeline ships -- its last valid one -- with no post-hoc
selection. Both ZEBRA variants still beat LLM-direct on every benchmark and budget, significantly in each. Absolute scores
drop but the ordering and the size of the gaps do not
(App.~\ref{app:last_valid}).
}\textbf{\chg{(iv) End-to-end budgeting.}} \chg{The budget covers phase spend, with the
controller call reported separately as overhead. Under the stricter
accounting in which the controller is paid from the same budget, ZEBRA still
wins at $\alpha = 0.8$: $+0.014$ ($p = 0.014$) on $150$ APPS tasks, and
$+0.053$ and $+0.048$ F1 ($p = 2\!\times\!10^{-4}$) on HotpotQA. At
$\alpha = 0.5$ on HotpotQA the controller consumes too large a share of a
small budget and the advantage is no longer significant: ZEBRA pays for an
allocation, so it is worth running only when the budget is large enough that
something remains to allocate after the controller is paid
(App.~\ref{app:ctrl_budget}).
}

\textbf{\chg{(v) Cost normalization.}} \chg{Realized spend is not matched across strategies,
so we also report the cost-normalized view
(App.~\ref{app:cost_normalization}): }\texttt{\chg{mult\_offset}} \chg{beats LLM-direct on
score-per-dollar at $\alpha = 0.5$, while LLM-direct wins the other cells.
The difference is under-spending rather than over-spending - all strategies
allocate the same total, but LLM-direct utilizes only $70\%$ of its allocation
and strands $30.2\%$ of the budget on phases that cannot consume their cap
($40$--$55\%$ utilization on }\texttt{\chg{decompose}}\chg{$/$}\texttt{\chg{implement}}\chg{), against
$80\%$ utilization and $20.1\%$ stranded for ZEBRA.
}\textbf{\chg{(vi) Pipeline monotonicity.}} Recent work on
overthinking~\citep{chen2024overthinking,sui2025stop} shows
non-decreasing utility can fail for individual prompts. However, within-task
fixed-effects regressions on $18{,}000$ runs yield positive per-tier
slopes ($\hat\beta_{\text{easy}} = 9.4$,
$\hat\beta_{\text{med}} = 5.2$, $\hat\beta_{\text{hard}} = 1.8$;
all $p < 10^{-6}$), validating the assumption for our pipeline.

\subsection{Transfer to HumanEval~\citep{chen2021codex} and CodeContests~\citep{li2022alphacode}}
\label{sec:results_transfer}

We run two additional benchmarks at $\alpha = 0.5$
(Appendix~\ref{app:transfer}): HumanEval
($29$ stable tasks; $28$E), and
CodeContests ($23$ tasks; $8$E/$5$M/$10$H).
On HumanEval, \texttt{zebra-mult\_offset} beats LLM-direct by
$+0.025$ ($p_{\text{raw}} = 0.025$, W/L/T = $11$/$1$/$17$); on
CodeContests, both ZEBRA variants finish above LLM-direct in mean
and NB-retention, but neither reaches significance at
$n = 23$. However, the overall signal is consistent with
APPS: gains concentrate where the budget is most binding.

\subsection{Generality: a non-coding domain and a different pipeline (HotpotQA)}
\label{sec:results_hotpotqa}

The HumanEval and CodeContests transfers above hold the
\emph{coding} domain and a $4$-phase LangGraph pipeline fixed; they
test whether ZEBRA generalizes across coding benchmarks but not
whether it generalizes beyond coding or the orchestration
framework we built it on. To stress this we run a third transfer
that changes both: multi-hop question answering on HotpotQA's ~\citep{yang2018hotpotqa}, with a plain-Python
(no LangGraph) pipeline of three phases:
\texttt{direct} (one-shot answer) $\to$
\texttt{research} (sub-question decomposition that yields
\emph{evidence only}, not a candidate answer) $\to$
\texttt{synthesize} (combine the direct answer with the evidence,
keeping the direct answer unless contradicted). We screen \chg{$400$
}multi-hop tasks with the same protocol and retain \chg{$114$ }stable
tasks (Appendix~\ref{app:hotpotqa}); we compare both ZEBRA objectives,
LLM-direct, and a Uniform $\frac{1}{3}/\frac{1}{3}/\frac{1}{3}$
baseline at $\alpha = 0.5$, $15$ runs per cell, with token-F1
against the gold answer as the score. The solver and controller
code are unchanged from APPS. Full numbers are in
Table~\ref{tab:hotpotqa_overall}; per-phase allocations are in
Table~\ref{tab:hotpotqa_alloc}.

{\bf ZEBRA significantly beats LLM-direct on a non-coding
pipeline.} \texttt{zebra-additive} retains \chg{$92.4\%$ }of the
no-budget F1 versus \chg{$82.9\%$ }for LLM-direct, a \chg{$+9.5$}pp NB-retention
gap (corresponding to \chg{$+0.043$ }on absolute mean F1: \chg{$0.474$ }vs.\
\chg{$0.430$; $p = 0.0012$}, paired Wilcoxon on per-task mean
F1, \chg{$n = 114$}); the gap widens to \chg{$+13.6$}pp NB retention on the hard
tier (\chg{$p = 0.040$}). LLM-direct again uses a fixed, ill-calibrated
split -- this time overweighting \texttt{direct} (\chg{$\sim 67\%$}) and
starving \texttt{synthesize} (\chg{$\sim 11\%$}) -- so the same failure
mode we see on APPS reproduces on a domain and pipeline structure
with no shared phases or framework code
(Table~\ref{tab:hotpotqa_alloc}).

{\bf Interesting finding: ZEBRA and Uniform are statistically indistinguishable.} \texttt{zebra-additive}
vs.\ Uniform: \chg{$+0.05$}pp NB retention overall (\chg{$+0.002$ }on absolute
mean F1; \chg{$p = 0.28$}); per tier \chg{$+0.5$}pp easy (\chg{$p = 0.29$) and
$-0.5$}pp hard (\chg{$p = 0.58$}). Unlike APPS, where the right allocation is
strongly skewed toward \texttt{refine} and Uniform leaves $\sim 5$
points on the table, on HotpotQA the optimal allocation is close to
balanced -- the three QA phases contribute comparably to the final
F1, and any near-balanced split lands close to the right answer.
Crucially, \emph{ZEBRA discovers this on its own}: averaged across
all \chg{$114$ }tasks its allocation is \chg{$31/39/30$
}(Table~\ref{tab:hotpotqa_alloc}), within a few points of the
$33/33/33$ uniform reference. On APPS the same controller had
learned a sharply skewed $\sim 11/14/24/51$ split; here it
converges back toward balance because that is what the task
structure asks for.

{\bf Allocation is also consistent \emph{across tiers},
unlike on APPS.} On APPS, ZEBRA's per-phase shares move visibly
between easy and medium/hard tasks -- \texttt{refine} grows from
$\sim 30$--$45\%$ on easy to $\sim 60$--$66\%$ on medium/hard
(Sec.~\ref{sec:results_allocations},
Table~\ref{tab:allocation_distribution}). On HotpotQA the
allocation is essentially \emph{flat} across tiers: the per-phase
shares change by less than \chg{$\sim 1$}pp from easy to hard
(Table~\ref{tab:hotpotqa_alloc}; \texttt{additive} \chg{$31.0/38.7/30.3$ }easy
vs.\ \chg{$30.0/39.7/30.2$ }hard). Non-adaptation here is correct: on
this benchmark the optimal split does not change with
difficulty, and ZEBRA reflects that. The framework adapts when
this is warranted (APPS), and stays balanced otherwise
(HotpotQA), without that behavior being hard-coded. %

\section{Conclusion and Limitations}
\label{sec:conclusion}
On APPS, both
ZEBRA objectives outperform direct LLM allocation, %
and the gap widens as the budget tightens. %
This reproduces
on HumanEval/CodeContests, and on a non-coding %
HotpotQA pipeline. %
The
result also survives $50\%$ curve noise, controller-LLM swaps \chg{(including
open-weight controllers) }and
a Uniform comparator given $60\%$ more budget.
Studying the allocations shows \emph{why}:
ZEBRA adapts its phase split to task type, task difficulty, and
budget pressure -- skewed toward \texttt{refine} on APPS,
near-balanced on HotpotQA -- while LLM-direct's split is fixed and
miscalibrated in both regimes. More broadly, a small amount of
optimization-theoretic logic applied at inference time can
noticeably improve the economic behavior of autonomous multi-agent
systems.

\textbf{Limitations.} ZEBRA allocates once before execution;
recursive/online reallocation and budget-aware downstream agents (rather than capped LLM interfaces) are natural
extensions. We did explore a hybrid mid-pipeline re-allocation variant
on the APPS pipeline, but it did not improve in our ZEBRA case
(Appendix~\ref{app:hybrid}). The $\sim$$33\%$ controller overhead is empirically
justified (Appendix~\ref{app:overhead}), but can strain prohibitive
budgets. \chg{This delimits the regime in which ZEBRA is useful: the
controller call costs a roughly fixed amount per task - so its share of spend grows as the budget
tightens, and the allocation it buys repays that fixed price only once the
per-task budget is comfortably above it.  At the extreme, when a task's entire budget is about the price of a single controller call, no allocator can help. ZEBRA is therefore aimed at the higher-budget regime,
increasingly the common one as agentic pipelines grow longer and more
expensive, rather than at sub-cent budgets. }The framework
presupposes a pipeline of discrete phases, limiting applicability
to phaseless agentic loops.  
\textbf{Future work.} Several extensions are natural but not addressed
here: workflows with branching/merging structure expressed as DAGs; per-phase model choice; and explicit prefill/decode cost separation. %

\begin{ack}

\chg{We thank Rana Shahout for her thoughtful and insightful feedback, and Eden Saig for his helpful comments and suggestions. This work was supported by the European Research Council (ERC) under the European Union’s Horizon 2020 research and innovation program (grant agreement No. 101077862, project ALGOCONTRACT), by the Israel Science Foundation (grant No. 3331/24), by the NSF-BSF (grant No. 2021680), by a Google Research Scholar Award, by a Microsoft AIEI fellowship, and by the AISI Alignment Project. The authors declare no competing interests.
}\end{ack}

\bibliographystyle{abbrvnat}
\bibliography{references}

\newpage

\appendix

\section{Extended Related Work}
\label{app:related_work}

\subsection{Per-axis comparison with prior work}
\label{app:rw_table}

Table~\ref{tab:rw_comparison} contrasts ZEBRA with the closest budgeted-LLM
and resource-allocation methods discussed in
Section~\ref{sec:related_work} along four axes:
\textbf{Perspective} -- whether the budget is set from the
\emph{provider} side (an inference operator splitting compute across
many independent user queries) or the \emph{orchestration} side (a
controller allocating budget within a single multi-agent task);
\textbf{Decision (params)} -- the variable each method picks (its scope
and form, including whether it is discrete -- item/model selection -- or
continuous -- a real-valued budget share);
\textbf{Zero-shot} -- whether per-task training data, fine-tuning, or RL
is required;
\textbf{Dependency} -- whether the items sharing the budget are
independent (e.g., batched queries, weakly-coupled MDPs) or compose into
a single joint outcome (the failure of one phase voids the others).

\begin{center}
\scriptsize
\setlength{\tabcolsep}{3pt}
\renewcommand{\arraystretch}{1.18}
\begin{longtable}{@{}p{1.9cm} p{1.3cm} p{5.2cm} p{1.9cm} p{2.4cm}@{}}
\caption{Per-axis comparison of ZEBRA with prior budgeted-reasoning,
multi-agent cost-control, budget-aware agent-control, and
resource-allocation methods. Superscripts mark terms defined in the
legend below the table.}
\label{tab:rw_comparison} \\
\toprule
\textbf{Method} & \textbf{Persp.} & \textbf{Decision (params)} & \textbf{Zero-shot} & \textbf{Dependency} \\
\midrule
\endfirsthead
\toprule
\textbf{Method} & \textbf{Persp.} & \textbf{Decision (params)} & \textbf{Zero-shot} & \textbf{Dependency} \\
\midrule
\endhead
\bottomrule
\endlastfoot
\multicolumn{5}{@{}l}{\emph{Inference-time token budget allocation}} \\
TALE \citep{han2024tokenbudgetaware} & Provider & Discrete: per-query token budget, injected as prompt constraint & Both$^{a}$ & Independent queries \\
Pred.\ Scheduling \citep{brown2025predictivescheduling} & Provider & Discrete: greedy allocation of 16-token windows across queries & No (MLP on hidden states) & Independent queries \\
ROI-Reasoning \citep{zhao2026roireasoning} & Provider & Discrete: per-query token budget + skip decision (ordered-subset MCKP) & No (RL fine-tuning) & Independent queries \\
HBPO \citep{hbpo2025} & Provider & Discrete: per-query hierarchical token-budget tier & No (RL fine-tuning) & Independent queries \\
\midrule
\multicolumn{5}{@{}l}{\emph{Multi-agent cost management and routing}} \\
FrugalGPT \citep{chen2024frugalgpt} & Provider & Discrete: which model in a per-query cascade & No (trained cascade) & Independent queries \\
CoRL \citep{jin2025corl} & Provider & Discrete: which expert LLM to invoke per query & No (RL controller) & Independent queries \\
BAMAS \citep{yang2025bamas} & Orchestr. & Discrete: pre-execution choice of agent team (via ILP$^{b}$) + collaboration topology$^{c}$ (via RL) & No (ILP + RL) & Dependent (chosen agents compose a pipeline) \\
DAAO \citep{su2026daao} & Orchestr. & Discrete: per-query workflow structure -- which agents, in what arrangement & No (VAE difficulty estimator) & Dependent (chosen agents compose a workflow) \\
\midrule
\multicolumn{5}{@{}l}{\emph{Budget-aware agent control (inside a running agent)}} \\
BATS \citep{bats2025} & Orchestr. & Discrete per-step choice inside a single agent: ``dig deeper'' (verify/elaborate) vs.\ ``pivot'' (try a new branch) & Yes (plug-in) & Sequential single-agent \\
INTENT \citep{liu2026intent} & Orchestr. & Discrete: next tool call inside a single agent's online planning loop & No (hierarchical world model$^{d}$) & Sequential single-agent \\
BAVT \citep{li2026bavt} & Orchestr. & Discrete: which node of the single-agent multi-hop reasoning tree to expand next$^{e}$ & Yes (heuristic$^{f}$) & Sequential single-agent \\
\midrule
\multicolumn{5}{@{}l}{\emph{Resource allocation theory and knapsack methods}} \\
Budget MDPs \citep{boutilier2016budgetmdps} & Neither (Theory) & Discrete: budget chunks across weakly-coupled$^{g}$ MDPs via MCKP over piecewise-linear concave utility curves & No (known BMDP$^{h}$; offline DP) & Independent (weak coupling) \\
Knapsack RL \citep{li2025knapsackrl} & Neither (Training) & Discrete: RL rollout distribution across training tasks & No (training-time) & Independent training batches \\
\midrule
\textbf{ZEBRA (ours)} & \textbf{Orchestr.} & \textbf{Continuous} per-phase monetary budget across the composing phases of a multi-agent pipeline; nonlinear continuous knapsack solved via water-filling on the Lagrange multiplier & \textbf{Yes} -- LLM-elicited per-phase curves; no fine-tuning, RL, or historical data & \textbf{Dependent} -- multiplicative-product aggregation; one phase's failure voids the joint outcome \\
\end{longtable}
\end{center}

\vspace{0.2em}
{\noindent\scriptsize
\textbf{Legend.}
\textsuperscript{a}\,Both: TALE has a prompting variant (zero-shot) and a fine-tuning variant (not zero-shot).
\textsuperscript{b}\,ILP = Integer Linear Programming, used to choose which agents form the team subject to a budget constraint.
\textsuperscript{c}\,Topology = the message-passing graph specifying which agents communicate with which (the structure of the multi-agent collaboration).
\textsuperscript{d}\,Intention-aware hierarchical world model: a learned model trained on past tool-use episodes that predicts future tool calls, costs, and outcomes; planning is done against this learned model.
\textsuperscript{e}\,The reasoning tree models multi-hop reasoning as a tree where each node is a reasoning step; the search decides which child node to expand next.
\textsuperscript{f}\,Heuristic: BAVT rescales each node's value by the remaining-budget ratio (used as a scaling exponent) -- no learned component.
\textsuperscript{g}\,Weakly-coupled MDPs share only the budget constraint; their state, action, and reward dynamics are otherwise independent.
\textsuperscript{h}\,The method requires a fully specified Budgeted MDP (BMDP) model whose per-MDP value functions are precomputed offline via dynamic programming.
\par}

\subsection{Economic mechanisms in multi-agent systems}
\label{app:rw_economic}

\citep{hua2025shapleycoop} apply Shapley values to credit assignment and
pricing negotiation among self-interested agents, demonstrating that
principled algorithmic guidance can meaningfully improve coordination
outcomes in multi-agent systems.

\subsection{Detailed per-paper descriptions (full prose, moved from Section~\ref{sec:related_work})}
\label{app:rw_detail}

This subsection preserves the full per-paper exposition that appeared
in earlier drafts of Section~\ref{sec:related_work}; the body retains
only a citation-list summary.

\paragraph{Inference-time token budget allocation.}
A large body of work studies how inference operators can redistribute a fixed
token budget across independent user queries to maximize aggregate
accuracy.
Difficulty estimation is a common thread: \citep{han2024tokenbudgetaware}
estimate a per-query token budget via prompting or fine-tuning and inject it
as a reasoning constraint (TALE); \citep{brown2025predictivescheduling} train
an MLP on transformer hidden states to predict early-stopping probabilities,
then greedily allocate 16-token windows to whichever query yields the highest
marginal gain.
Methods that require RL fine-tuning go further by baking budget awareness
directly into model weights: \citep{zhao2026roireasoning} train a model to
predict token need and skip low-ROI queries (ROI-Reasoning);
\citep{hbpo2025} partition the token budget into hierarchical tiers and learn
per-tier reward structures (HBPO).
All four take the \emph{provider} perspective, operate on batches of
\emph{independent} queries, and pick a \emph{discrete} per-query token
budget; most also require task-specific training data or model
fine-tuning. ZEBRA differs on every axis (Table~\ref{tab:rw_comparison}):
it is an \emph{orchestration}-side controller that splits a
\emph{continuous} monetary budget across the \emph{dependent} phases of a
single multi-agent pipeline, zero-shot.

\paragraph{Multi-agent cost management and routing.}
A parallel line reduces cost through discrete model selection and
agent routing rather than continuous budget allocation:
\citep{chen2024frugalgpt} build an LLM cascade that escalates to
expensive models only when needed (FrugalGPT);
\citep{jin2025corl} train an RL controller to pick which expert LLM
to invoke per query under a budget (CoRL); \citep{yang2025bamas}
combine ILP-based agent selection with RL-fixed collaboration
topology (BAMAS); \citep{su2026daao} use a VAE difficulty estimator
to assemble per-query workflows (DAAO);
\citep{kandogan2025orchestrating} treat budget as a constraint in
enterprise pipeline design; and \citep{amayuelas2025selfresource}
find that upfront worker-model selection beats reactive
orchestration. In every case the decision variable is
\emph{discrete and combinatorial} and the controller is trained
(Table~\ref{tab:rw_comparison}).

\paragraph{Budget-aware agent control.}
A concurrent line studies budget-aware control at inference time
\emph{inside} a single agent. BATS~\citep{bats2025} pairs a Budget
Tracker plug-in with a per-step ``dig deeper vs.\ pivot'' decision
under a remaining tool-call budget; INTENT~\citep{liu2026intent}
trains an intention-aware hierarchical world model that
risk-calibrates next-tool calls online for a single agent under a
strict monetary budget; BAVT~\citep{li2026bavt} models multi-hop
reasoning as a dynamic value tree and uses the remaining-resource
ratio as a scaling exponent on node values. All three are
\emph{step-by-step} controllers inside a single running agent.
ZEBRA targets a different and complementary problem \emph{above}
the pipeline: before any agent runs, it solves a one-shot
continuous knapsack across all phases. The two layers stack --
ZEBRA's per-phase budgets define the caps within which a step-level
controller of this kind would operate.

\paragraph{Resource allocation theory and knapsack methods.}
The knapsack problem and its variants provide the theoretical
backbone for budget
allocation~\citep{katohIbarakiHandbook, kellerer2004knapsack,
boydVandenberghe}. \citet{boutilier2016budgetmdps} allocate a fixed
budget across weakly-coupled MDPs via Multiple-Choice Knapsack
(MCKP), assuming full independence between subprocesses and
predating LLMs entirely. Recent LLM work has adopted knapsack
formulations for related problems:
\citep{zhao2026roireasoning} use an ordered-subset MCKP across
independent queries, \citep{brown2025predictivescheduling} solves a
greedy knapsack over fixed-size token windows, and
\citep{li2025knapsackrl} models RL rollout distribution as a
knapsack. All operate on independent queries or training batches
rather than across the composing phases of a single running
pipeline.

\paragraph{Hybrid LLM + knapsack-solver systems and LLM weakness on
NP-hard problems.}
The methodological precedent closest to ZEBRA is
SEER~\citep{tonglet2023seer}, which formulates in-context exemplar
selection for HybridQA as an integer-linear-program knapsack with
diversity and capacity constraints, demonstrating that wrapping an
LLM system with an explicit combinatorial-optimization layer beats
heuristic selection. SEER and ZEBRA share the high-level move --
delegate a constrained selection/allocation step to a knapsack
solver rather than to the LLM itself -- but differ in scope and
problem class. SEER addresses one-prompt, discrete exemplar choice
over an independent candidate set; ZEBRA addresses zero-shot,
continuous, orchestration-time monetary-budget allocation across the
\emph{dependent} phases of a multi-phase pipeline whose outputs
compose into a single joint outcome. The choice to delegate the
allocation step at all is independently supported by
\citet{duchnowski2025knapsack} (EHOP), who introduce a benchmark of
NP-hard problems -- including knapsack -- expressed in natural
language and show that frontier LLMs, including reasoning models,
systematically solve textbook formulations more accurately than
real-life or rule-inverted variants and exhibit high variance across
surface presentations. Their finding -- that current LLMs lack a
robust, presentation-invariant solver for NP-hard problems --
motivates ZEBRA's design.

\clearpage
\section{Background and Method Details}
\label{app:opt_details}
\label{app:method_details}
\subsection{\chg{The full ZEBRA allocation procedure}}
\label{app:zebra_algorithm}

\chg{Algorithm~\ref{alg:zebra} states the complete allocation procedure in one
place: curve elicitation from the controller (Section~\ref{sec:curves}), the
choice of aggregation objective and its closed-form per-phase response
(Section~\ref{sec:objectives}), and the dual search that enforces the budget
(Section~\ref{sec:dual}).  The controller is queried exactly once per
task, before any phase runs; everything after that query is deterministic
arithmetic, and the solver itself is classical, the contribution is the
reduction, not the water-filling. For compactness the algorithm states both
objectives in terms of the single offset curve
$g_i(x) = 1 - a_i e^{-b_i x}$ of \eqref{eq:offset}: this is exactly the
multiplicative case, and it also covers the additive case of
\eqref{eq:zebra_additive}, since $g_i$ and $f_i$ differ only by the additive
constant $1 - a_i$ and therefore have identical derivatives and identical
maximizers.
}

\begingroup\chgon
\begin{algorithm}[t]\chgon 
\caption{ZEBRA allocation}
\label{alg:zebra}
\begin{algorithmic}
\Statex \textbf{Input:}\ \ budget $B$; per-phase curves $(a_i, b_i)$ estimated by
        prompting the LLM controller
\Statex \textbf{Output:}\ allocation $\{x_i\}$ with $\sum_i x_i = B$
\Statex
\Statex \textbf{Per-phase quality curve:}\quad
        $g_i(x) \;=\; 1 - a_i \cdot e^{-b_i x}$
\Statex
\Statex \textbf{Objective} (separable concave program):
\Statex \quad $\max\ \sum_i h_i(x_i)$
        \qquad s.t.\ $\sum_i x_i = B$, \quad $x_i \ge 0$
\Statex \quad\quad \makebox[6.2em][l]{additive:} $h_i = g_i$
\Statex \quad\quad \makebox[6.2em][l]{multiplicative:} $h_i = \log g_i$
        \qquad (i.e.\ $\max \prod_i g_i$)
\Statex
\Statex KKT $\Rightarrow$ $h_i'(x_i) = \lambda$ at the optimum. Invert for $x_i(\lambda)$:
\Statex \quad\quad \makebox[6.2em][l]{additive:}
        $x_i(\lambda) = \max\bigl(0,\ (1/b_i) \cdot \ln( a_i b_i / \lambda )\bigr)$
\Statex \quad\quad \makebox[6.2em][l]{multiplicative:}
        $x_i(\lambda) = \max\bigl(0,\ (1/b_i) \cdot \ln( a_i (b_i + \lambda) / \lambda )\bigr)$
\Statex Phase $i$ is starved ($x_i = 0$) once $\lambda \ge h_i'(0)$.
\Statex
\Statex \textit{Bisection on the Lagrange multiplier $\lambda$ (the budget's price)}
\State $\lambda_{\mathrm{lo}} \gets 0$
\State $\lambda_{\mathrm{hi}} \gets \max_i h_i'(0)$
       \Comment{$\lambda_{\mathrm{hi}} \Rightarrow$ every $x_i = 0$}
\Repeat
  \State $\lambda \gets (\lambda_{\mathrm{lo}} + \lambda_{\mathrm{hi}}) / 2$
  \If{$\sum_i x_i(\lambda) > B$}
    \State $\lambda_{\mathrm{lo}} \gets \lambda$
           \Comment{overspending $\rightarrow$ raise price}
  \Else
    \State $\lambda_{\mathrm{hi}} \gets \lambda$
           \Comment{underspending $\rightarrow$ lower price}
  \EndIf
\Until{$\lambda_{\mathrm{hi}} - \lambda_{\mathrm{lo}} < \varepsilon$}
\Statex
\State \Return $\{x_i(\lambda^\star)\}$, rescaled so $\sum_i x_i = B$ exactly
\end{algorithmic}
\end{algorithm}
\endgroup

\subsection{Full assumptions and convexity (from Section~\ref{sec:cnk})}
\label{app:opt_full_assumptions}

We restate the full optimization problem with optional per-phase upper bounds:
\begin{equation}
  \max_{\mathbf{x}} \;\sum_{i=1}^n f_i(x_i) \quad \text{s.t.} \quad
  \sum_{i=1}^n x_i \le B, \;\; 0 \le x_i \le u_i \;\forall i,
  \label{eq:primal_full}
\end{equation}
where $u_i$ is an optional per-phase upper bound (e.g., a hard cap on tokens
or tool calls).

We assume each $f_i$ is non-decreasing (more budget never hurts), concave
(diminishing returns), and differentiable. Under these assumptions
$\sum_i f_i(x_i)$ is concave on a convex feasible set, so
\eqref{eq:primal_full} is a convex program with no spurious local optima
and with optimality fully characterized by KKT
conditions~\citep{boydVandenberghe}. Without a fully convex feasible set,
related continuous knapsack variants can require nonconvex
machinery~\citep{concaveCardinalityKnapsack}.

\subsection{KKT conditions and dual-search algorithm (full statement, from Section~\ref{sec:dual})}
\label{app:opt_dual}

Introduce a Lagrange multiplier $\lambda \ge 0$ for the budget constraint and
box-multipliers for $0 \le x_i \le u_i$. Because the problem is convex, the
KKT conditions are necessary and sufficient~\citep{boydVandenberghe} and yield
the familiar \emph{marginal equalization} rule
\begin{equation}
  f_i'(x_i^\star) = \lambda \quad \text{for } 0 < x_i^\star < u_i,
  \label{eq:marginal_full}
\end{equation}
with the natural boundary modifications $f_i'(x_i^\star) \le \lambda$ at
$x_i^\star = 0$ and $f_i'(x_i^\star) \ge \lambda$ at $x_i^\star = u_i$.
Economically, $\lambda$ is the shadow price of budget: budget flows to each
phase until all \emph{active} phases have equal marginal utility, matching
the water-filling solution from communication
systems~\citep{boydVandenberghe}.

Defining the per-phase response
$x_i(\lambda) = \arg\max_{0 \le x \le u_i} \bigl(f_i(x) - \lambda x\bigr)$,
the total allocation $S(\lambda) = \sum_i x_i(\lambda)$ is non-increasing in
$\lambda$, so $\lambda^\star$ can be found by bisection in
$O(n \log\tfrac{1}{\varepsilon})$ time, matching the classical
resource-allocation result~\citep{katohIbarakiHandbook,katohIbarakiMIT}.

\paragraph{From products to sums.}
A weighted product $\prod_i h_i(x_i)^{w_i}$ with positive concave $h_i$
reduces, via $\log$, to $\sum_i w_i \log h_i(x_i)$, again a separable
concave program solvable by the same dual-search machinery. ZEBRA uses this
transformation to handle all multiplicative and offset objectives of
Section~\ref{sec:method} with one solver; full derivation in
Appendix~\ref{app:log_transform}.

\subsection{Saturating-exponential derivatives (from Section~\ref{sec:curves})}
\label{app:method_derivatives}

The first and second derivatives of the saturating exponential
$f_i(x) = a_i(1 - e^{-b_i x})$ are
\begin{equation}
f_i'(x) = a_i b_i e^{-b_i x}, \qquad
f_i''(x) = -a_i b_i^2 e^{-b_i x} \le 0,
\end{equation}
which establish that $f_i$ is monotone non-decreasing and concave for
$a_i, b_i > 0$.

\subsection{Two-point curve fitting (full derivation, from Section~\ref{sec:curves})}
\label{app:method_twopoint}

Rather than asking the controller for abstract curve parameters, we elicit
two intuitive operating points per phase. The controller estimates:
(i) $n_{\text{basic},i}$, the total output tokens the phase needs to
reach roughly $50\%$ of its potential quality, and (ii) $n_{\text{great},i}$,
the total output tokens to reach roughly $90\%$ (named \texttt{tokens\_basic}
and \texttt{tokens\_great} in the controller prompt;
App.~\ref{app:example:prompts}). These correspond to the equations
$f_i(n_{\text{basic}}) = 0.5\,a_i$ and
$f_i(n_{\text{great}}) = 0.9\,a_i$, yielding two estimates:
\begin{equation}
b_{\text{basic}} = \frac{\ln 2}{n_{\text{basic}}}, \qquad
b_{\text{great}} = \frac{\ln 10}{n_{\text{great}}}.
\end{equation}
We combine these as
$b_i = \sqrt{b_{\text{basic}} \cdot b_{\text{great}}}$.
Because the global budget $B$ is monetary (USD), we convert each phase's
token operating points to USD via that phase's model pricing -- which
differs across phases (e.g., \texttt{gpt-4o} for \texttt{refine} vs.\
\texttt{gpt-4o-mini} for the others) -- before computing $b_i$, so that
$b_i$ has units of inverse-USD and the per-phase response $x_i(\lambda)$
in Section~\ref{sec:additive} is in USD. The equations above are
unit-agnostic: the same expression for $b$ holds whether $n$ is in tokens
or in dollars; the conversion just rescales $b$ by the per-phase USD-per-token rate.
In addition to $b_i$, the controller also estimates $a_i \in (0,1]$, the
quality ceiling of each phase. (A propagation-weighted variant that reuses
$a_i$ as a per-phase weight on the multiplicative offset objective is
defined and evaluated in Appendix~\ref{app:prop_offset}.)

\subsection{Multiplicative offset closed form (from Section~\ref{sec:mult_offset})}
\label{app:method_mult_closed_form}

Taking logs (Appendix~\ref{app:log_transform}) reduces \eqref{eq:mult_offset}
to the separable concave program $\max \sum_i \log g_i(x_i)$, solvable by
the same dual search. The KKT condition $g_i'(x_i)/g_i(x_i) = \lambda$
yields the closed-form per-phase response
\begin{equation}
x_i(\lambda) \;=\; \max\!\Bigl(0,\, \tfrac{1}{b_i}\ln\tfrac{a_i(b_i + \lambda)}{\lambda}\Bigr).
\end{equation}
Phase $i$ is starved exactly when its \emph{log-marginal at zero},
$g_i'(0)/g_i(0) = a_i b_i / (1 - a_i)$, falls below the shadow price
$\lambda$. The form is parallel to the additive case (marginal at zero
$f_i'(0) = a_i b_i$ vs.\ $\lambda$): in both objectives, a phase is starved
when its initial return on budget is too low to beat the going price. The
shift from $f'$ to $g'/g$ is just the chain rule applied to $\log g_i$,
since the multiplicative objective is a sum of \emph{logs} after the
transformation.

\section{Logarithmic Transformation Derivation}
\label{app:log_transform}

A natural extension of the additive formulation in
Section~\ref{sec:cnk} replaces the sum with a product:
\begin{equation}
\max_{\mathbf{x}} \quad \prod_{i=1}^{n} h_i(x_i) \quad
\text{s.t.}\ \sum_{i=1}^n x_i \le B,\ x_i \ge 0,
\label{eq:product_obj}
\end{equation}
where each $h_i$ is positive, concave, and non-decreasing. Because the
logarithm is strictly increasing, the product objective shares its
maximizer with the log-sum:
\begin{equation}
\max_{\mathbf{x}} \quad \sum_{i=1}^{n} \log h_i(x_i) \quad
\text{s.t.}\ \sum_{i=1}^n x_i \le B,\ x_i \ge 0.
\label{eq:log_obj}
\end{equation}
When each $h_i$ is positive and log-concave (which is guaranteed when
$h_i$ is concave and positive), each term $\log h_i(x_i)$ is concave,
so~\eqref{eq:log_obj} is again a separable concave program with a
linear constraint. The water-filling machinery of
Section~\ref{sec:dual} applies directly: one searches for a Lagrange
multiplier $\lambda$ such that the log-marginals
$\frac{d}{dx}\log h_i(x_i) = h_i'(x_i)/h_i(x_i)$ are equalized across
active phases.

More generally, a \emph{weighted} product
$\prod_i h_i(x_i)^{w_i}$ with $w_i > 0$ transforms to
$\sum_i w_i \log h_i(x_i)$, and the optimality condition becomes
$w_i \cdot h_i'(x_i) / h_i(x_i) = \lambda$ for all active phases. This
observation is central to ZEBRA: it allows us to solve multiplicative,
propagation, and offset objectives (Section~\ref{sec:method}) using the
same dual-search template as the additive case, differing only in the
closed-form expression for $x_i(\lambda)$ at each step.

\section{Additional Experiment Results}
\label{app:additional_results}
\subsection{Pipeline phases}
\label{app:phases}
See Table~\ref{tab:phases} for details. 
\begin{table}[t]
\centering
\small
\setlength{\tabcolsep}{6pt}
\renewcommand{\arraystretch}{1.15}
\begin{tabularx}{\columnwidth}{@{} l l X @{}}
\toprule
\textbf{Phase} & \textbf{Model} & \textbf{Purpose} \\
\midrule
{\ttfamily plan}      & \texttt{gpt-4o-mini} & Produce a high-level plan for solving the task. \\
{\ttfamily decompose} & \texttt{gpt-4o-mini} & Decompose the plan into implementable subtasks. \\
{\ttfamily implement} & \texttt{gpt-4o-mini} & Write solution code across subtasks. \\
{\ttfamily refine}    & \texttt{gpt-4o}      & Review$\to$revise loop: identify real bugs via spec comparison, then fix flagged issues only. Up to 3 iterations. \\
\bottomrule
\end{tabularx}
\caption{Workflow phases shared by all strategies. The refine phase uses a stronger model (\texttt{gpt-4o}) for higher-quality code review; all other phases use \texttt{gpt-4o-mini}.}
\label{tab:phases}
\end{table}

\subsection{Benchmark construction details}
\label{app:benchmark_construction}

\paragraph{Raw pool.} We start from the APPS benchmark~\citep{hendrycks2021apps}
at the \emph{interview} difficulty level. To keep scoring reliable we
restrict to tasks with enough test cases to distinguish near-solutions
from fully-correct ones. We use two ranges of test-suite size: tasks
with at least $100$ tests ($103$ tasks) and tasks
with $50$--$99$ tests ($338$ tasks\chg{)},  \chg{both extracted by selecting
}\emph{\chg{interview}}\chg{-level APPS problems according to the number of test cases}.

\paragraph{Screening for stability.} We screen every task in both pools
with $30$ independent \texttt{no\_budget} runs of the same pipeline used
for the benchmark, using the same models. For each task we record the
mean and coefficient of variation of cost and of score across the $30$
runs. A task is retained if $\mathrm{CV}_{\text{cost}} < 0.35$ and its
cost variance is non-zero (trivially solved tasks are excluded). This
retains $49$ of $103$ tasks in the $100+$-test pool and $170$ of $338$
tasks in the $50$--$99$-test pool. We save the per-task means.

\paragraph{Balancing across tiers.} We assign each retained task a
difficulty tier from its mean no-budget score $\bar{s}$:
\emph{easy} ($\bar{s} \ge 0.8$), \emph{medium} ($0.4 \le \bar{s} < 0.8$),
\emph{hard} ($\bar{s} < 0.4$). The $49$-task pool distributes as
$22$~easy / $20$~medium / $7$~hard, which is too hard-light for a
per-tier analysis. We therefore sample from the $50$--$99$-test stable
pool to fill each tier to $50$ tasks:

\begin{center}
\small
\begin{tabular}{l c c c}
\toprule
Tier & Pool 1 (100+) & Pool 2 (50--99) & Total \\
\midrule
Easy    & $22$ & $28$ & $50$ \\
Medium  & $20$ & $30$ & $50$ \\
Hard    & $7$  & $43$ & $50$ \\
\midrule
Total   & $49$ & $101$ & $150$ \\
\bottomrule
\end{tabular}
\end{center}

\paragraph{Job counts.} The main benchmark comprises
$150 \times 2 \times 3 \times 15 = 13{,}500$ runs for the main
($\alpha \in \{0.5, 0.8\}$) portion plus $50 \times 3 \times 15 = 2{,}250$
runs for the $\alpha = 0.3$ easy-tasks portion, on top of the
$150 \times 30 = 4{,}500$ \texttt{no\_budget} screening runs.

\subsection{Why we drop the budget-aware $a$-rescaling}
\label{app:rescale_drop}

An earlier draft of ZEBRA included a budget-aware rescaling of the
quality ceilings, $a'_i = a_i^{\kappa}$ with
$\kappa = \max(1,\, 1 + 2\ln\tfrac{4}{c})$, intended to let the offset
objectives meaningfully starve phases when the budget is tight. We ran
the full $150$-task benchmark with and without rescaling on
\texttt{prop\_offset} and \texttt{mult\_offset}; the no-rescale
($\kappa \equiv 1$) variants matched or beat the rescaling variants on
every aggregate metric at both $\alpha \in \{0.5, 0.8\}$. The
rescaling step is therefore dropped throughout this paper, and all
results in Section~\ref{sec:results_main}--%
\ref{sec:results_ablations} use the simpler $\kappa \equiv 1$
formulation. Removing the rescaling has the additional benefit that
the curves estimated by the controller are used as-is, without a
budget-dependent post-hoc transformation, which makes the controller's
output easier to inspect.

\subsection{Paired Wilcoxon details}
\label{app:wilcoxon}

For every comparison in
Sections~\ref{sec:results_main}--\ref{sec:results_allocations}, we run
paired Wilcoxon signed-rank tests on three per-task metrics (mean
score, median score, success rate), pairing each ZEBRA variant against
LLM-direct on the same tasks. Below we report the mean-score test
results in full; the median-score and success-rate tests are
qualitatively similar (always significant where the mean test is
significant).

The $18$ tests in Table~\ref{tab:wilcoxon} form a single corrected
family. We report the raw two-sided $p$-value
($p_{\text{raw}}$) alongside the Benjamini--Hochberg adjusted
$p$-value ($p_{\text{BH}}$) at $q = 0.05$ over $m = 18$ tests; the
significance markers in the body and in this table reflect
$p_{\text{BH}}$. Of the $13$ raw-significant cells, all $13$ remain
significant under BH-FDR.

\begin{table*}[t]
\centering
\small
\setlength{\tabcolsep}{5pt}
\renewcommand{\arraystretch}{1.1}
\begin{tabular}{l l c c c c c}
\toprule
Condition & Strategy & Mean-diff & W/L/T & $p_{\text{raw}}$ & $p_{\text{BH}}$ & Sig. \\
\midrule
$\alpha=0.5$, all ($n=150$)   & zebra-additive     & $+0.0220$ & $74/38/38$ & $1.27\!\times\!10^{-3}$ & $4.59\!\times\!10^{-3}$ & ** \\
$\alpha=0.5$, all ($n=150$)   & zebra-mult\_offset & $+0.0359$ & $79/34/37$ & $2.63\!\times\!10^{-7}$ & $4.74\!\times\!10^{-6}$ & *** \\
\midrule
$\alpha=0.8$, all ($n=150$)   & zebra-additive     & $+0.0216$ & $77/38/35$ & $8.36\!\times\!10^{-5}$ & $5.01\!\times\!10^{-4}$ & *** \\
$\alpha=0.8$, all ($n=150$)   & zebra-mult\_offset & $+0.0187$ & $75/39/36$ & $6.95\!\times\!10^{-4}$ & $3.13\!\times\!10^{-3}$ & ** \\
\midrule
$\alpha=0.5$, easy ($n=50$)   & zebra-additive     & $-0.0110$ & $11/10/29$ & $0.1909$                & $0.2291$                & ns \\
$\alpha=0.5$, easy ($n=50$)   & zebra-mult\_offset & $+0.0060$ & $13/7/30$  & $0.1790$                & $0.2291$                & ns \\
$\alpha=0.5$, medium ($n=50$) & zebra-additive     & $+0.0491$ & $32/14/4$  & $2.48\!\times\!10^{-3}$ & $7.43\!\times\!10^{-3}$ & ** \\
$\alpha=0.5$, medium ($n=50$) & zebra-mult\_offset & $+0.0673$ & $34/12/4$  & $4.09\!\times\!10^{-5}$ & $3.68\!\times\!10^{-4}$ & *** \\
$\alpha=0.5$, hard ($n=50$)   & zebra-additive     & $+0.0278$ & $31/14/5$  & $4.22\!\times\!10^{-3}$ & $7.95\!\times\!10^{-3}$ & ** \\
$\alpha=0.5$, hard ($n=50$)   & zebra-mult\_offset & $+0.0344$ & $32/15/3$  & $4.42\!\times\!10^{-3}$ & $7.95\!\times\!10^{-3}$ & ** \\
\midrule
$\alpha=0.8$, easy ($n=50$)   & zebra-additive     & $+0.0051$ & $11/13/26$ & $0.6071$                & $0.6428$                & ns \\
$\alpha=0.8$, easy ($n=50$)   & zebra-mult\_offset & $+0.0076$ & $15/9/26$  & $0.2652$                & $0.2983$                & ns \\
$\alpha=0.8$, medium ($n=50$) & zebra-additive     & $+0.0418$ & $34/12/4$  & $4.27\!\times\!10^{-3}$ & $7.95\!\times\!10^{-3}$ & ** \\
$\alpha=0.8$, medium ($n=50$) & zebra-mult\_offset & $+0.0343$ & $32/17/1$  & $0.0219$                & $0.0303$                & * \\
$\alpha=0.8$, hard ($n=50$)   & zebra-additive     & $+0.0179$ & $32/13/5$  & $4.07\!\times\!10^{-3}$ & $7.95\!\times\!10^{-3}$ & ** \\
$\alpha=0.8$, hard ($n=50$)   & zebra-mult\_offset & $+0.0142$ & $28/13/9$  & $0.0187$                & $0.0296$                & * \\
\midrule
$\alpha=0.3$, easy ($n=50$)   & zebra-additive     & $+0.0154$ & $19/8/23$  & $0.0197$                & $0.0296$                & * \\
$\alpha=0.3$, easy ($n=50$)   & zebra-mult\_offset & $+0.0026$ & $11/15/24$ & $0.8291$                & $0.8291$                & ns \\
\bottomrule
\end{tabular}
\caption{\textbf{Paired Wilcoxon signed-rank tests on per-task mean
score, ZEBRA variants vs LLM-direct.} Two-sided test. $p_{\text{raw}}$
is the unadjusted Wilcoxon $p$-value; $p_{\text{BH}}$ is the
Benjamini--Hochberg adjusted $p$-value over the $m = 18$ tests in this
family. Sig.\ codes use $p_{\text{BH}}$:
$^{*}p<0.05$, $^{**}p<0.01$, $^{***}p<0.001$, ns~=~not significant.
W/L/T = number of tasks where ZEBRA wins / loses / ties LLM-direct on
per-task mean.}
\label{tab:wilcoxon}
\end{table*}

\subsection{API model snapshots and sampling parameters}
\label{app:model_snapshots}

All experiments use OpenAI's Chat Completions API. To ensure
reproducibility, we report the exact model snapshots resolved by
each alias during our experiment window:

\begin{center}
\small
\begin{tabular}{l l l}
\toprule
Alias & Snapshot & Used for \\
\midrule
\texttt{gpt-4o-mini} & \texttt{gpt-4o-mini-2024-07-18} &
  \texttt{plan}, \texttt{decompose}, \texttt{implement} \\
\texttt{gpt-4o}      & \texttt{gpt-4o-2024-08-06}      &
  \texttt{refine}, controller \\
\bottomrule
\end{tabular}
\end{center}

Snapshots were confirmed via OpenAI's published model lifecycle
documentation. Sampling temperature is set per call type to balance
determinism (controller decisions, code generation) against diversity
(planning):

\begin{center}
\small
\begin{tabular}{l c}
\toprule
Call type & \texttt{temperature} \\
\midrule
\texttt{plan}                              & $0.7$ \\
\texttt{decompose}                         & $0.5$ \\
\texttt{implement}, \texttt{refine}, \texttt{review} & $0.3$ \\
Controller (ZEBRA, LLM-direct allocator)   & $0.3$ \\
\bottomrule
\end{tabular}
\end{center}

\texttt{top\_p} is left at the SDK default ($1.0$). Run-to-run
variability is addressed by averaging over $15$ independent runs per
(task, $\alpha$, strategy) condition.

\subsection{Ablations on the allocation step}
\label{app:ablations}
\label{sec:appendix_ablation}

We hold the rest of the pipeline fixed (same phases, same models,
same total budget, same prompts) and replace only ZEBRA's allocator,
to isolate the contribution of (i) the knapsack solver and (ii) the
phase utility curves themselves. Both ablations use
\texttt{zebra-mult\_offset} (no rescale) as the reference strategy.
Results are summarized in Table~\ref{tab:ablations}.

\paragraph{Ablation 1: LLM allocator on the same fitted curves.}
We give the controller (\texttt{gpt-4o}, same model used by ZEBRA's
estimation step) the same fitted per-phase curves
$\{(a_i, b_i)\}_{i=1}^n$ that ZEBRA estimates and ask it to output an
allocation directly. This isolates the contribution of the
\emph{optimization step}: any difference between this ablation and
ZEBRA must come from the controller's ad-hoc allocation versus the
knapsack solution on identical inputs. The controller's allocations
score $-4.20$ points worse at $\alpha = 0.5$ ($0.4977$ vs.\ $0.5397$)
and $-4.31$ points worse at $\alpha = 0.8$ ($0.5146$ vs.\ $0.5576$),
both significant at $p < 10^{-4}$ ($n = 150$ paired Wilcoxon
signed-rank).

\paragraph{Ablation 2: Uniform $25/25/25/25$ split.}
We discard the curves entirely and use a flat per-phase split. This
isolates the contribution of \emph{curve-aware allocation}: the gap to
ZEBRA reflects the value of estimating phase utilities at all. The
uniform split scores $-5.14$ points worse at $\alpha = 0.5$
($0.4883$ vs.\ $0.5397$) and $-5.92$ points worse at $\alpha = 0.8$
($0.4985$ vs.\ $0.5576$), again significant at $p < 10^{-4}$.

\paragraph{Ablation 3: Chain-of-thought prompting on LLM-direct.}
A natural concern is whether LLM-direct's underperformance
in Table~\ref{tab:main_overall} simply reflects weak prompting -- if
the controller were asked to think step-by-step before allocating,
would the gap close? We test this directly with
\textsc{LLM-CoT}, identical to LLM-direct but with a chain-of-thought
scaffold (the prompt is reproduced in
Appendix~\ref{app:examples_llm_cot}). We run the full $150$-task
benchmark at $\alpha = 0.5$. The result is unambiguous: \textbf{CoT
does not help.} \textsc{LLM-CoT} scores $0.4968$ overall vs.\ $0.5038$
for plain LLM-direct ($\Delta = -0.0070$, paired Wilcoxon
$p_{\text{raw}} = 0.054$, n.s., $n = 150$); the trend is in the wrong
direction (W/L/T = $42$/$75$/$33$), not the right one. ZEBRA still
beats CoT by a wide margin: \texttt{zebra-mult\_offset} scores
$+0.0429$ over \textsc{LLM-CoT} (W/L/T = $90$/$29$/$31$, $p_{\text{raw}}
< 10^{-4}$) and \texttt{zebra-additive} scores $+0.0290$ over
\textsc{LLM-CoT} ($p_{\text{raw}} = 0.0006$). The gap is therefore
not a prompting artefact; the LLM controller cannot match the knapsack
solution even with explicit reasoning steps.

\paragraph{Takeaway.} All three ingredients matter independently of one
another. Removing the solver costs $\sim$$4$ points; removing the
curves costs $\sim$$5$--$6$ points; adding CoT to the LLM-direct
controller does not recover any of the gap. Together these results
explain why LLM-direct in
Tables~\ref{tab:main_overall}--\ref{tab:main_by_tier} underperforms
even when it has access to the same controller model: even given the
fitted curves \emph{and} explicit reasoning steps, the controller
cannot match the knapsack solution.

\begin{table*}[t]
\centering
\small
\setlength{\tabcolsep}{6pt}
\renewcommand{\arraystretch}{1.15}
\begin{tabularx}{\textwidth}{@{} l l
  >{\centering\arraybackslash}X
  >{\centering\arraybackslash}X
  >{\centering\arraybackslash}X
  >{\centering\arraybackslash}X @{}}
\toprule
$\boldsymbol{\alpha}$ & \textbf{Allocator (curves used?)} &
\textbf{Mean} & \textbf{Succ\%} & $\boldsymbol{\Delta}$\textbf{Mean} & $\boldsymbol{p}$ \\
\midrule
\multirow{3}{*}{$0.5$}
& ZEBRA knapsack (yes)        & $\mathbf{0.5397}$ & $\mathbf{54.1\%}$ & ---       & ---               \\
& LLM, same curves            & $0.4977$          & $48.4\%$          & $-0.0420$ & $<\!10^{-4\,***}$ \\
& Uniform $25/25/25/25$       & $0.4883$          & $47.2\%$          & $-0.0514$ & $<\!10^{-4\,***}$ \\
\midrule
\multirow{3}{*}{$0.8$}
& ZEBRA knapsack (yes)        & $\mathbf{0.5576}$ & $\mathbf{55.2\%}$ & ---       & ---               \\
& LLM, same curves            & $0.5146$          & $50.5\%$          & $-0.0431$ & $<\!10^{-4\,***}$ \\
& Uniform $25/25/25/25$       & $0.4985$          & $48.2\%$          & $-0.0592$ & $<\!10^{-4\,***}$ \\
\bottomrule
\end{tabularx}
\caption{\textbf{Ablations on the allocation step.} Reference is
\texttt{zebra-mult\_offset}. Both
ablations hold the rest of the pipeline fixed (same phases, same
models, same budget, same prompts) and replace only ZEBRA's allocator.
Ablation~1 hands the controller the same fitted phase curves and asks
it to output an allocation directly, isolating the contribution of the
knapsack solver. Ablation~2 ignores curves entirely and uses a flat
$25\%$ split per phase, isolating the contribution of curve-aware
allocation. Both ablations are significantly worse than ZEBRA at both
budget levels (paired Wilcoxon signed-rank one-sided $p < 10^{-4}$,
$n = 150$). }
\label{tab:ablations}
\end{table*}

\subsection{Empirical validation of controller phase utility curves}
\label{app:curve_calibration}

\begin{figure}[t]
  \centering
  \includegraphics[width=0.98\textwidth]{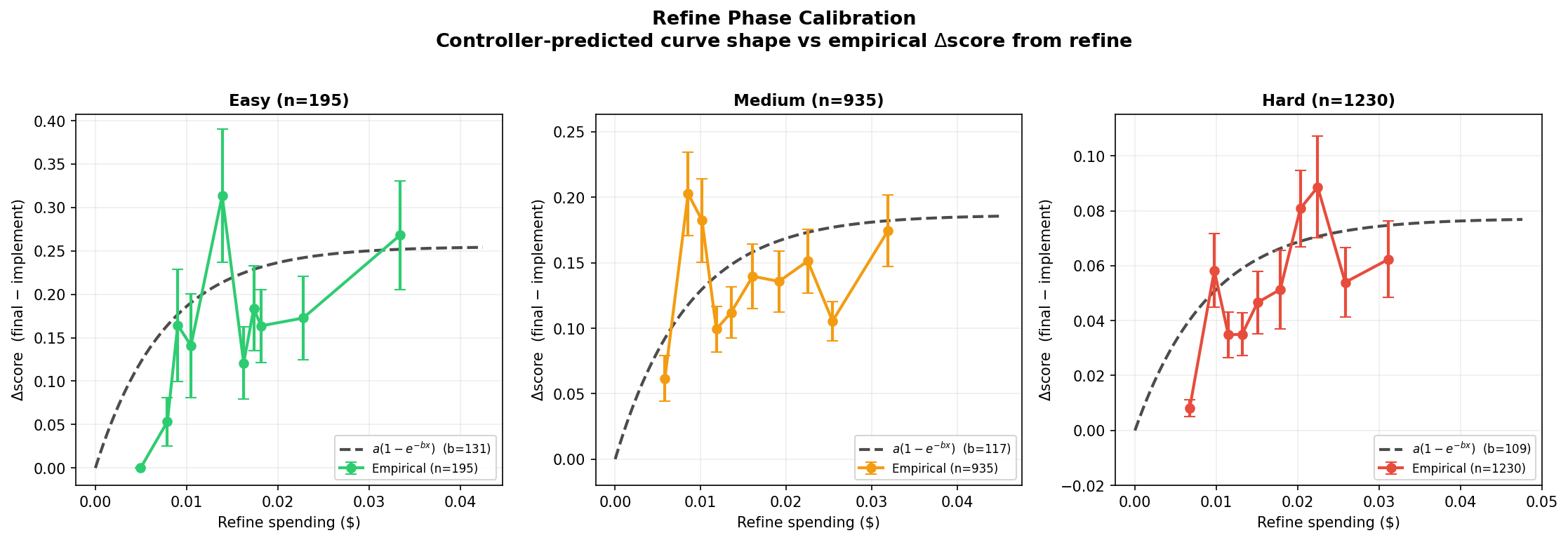}
  \caption{\textbf{Refine-phase calibration: controller-predicted
  curve shape vs.\ empirical $\Delta$score from refine.} Empirical
  points are mean $\Delta\text{score} = $ final $-$ implement within
  ten equal-count quantile bins of realized refine spend, with
  standard-error bars; runs are capped at $30$ per task before
  binning. The dashed curve is the controller's predicted shape
  $a(1 - e^{-bx})$ at the tier-average cost-adjusted $b$, normalized
  so its plateau matches the empirical plateau (top-three-bin mean).}
  \label{fig:refine_calibration}
\end{figure}

The ZEBRA solver consumes per-phase utility curves
$u_i(x) = a_i (1 - e^{-b_i x})$ produced by the controller. The
ablations in Appendix~\ref{app:ablations} show that removing those
curves (uniform split) is the single most damaging ingredient to
remove, but they do not directly check whether the controller's
\emph{shape} prediction matches the empirical phase response. This
appendix reports a post-hoc calibration of the controller's curve
for the \texttt{refine} phase against the empirical $\Delta$score it
produces in our run logs (Figure~\ref{fig:refine_calibration}). We
focus on \texttt{refine} because, unlike the upstream phases whose
intermediate outputs are not independently scored, \texttt{refine}'s
contribution can be cleanly isolated from the difference between the
implement-stage candidate and the final selected candidate.

\paragraph{Data extraction.} For each run log we parse, via regex,
the controller-printed line
\texttt{Controller curves (...): \{...\}} to recover $(a_i, b_i)$ for
each phase. The \texttt{refine} curve is fit by the controller in
token space; we replace its $b_{\texttt{refine}}$ with the
cost-adjusted value reported by the
\texttt{Refine cost adjustment:...adjusted b=X} line, which divides
the token-space $b$ by $\sim$$16.7\times$ (the
\texttt{gpt-4o-mini}~$\to$~\texttt{gpt-4o} cost ratio) so that the
curve is expressed in dollars rather than tokens. We then read three
quantities from the same log: the implement-stage score
(\texttt{[refine] candidate 0 (implement): score=X}), the final
selected score
(\texttt{[refine] Selected candidate N \dots\ with score=X}; if the
log records \texttt{Final code = implement output} the final score
equals the implement score), and the realized refine spend
(\texttt{[refine] Spent=\$X}). The empirical refine contribution is
$\Delta\text{score} = \text{final} - \text{implement}$.

\paragraph{Difficulty assignment and balancing.}
Tasks are tier-labelled from \texttt{nb\_costs.json} (the no-budget
screening means used in Appendix~\ref{app:benchmark_construction}):
easy ($\bar{s} \ge 0.8$), medium ($0.4 \le \bar{s} < 0.8$),
hard ($\bar{s} < 0.4$). Because some tasks have many more runs than
others (up to $90$ vs.\ as few as $1$), we cap each task at $30$ runs
by random sampling (seed $=42$) before pooling, so that no single
task dominates a tier.

\paragraph{Binning and curve overlay.} Within each tier, runs are
sorted by realized refine spend and split into ten equal-count
quantile bins. We plot each bin's mean $\Delta$score with its
standard error (empirical points). The controller's predicted shape
is overlaid as $a(1 - e^{-bx})$ using the tier-average $a$ and
cost-adjusted $b$, with $a$ rescaled so that the curve's plateau
matches the empirical plateau (mean of the top three bins). This
rescaling absorbs any tier-level offset in $a$ and tests only the
\emph{shape} (saturation rate) determined by $b$.

\paragraph{Result.} Across all three tiers the empirical points lie
close to the controller's predicted curve, with mild over-prediction
in the lowest-spend bins and good agreement past
$\sim$\$$0.015$ refine spend. The fitted saturation rates
($b \approx 109$--$131$ across tiers) reproduce the empirical knee
location, supporting the use of the estimated curves as inputs to
the knapsack solver in Section~\ref{sec:method}. \chg{This calibration is one of several checks on the curves; Appendix~\ref{app:curve_validation} collects them and states what they jointly establish.
}

\subsection{Per-task scatter view and binding-budget intuition}
\label{app:per_task_scatter}

\begin{figure}[t]
  \centering
  \includegraphics[width=0.95\columnwidth]{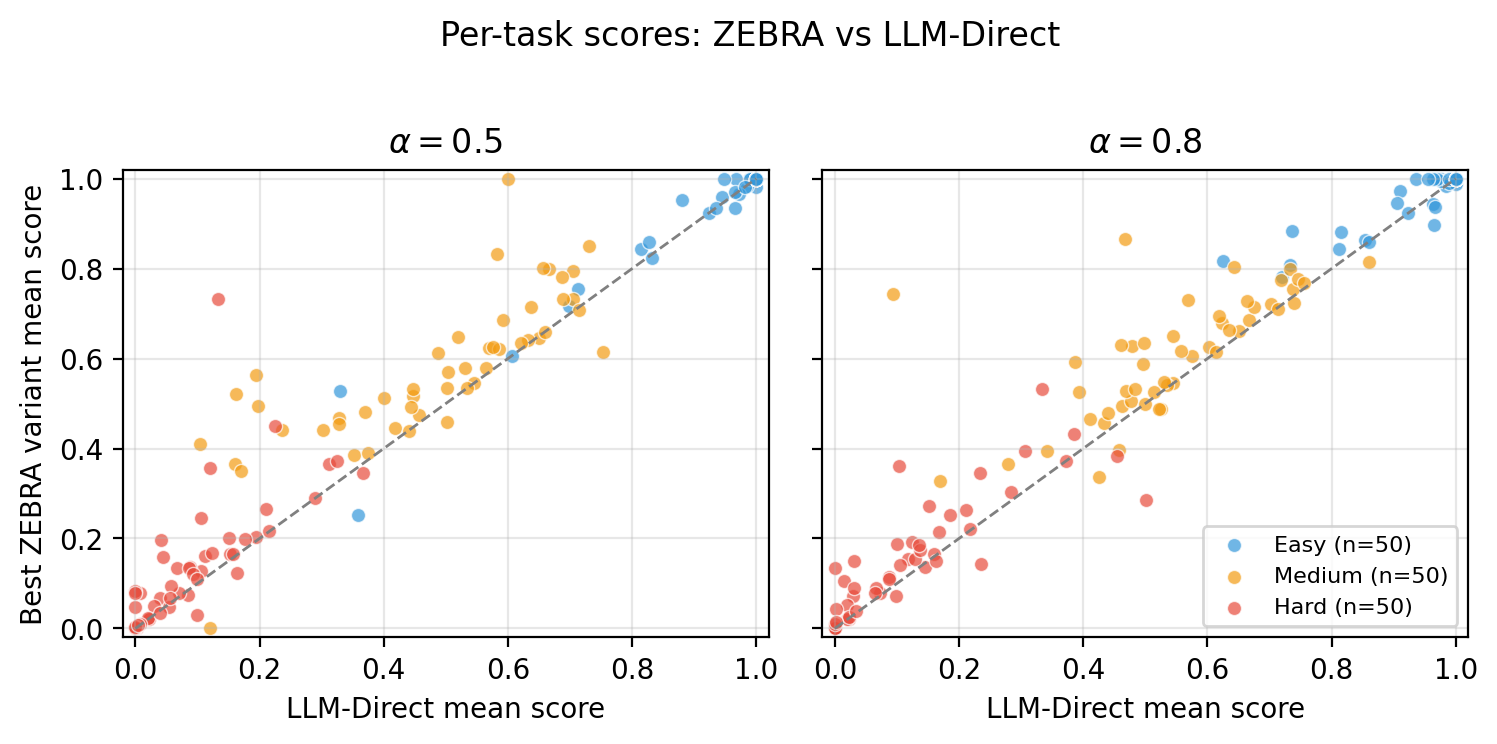}
  \caption{\textbf{Per-task scores: best ZEBRA variant vs LLM-direct.}
  Each dot is a task; $y$-axis is the best ZEBRA variant's per-task
  mean score, $x$-axis is LLM-direct's. Dashed line is $y = x$. Points
  above the diagonal are tasks where ZEBRA wins. Easy tasks (blue)
  cluster near the diagonal in the upper right; the per-task gap grows
  as tasks get harder, with a visible swarm of red points well above
  the diagonal at low LLM scores -- those are tasks where LLM-direct
  scores near zero and ZEBRA recovers some signal.}
  \label{fig:per_task_scatter}
\end{figure}

The per-tier structure in Section~\ref{sec:results_by_tier} is
consistent with the intuition behind ZEBRA: allocation quality only
matters when the budget is binding. When the task is easy, any
reasonable split works; when the task is hard or the budget is
tight, the choice of split becomes the main lever.
Figure~\ref{fig:per_task_scatter} visualises this per-task: each
point is a task, plotted as the best ZEBRA variant's mean score
(vertical) versus LLM-direct's mean score (horizontal), with the
diagonal indicating parity. Easy tasks cluster on the diagonal,
while medium and especially hard tasks drift above it -- the drift
away from the diagonal grows with task difficulty.

\subsection{Detailed budget-pressure results}
\label{app:budget_pressure}

\begin{figure}[h]
  \centering
  \includegraphics[width=0.85\columnwidth]{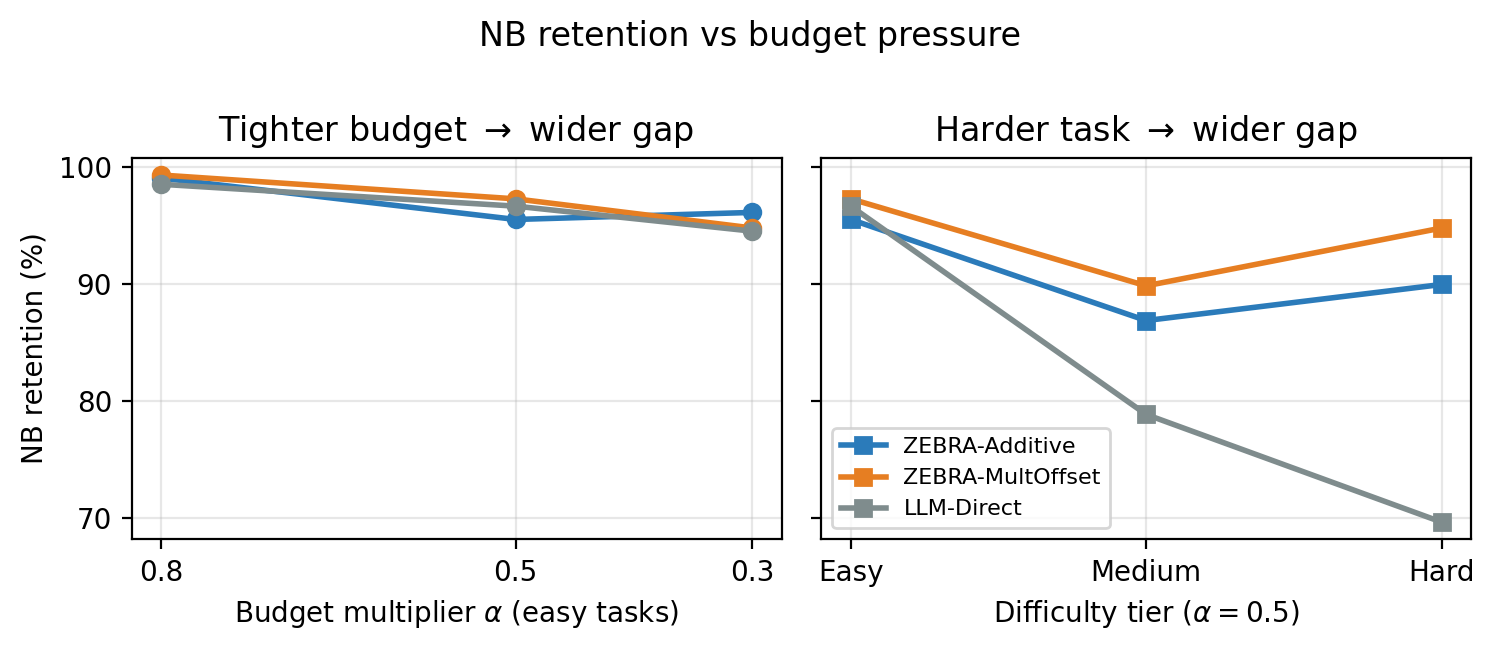}
  \caption{\textbf{NB retention vs budget tightness.} Fraction of
  unconstrained quality recovered by each strategy as the budget
  multiplier $\alpha$ tightens. The ZEBRA-vs-LLM gap is essentially
  zero at $\alpha = 0.8$ on easy tasks and grows as either the budget
  tightens (easy: $\alpha = 0.5 \to 0.3$) or the tier hardens
  ($\alpha = 0.5$, easy $\to$ medium $\to$ hard).}
  \label{fig:pressure_retention}
\end{figure}

The very-tight ($\alpha = 0.3$) regime on the $50$ easy tasks
(Section~\ref{sec:results_pressure}) is reported in full in
Table~\ref{tab:easy_alpha03}. \texttt{zebra-additive} recovers
$96.1\%$ of the no-budget score versus $94.5\%$ for LLM-direct,
beating LLM-direct on per-task mean in $19/50$ tasks (loses $8$,
ties $23$; paired Wilcoxon $p_{\text{raw}} = 0.0197$,
$p_{\text{BH}} = 0.030$) and winning outright on $38/50$ tasks.
Figure~\ref{fig:pressure_retention} visualises NB-retention vs.\
budget tightness across all conditions.

\begin{table}[t]
\centering
\small
\setlength{\tabcolsep}{5pt}
\renewcommand{\arraystretch}{1.15}
\begin{tabularx}{\columnwidth}{@{} l
  >{\centering\arraybackslash}X
  >{\centering\arraybackslash}X
  >{\centering\arraybackslash}X
  >{\centering\arraybackslash}X @{}}
\toprule
\textbf{Strategy} &
\textbf{Mean} &
\textbf{Succ\%} &
\textbf{NB ret.} &
\textbf{vs LLM (W/L/T)} \\
\midrule
\texttt{zebra-additive}     & $\mathbf{0.927}^{*}$ & $\mathbf{94.7}$ & $\mathbf{96.1}$ & $\mathbf{19/8/23}$ \\
\texttt{zebra-mult\_offset} & $0.915$              & $92.9$          & $94.8$          & $11/15/24$         \\
\texttt{llm}                & $0.912$              & $93.1$          & $94.5$          & ---                \\
\bottomrule
\end{tabularx}
\caption{\textbf{Very-tight budget on easy tasks ($\alpha = 0.3$,
$n = 50$).} \texttt{zebra-additive} beats LLM-direct on per-task mean
score (paired Wilcoxon $p_{\text{raw}} = 0.0197$, $p_{\text{BH}} =
0.0296$, marked $^{*}$).
\texttt{zebra-mult\_offset} essentially ties LLM-direct
($p_{\text{raw}} = 0.8291$, ns), with W/L/T = $11$/$15$/$24$ on
per-task mean. Bold = best in column. Asterisks reflect BH-adjusted
$p$-values from the $18$-test APPS main family
(Appendix~\ref{app:wilcoxon}).}
\label{tab:easy_alpha03}
\end{table}

\paragraph{Per-phase allocation at $\alpha = 0.3$.}
Table~\ref{tab:alloc_alpha03} reports the mean per-phase budget
fractions for the very-tight regime, supplementing
Table~\ref{tab:allocation_distribution} (which covers $\alpha = 0.5$
and $\alpha = 0.8$). Both ZEBRA variants compress \texttt{refine}
further at $\alpha = 0.3$ ($10.7\%$ for \texttt{additive},
$19.3\%$ for \texttt{mult\_offset}) than at $\alpha = 0.5$ on the
same easy tier ($30.5\%$, $45.1\%$); LLM-direct's \texttt{refine}
share is essentially unchanged across budgets ($47.3\%$ at
$\alpha = 0.3$ vs.\ $45.8\%$ at $\alpha = 0.5$).

\begin{table}[h]
\centering
\small
\setlength{\tabcolsep}{8pt}
\renewcommand{\arraystretch}{1.1}
\begin{tabular}{@{} l c c c c @{}}
\toprule
\textbf{Strategy} &
\texttt{plan} & \texttt{decomp} & \texttt{impl} & \texttt{refine} \\
\midrule
zebra-additive     & $22.4$ & $26.4$ & $40.4$ & $\mathbf{10.7}$ \\
zebra-mult\_offset & $20.3$ & $23.7$ & $36.8$ & $\mathbf{19.3}$ \\
llm                & $12.3$ & $14.2$ & $26.2$ & $\mathbf{47.3}$ \\
\bottomrule
\end{tabular}

\caption{\textbf{Per-phase budget fractions at $\alpha = 0.3$ (very-tight,
easy tasks, $n = 50$, $15$ runs per cell).} Mean allocation (\%)
across the four pipeline phases. \textbf{Both ZEBRA variants compress
\texttt{refine} further as the budget tightens} -- to $10.7\%$
(\texttt{additive}) and $19.3\%$ (\texttt{mult\_offset}) at
$\alpha = 0.3$, down from their easy-tier $\alpha = 0.5$ shares of
$30.5\%$ and $45.1\%$ respectively
(Table~\ref{tab:allocation_distribution}). LLM-direct keeps
\texttt{refine} near $47\%$, essentially unchanged from its
$\alpha = 0.5$ easy-tier share of $45.8\%$. This empirically
substantiates the behavioral claim in
Section~\ref{sec:results_allocations}.}
\label{tab:alloc_alpha03}
\end{table}

\subsection{Per-phase allocation breakdown}
\label{app:allocation_distribution}

\begin{figure}[t]
  \centering
  \includegraphics[width=0.85\columnwidth]{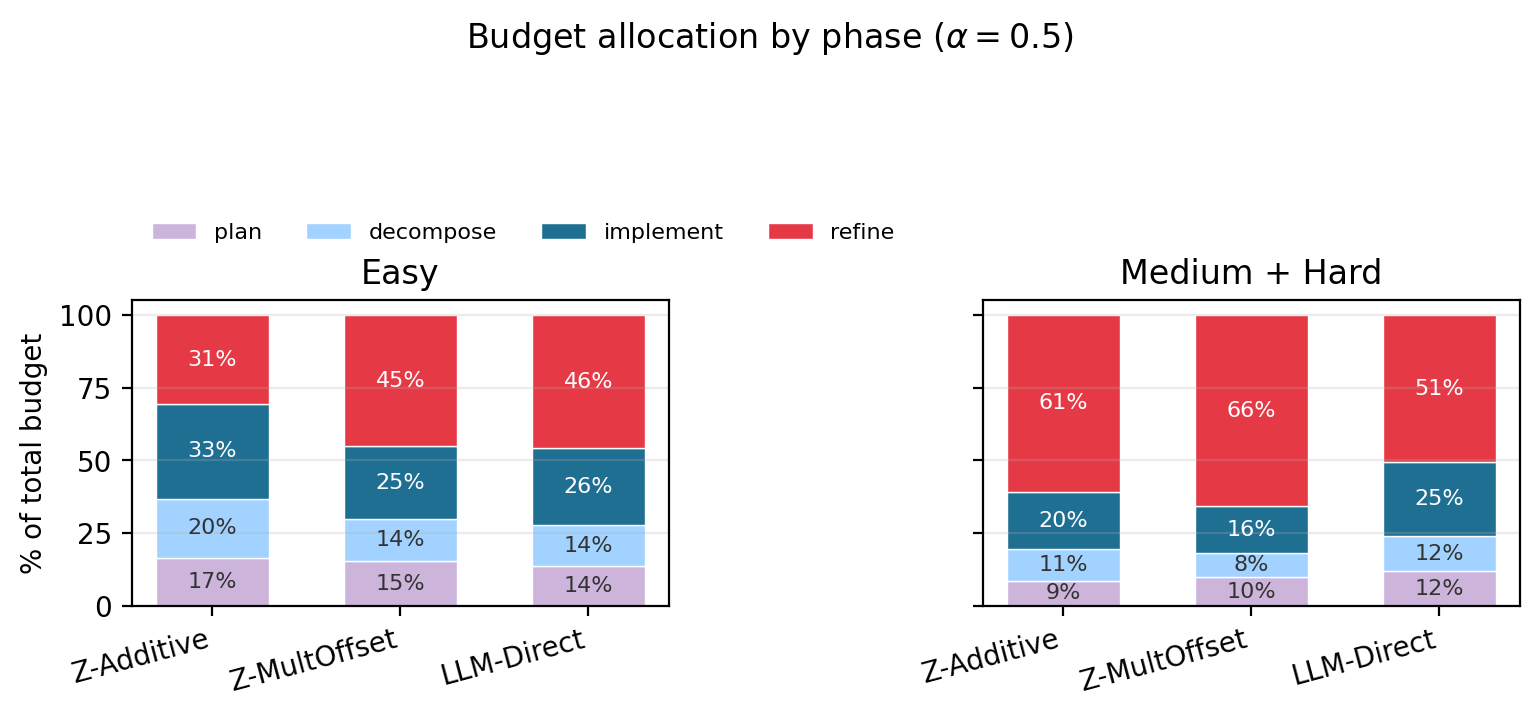}
  \caption{\textbf{Allocation distributions at $\alpha = 0.5$.} Stacked
  bars show the mean fraction of the total budget spent on each phase,
  per strategy, on easy (left) versus medium+hard (right) tasks. ZEBRA
  shifts spend from \texttt{implement} (easy) to \texttt{refine}
  (medium+hard); LLM-direct uses a near-identical split in both
  regimes.}
  \label{fig:alloc_bars}
\end{figure}

Table~\ref{tab:allocation_distribution} reports the average
per-phase budget fractions referenced in
Section~\ref{sec:results_allocations}, broken down by tier and budget
multiplier; Figure~\ref{fig:alloc_bars} visualises the same
distributions.

\begin{table*}[t]
\centering
\small
\setlength{\tabcolsep}{4pt}
\renewcommand{\arraystretch}{1.1}
\begin{tabularx}{\textwidth}{@{} l l
  >{\centering\arraybackslash}X
  >{\centering\arraybackslash}X
  >{\centering\arraybackslash}X
  >{\centering\arraybackslash}X
  >{\centering\arraybackslash}X
  >{\centering\arraybackslash}X
  >{\centering\arraybackslash}X
  >{\centering\arraybackslash}X @{}}
\toprule
\textbf{Tier} & \textbf{Strategy} &
\multicolumn{4}{c}{\textbf{Mean allocation (\%), $\boldsymbol{\alpha = 0.5}$}} &
\multicolumn{4}{c}{\textbf{Mean allocation (\%), $\boldsymbol{\alpha = 0.8}$}} \\
\cmidrule(lr){3-6} \cmidrule(lr){7-10}
 & & \texttt{plan} & \texttt{decomp} & \texttt{impl} & \texttt{refine}
   & \texttt{plan} & \texttt{decomp} & \texttt{impl} & \texttt{refine} \\
\midrule

\multirow{3}{*}{\textbf{Easy}}
& zebra-additive     & $16.6$ & $20.3$ & $32.6$ & $30.5$ & $11.6$ & $14.3$ & $23.2$ & $50.9$ \\
& zebra-mult\_offset & $15.5$ & $14.5$ & $24.9$ & $45.1$ & $12.2$ & $11.3$ & $19.8$ & $56.7$ \\
& llm                & $13.7$ & $14.2$ & $26.3$ & $45.8$ & $13.7$ & $13.2$ & $26.7$ & $46.4$ \\
\midrule

\multirow{3}{*}{\textbf{Medium + Hard}}
& zebra-additive     & $8.7$  & $10.8$ & $19.8$ & $60.7$ & $6.2$  & $7.8$  & $14.6$ & $71.3$ \\
& zebra-mult\_offset & $9.9$  & $8.3$  & $16.2$ & $65.7$ & $8.2$  & $6.7$  & $13.7$ & $71.3$ \\
& llm                & $11.9$ & $12.3$ & $25.2$ & $50.6$ & $12.3$ & $12.7$ & $24.1$ & $50.9$ \\

\bottomrule
\end{tabularx}
\caption{\textbf{Where each strategy spends its budget.} Mean
per-phase budget fractions at $\alpha = 0.5$ and $\alpha = 0.8$,
grouped by difficulty tier. \textbf{Both ZEBRA variants reallocate
across tiers}: on easy tasks they assign \texttt{refine} only $30$--$57\%$
of the budget, while on medium+hard tasks both variants pull strongly
toward \texttt{refine} ($60$--$71\%$). LLM-direct uses essentially
the same $\sim$$25\%$ \texttt{implement} $/$ $\sim$$50\%$ \texttt{refine}
split across tiers and budgets. The within-strategy shift is largest
for \texttt{additive} (a $30$-point drop in \texttt{refine} share between
tiers at $\alpha = 0.5$). This adaptation is the proximate cause of
the score gaps in Table~\ref{tab:main_by_tier}.}
\label{tab:allocation_distribution}
\end{table*}

\subsection{Per-task win counts across objectives}
\label{app:variant_wins}

Table~\ref{tab:variant_wins} accompanies the discussion of
``Additive vs.\ \texttt{mult\_offset}: when does the product
aggregator help?'' in Section~\ref{sec:results_allocations}: it reports
per-task win counts for all four strategies (\texttt{additive},
\texttt{mult\_offset}, \texttt{prop\_offset}, and LLM-direct) across
tiers and budget multipliers. The \texttt{prop\_offset} column is
included for completeness; that variant and its head-to-head with
\texttt{mult\_offset} are discussed in
Appendix~\ref{app:prop_offset}.

\begin{table}[t]
\centering
\small
\setlength{\tabcolsep}{4pt}
\renewcommand{\arraystretch}{1.1}
\begin{tabularx}{\columnwidth}{@{} l c
  >{\centering\arraybackslash}X
  >{\centering\arraybackslash}X
  >{\centering\arraybackslash}X
  >{\centering\arraybackslash}X @{}}
\toprule
$\boldsymbol{\alpha}$ & \textbf{Tier ($n$)} &
\textbf{add.} & \textbf{prop.} & \textbf{mult.} & \textbf{llm} \\
\midrule
$0.3$ & Easy ($50$)        & $\mathbf{38}$ & $5$  & $3$  & $4$  \\
\midrule
\multirow{4}{*}{$0.5$}
      & Easy ($50$)        & $\mathbf{32}$ & $7$  & $2$  & $9$  \\
      & Medium ($50$)      & $\mathbf{18}$ & $13$ & $13$ & $6$  \\
      & Hard ($50$)        & $14$          & $9$  & $\mathbf{21}$ & $6$  \\
      & \emph{All} ($150$) & $\mathbf{64}$ & $29$ & $36$ & $21$ \\
\midrule
\multirow{4}{*}{$0.8$}
      & Easy ($50$)        & $\mathbf{33}$ & $4$  & $7$  & $6$  \\
      & Medium ($50$)      & $\mathbf{18}$ & $16$ & $10$ & $6$  \\
      & Hard ($50$)        & $\mathbf{21}$ & $9$  & $13$ & $7$  \\
      & \emph{All} ($150$) & $\mathbf{72}$ & $29$ & $30$ & $19$ \\
\bottomrule
\end{tabularx}
\caption{\textbf{Per-task win counts.} Number of tasks (out of the
indicated $n$) on which each strategy achieves the highest per-task
mean score among the four reported strategies. \texttt{add.}\ =
\texttt{zebra-additive}, \texttt{prop.}\ = \texttt{zebra-prop\_offset},
\texttt{mult.}\ = \texttt{zebra-mult\_offset}. The $\alpha = 0.3$ row
is restricted to the easy tasks
(Section~\ref{sec:results_pressure}); $\alpha \in \{0.5, 0.8\}$ rows
are from the full $150$-task run. \texttt{zebra-additive} is the
per-task win-count leader at every aggregate budget, and
\texttt{zebra-mult\_offset} takes over at the hard tier under
$\alpha = 0.5$. LLM-direct wins only $13$--$14\%$ of tasks at
$\alpha \ge 0.5$, falling to $8\%$ at $\alpha = 0.3$. Bold marks the
leader per row.}
\label{tab:variant_wins}
\end{table}

\subsection{Propagation-weighted variant: definition and comparison to \texttt{mult\_offset}}
\label{app:prop_offset}

The body of the paper compares two ZEBRA objectives, \texttt{additive}
(Eq.~\ref{eq:zebra_additive}) and \texttt{mult\_offset}
(Eq.~\ref{eq:mult_offset}). Here we define a third, propagation-weighted
variant -- which adds per-phase dependency weights on top of
\texttt{mult\_offset} -- and report its empirical comparison against
\texttt{mult\_offset}. The headline finding: the additional per-phase
weighting does \emph{not} reliably move the per-task score over the
symmetric \texttt{mult\_offset}, so we leave it out of the body and
report it here for completeness.

\paragraph{Definition.}
The multiplicative objective in Eq.~\eqref{eq:mult_offset} treats all
phases symmetrically. To allow per-phase emphasis -- motivated by the
intuition that early phases (e.g., \texttt{plan}) cascade more strongly
into downstream phases than late phases -- we introduce dependency
weights $w_i > 0$ and maximize:
\begin{equation}
\max_{\mathbf{x}} \quad \prod_{i=1}^n \big(1 - a_i\, e^{-b_i x_i}\big)^{w_i} \quad \text{s.t.}\ \sum_i x_i \le B,\ x_i \ge 0.
\label{eq:prop_offset}
\end{equation}
Via the log transformation (Appendix~\ref{app:log_transform}), this
becomes $\max \sum_i w_i \log(1 - a_i e^{-b_i x_i})$ and the per-phase
allocation is:
\begin{equation}
x_i(\lambda) = \max\!\bigg(0,\, \frac{1}{b_i}\ln\frac{a_i(w_i b_i + \lambda)}{\lambda}\bigg).
\end{equation}
Phase $i$ receives zero budget when $\lambda \ge a_i w_i b_i / (1 - a_i)$.
The weight $w_i$ is set to the quality ceiling $a_i$ already estimated
by the controller. Phases with higher ceilings  which contribute
more to the product when they succeed  therefore receive
proportionally more weight, while \texttt{prop\_offset} reduces to
\texttt{mult\_offset} when $w_i \equiv 1$ (uniform weighting).

\paragraph{Empirical comparison: \texttt{prop\_offset} vs.\ \texttt{mult\_offset}.}
We run \texttt{prop\_offset} on the same $150$-task benchmark with the
same $15$ runs per cell. Table~\ref{tab:prop_vs_mult} reports the
side-by-side comparison and a paired Wilcoxon signed-rank test on
per-task mean score against \texttt{mult\_offset} (the natural
$w_i \equiv 1$ control). The picture is consistent across both budget
levels: differences are small, do not have a consistent
sign across cells, and are not statistically significant at the
overall level.

\begin{table}[t]
\centering
\small
\setlength{\tabcolsep}{4pt}
\renewcommand{\arraystretch}{1.1}
\begin{tabularx}{\columnwidth}{@{} l l
  >{\centering\arraybackslash}X
  >{\centering\arraybackslash}X
  >{\centering\arraybackslash}X
  >{\centering\arraybackslash}X @{}}
\toprule
$\boldsymbol{\alpha}$ & \textbf{Tier ($n$)} &
\textbf{prop mean} & \textbf{mult mean} &
\textbf{$\Delta$ (prop$-$mult)} & \textbf{$p$ vs mult} \\
\midrule

\multirow{5}{*}{$0.5$}
& Easy ($50$)        & $0.9347$ & $0.9385$ & $-0.0039$ & $0.28$~\textsuperscript{ns} \\
& Medium ($50$)      & $0.5473$ & $0.5506$ & $-0.0033$ & $0.91$~\textsuperscript{ns} \\
& Hard ($50$)        & $0.1175$ & $0.1300$ & $-0.0125$ & $0.10$~\textsuperscript{ns} \\
& Med+Hard ($100$)   & $0.3324$ & $0.3403$ & $-0.0079$ & $0.33$~\textsuperscript{ns} \\
& \emph{All} ($150$) & $0.5331$ & $0.5397$ & $-0.0066$ & $0.15$~\textsuperscript{ns} \\
\midrule

\multirow{5}{*}{$0.8$}
& Easy ($50$)        & $0.9563$ & $0.9582$ & $-0.0019$ & $0.59$~\textsuperscript{ns} \\
& Medium ($50$)      & $0.5945$ & $0.5772$ & $+0.0173$ & $0.083$~\textsuperscript{ns} \\
& Hard ($50$)        & $0.1365$ & $0.1376$ & $-0.0011$ & $0.82$~\textsuperscript{ns} \\
& Med+Hard ($100$)   & $0.3655$ & $0.3574$ & $+0.0081$ & $0.22$~\textsuperscript{ns} \\
& \emph{All} ($150$) & $0.5624$ & $0.5576$ & $+0.0048$ & $0.40$~\textsuperscript{ns} \\

\bottomrule
\end{tabularx}
\caption{\textbf{\texttt{prop\_offset} vs.\ \texttt{mult\_offset}, head-to-head
on $150$ tasks ($15$ runs per cell).} Mean per-task score, the difference
$\Delta = \text{prop} - \text{mult}$, and the two-sided paired Wilcoxon
signed-rank $p$-value on per-task mean score against \texttt{mult\_offset}.
The displayed $p$ is raw; significance markers reflect Benjamini--Hochberg
adjustment over the $10$ tests in this family
($^{*}p_{\text{BH}}<0.05$, $^{**}p_{\text{BH}}<0.01$,
$^{***}p_{\text{BH}}<0.001$, ns~=~not significant).
\textbf{No cell is significant at $p_{\text{raw}} < 0.05$, and
therefore none is significant under BH.}
The sign of $\Delta$ flips between budget regimes
($\Delta < 0$ at $\alpha = 0.5$ overall, $\Delta > 0$ at $\alpha = 0.8$
overall). Win counts on the all-tasks rows
are prop $50$ / mult $62$ / ties $38$ at $\alpha = 0.5$ and prop $57$
/ mult $57$ / ties $36$ at $\alpha = 0.8$. Conclusion: the LLM-elicited
$\alpha_i$ weights do not move the per-task score reliably away from
the symmetric \texttt{mult\_offset} ($w_i \equiv 1$) baseline.}
\label{tab:prop_vs_mult}
\end{table}

\paragraph{Allocations are nearly identical.}
The mean per-phase budget shares of \texttt{prop\_offset} and
\texttt{mult\_offset} differ by at most $\sim$$2$ percentage points on
any phase at either budget level (e.g., at $\alpha = 0.5$:
\texttt{plan} $12.2\%$ vs.\ $11.7\%$; \texttt{decompose} $10.5\%$ vs.\
$10.3\%$; \texttt{implement} $20.7\%$ vs.\ $19.1\%$; \texttt{refine}
$56.6\%$ vs.\ $58.8\%$). Reusing $a_i$ as the weight, \texttt{prop\_offset}
does not pull the allocation meaningfully away from the symmetric
\texttt{mult\_offset} solution.

\paragraph{Why the per-phase weights do not help.} A few plausible
reasons, most of which point to the operationalization of the weights
rather than to the dependency-modeling idea itself:
\begin{itemize}
\itemsep=2pt
\item \emph{The weights are redundant to existing information.}
Setting $w_i = a_i$ reuses the same per-phase quality ceiling that
\texttt{mult\_offset} already incorporates through the offset curve
$1 - a_i e^{-b_i x_i}$. The additional $a_i$ exponent reweights the
product but adds no new signal, so allocations end up tracking the
symmetric $w_i \equiv 1$ solution.
\item \emph{On hard tasks, every phase matters.} In the regime where
\texttt{mult\_offset} most clearly dominates \texttt{additive} (hard
tier at $\alpha = 0.5$), tilting away from balance costs more than it
saves -- and we observe \texttt{prop\_offset} losing to
\texttt{mult\_offset} on hard tasks at $\alpha = 0.5$ ($0.1175$
vs.\ $0.1300$).
\end{itemize}

We expect dependency-weighting could help in settings where the weight
carries information beyond $a_i$  e.g., explicit per-phase weights
elicited from a separate prompt, or learned from observed cascades 
but reusing $a_i$ as the weight does not move the allocation
meaningfully in the $4$-phase coding pipeline we evaluate.

\subsection{Does this transfer? HumanEval and CodeContests}
\label{app:transfer}

To check whether ZEBRA's APPS gains generalize to other coding
benchmarks we run two transfer datasets, each at the tight budget
$\alpha = 0.5$ only:

\begin{itemize}\itemsep=2pt
\item \textbf{HumanEval}~\citep{chen2021codex}. We extract $50$
candidate tasks via \texttt{extract\_humaneval\_tasks.py}, screen them
with $15$ no-budget runs as in Appendix~\ref{app:benchmark_construction},
and retain $29$ tasks (after dropping trivially-solved cases and tasks
without sufficient cost variance). The retained pool is dominated by
the easy tier ($28$ easy, $1$ medium, $0$ hard) -- a regime where the
APPS main results predict that ZEBRA's advantage should be small.
\item \textbf{CodeContests}~\citep{li2022alphacode}. We extract $50$
candidate tasks via \texttt{extract\_codecontests\_tasks.py},
screen identically, and retain $23$ tasks. The retained pool is
balanced toward the harder end ($8$ easy, $5$ medium, $10$ hard) and is
the closest analogue to APPS' tier mix among the transfer datasets we
tested.
\end{itemize}

Both runs use the same pipeline, models and budget definition as the
APPS experiments; only the task pool changes.

\paragraph{HumanEval ($n = 29$, $\alpha = 0.5$).}

\begin{table}[h]
\centering
\small
\begin{tabular}{l c c c c}
\toprule
Strategy & Mean & NB ret. & vs LLM (W/L/T) & $p_{\text{raw}}$ \\
\midrule
\texttt{zebra-additive}     & $0.9626$ & $98.6\%$ & $9$/$2$/$18$  & $0.0754$ \textsuperscript{ns} \\
\texttt{zebra-mult\_offset} & $\mathbf{0.9672}$ & $\mathbf{99.1\%}$ & $\mathbf{11}$/$1$/$17$ & $0.0253$\textsuperscript{*} \\
\texttt{llm\_cot}           & $0.9463$ & $97.0\%$ & $8$/$6$/$15$  & $0.8752$ \textsuperscript{ns} \\
\texttt{llm}                & $0.9419$ & $96.5\%$ & ---           & --- \\
\bottomrule
\end{tabular}
\caption{\textbf{HumanEval transfer at $\alpha = 0.5$
($n = 29$ tasks, $15$ runs per cell).} Per-task mean F1 score, NB
retention, and paired Wilcoxon signed-rank test vs.\ LLM-direct.
\texttt{zebra-mult\_offset} reaches significance at the raw level;
\texttt{zebra-additive} is in the same direction but underpowered.
$28$ of $29$ retained tasks fall in the easy tier, where the APPS
main results predict the smallest advantage.}
\label{tab:humaneval_overall}
\end{table}

Both ZEBRA variants finish above LLM-direct on per-task mean and on
NB-retention; \texttt{zebra-mult\_offset} reaches significance at the
raw level ($p = 0.025$). The headline gaps are small in absolute
terms ($+2$--$3$ points) because $28$ of $29$ retained tasks are easy
and the budget is rarely binding -- exactly the regime where the APPS
main results show the smallest advantage
(Table~\ref{tab:main_by_tier}, easy column). LLM-CoT does not close
the gap: it scores $+0.004$ over plain LLM-direct, n.s.

\paragraph{CodeContests ($n = 23$, $\alpha = 0.5$).}

\begin{table}[h]
\centering
\small
\begin{tabular}{l c c c c}
\toprule
Strategy & Mean & NB ret. & vs LLM (W/L/T) & $p_{\text{raw}}$ \\
\midrule
\texttt{zebra-additive}     & $\mathbf{0.3824}$ & $\mathbf{80.3\%}$ & $\mathbf{12}$/$7$/$4$ & $0.1712$ \textsuperscript{ns} \\
\texttt{zebra-mult\_offset} & $0.3681$ & $77.3\%$ & $10$/$8$/$5$  & $0.3958$ \textsuperscript{ns} \\
\texttt{llm}                & $0.3534$ & $74.2\%$ & ---           & --- \\
\bottomrule
\end{tabular}
\caption{\textbf{CodeContests transfer at $\alpha = 0.5$
($n = 23$ tasks, $15$ runs per cell).} Per-task mean F1 score, NB
retention, and paired Wilcoxon signed-rank test vs.\ LLM-direct.
Both ZEBRA variants finish above LLM-direct on mean and
NB-retention with per-task wins favouring ZEBRA, but neither cell
reaches significance at this sample size; we treat CodeContests as
an underpowered generalization probe rather than a primary result.}
\label{tab:codecontests_overall}
\end{table}

Both ZEBRA variants finish above LLM-direct in mean and NB-retention,
with the per-task wins favouring ZEBRA ($12$/$7$/$4$ for additive),
but neither cell reaches significance at $n = 23$ -- the screened pool
is small. The signal is in the right direction but underpowered. We
treat CodeContests as a generalization probe rather than as a primary
result.

\paragraph{What transfers and what does not.} The patterns from the
APPS main results carry over: (i) ZEBRA's advantage is largest where
budget is binding and the task is non-trivial -- on HumanEval, where
$28/29$ tasks are easy and budget is loose, the gap is small but in
the right direction; (ii) curve-aware allocation does not hurt in the
easy regime even with mult's product objective, contradicting the
concern that the multiplicative formulation might over-fund refine on
trivially-solved tasks; (iii) chain-of-thought prompting on the
LLM-direct controller does not recover the gap on either dataset, just
as on APPS. We do not claim CodeContests is statistically
distinguishable from LLM-direct; we report it because the screened
pool was tier-mixed and the direction of the effect is consistent
with the APPS results.

\subsection{\chg{Candidate selection: shipping the last valid candidate}}
\label{app:last_valid}

\chg{Scores reported elsewhere in the paper take the best candidate a run
produces. We
therefore re-score every run under }\texttt{\chg{last\_valid}} \chg{selection: the score
is that of the last candidate the pipeline ships -- on APPS the most
recent valid }\texttt{\chg{refine}} \chg{candidate, on HotpotQA the last non-empty
candidate of the }\texttt{\chg{direct}} \chg{$\to$ }\texttt{\chg{research}} \chg{$\to$
}\texttt{\chg{synthesize}} \chg{chain -- with no post-hoc choice among candidates. The runs, models, budgets, and task pools are
unchanged; only the scoring rule differs.
}

\paragraph{\chg{APPS.}} \chg{The same $150$-task pool ($50$ easy / $50$ medium /
$50$ hard), $15$ runs per (task, strategy) cell, at both budget levels.
Results are in Table~\ref{tab:last_valid_apps}.
}

\begin{table}[h]
\centering
\small
\setlength{\tabcolsep}{4pt}
\begin{tabular}{l l c c c c c c}
\toprule
\chg{$\boldsymbol{\alpha}$ }& \textbf{\chg{Strategy}} & \textbf{\chg{Easy}} & \textbf{\chg{Medium}}
& \textbf{\chg{Hard}} & \textbf{\chg{M$+$H}} & \textbf{\chg{All}} & \chg{$\boldsymbol{\Delta}$ }\textbf{\chg{vs LLM}} \\
\midrule
\multirow{3}{*}{$0.5$}
& \texttt{\chg{zebra-additive}}     & \chg{$0.9068$          }& \chg{$\mathbf{0.4730}$ }& \chg{$\mathbf{0.0854}$ }& \chg{$\mathbf{0.2792}$ }& \chg{$\mathbf{0.4884}$ }& \chg{$+0.0300$\textsuperscript{**}  }\\
& \texttt{\chg{zebra-mult\_offset}} & \chg{$\mathbf{0.9122}$ }& \chg{$0.4532$          }& \chg{$0.0853$          }& \chg{$0.2693$          }& \chg{$0.4836$          }& \chg{$+0.0252$\textsuperscript{***} }\\
& \chg{LLM-direct                  }& \chg{$0.9018$          }& \chg{$0.3970$          }& \chg{$0.0764$          }& \chg{$0.2367$          }& \chg{$0.4584$          }& \chg{---                            }\\
\midrule
\multirow{3}{*}{$0.8$}
& \texttt{\chg{zebra-additive}}     & \chg{$0.9033$          }& \chg{$\mathbf{0.4882}$ }& \chg{$0.0866$          }& \chg{$0.2874$          }& \chg{$0.4927$          }& \chg{$+0.0277$\textsuperscript{**}  }\\
& \texttt{\chg{zebra-mult\_offset}} & \chg{$\mathbf{0.9199}$ }& \chg{$0.4874$          }& \chg{$\mathbf{0.0930}$ }& \chg{$\mathbf{0.2902}$ }& \chg{$\mathbf{0.5001}$ }& \chg{$+0.0351$\textsuperscript{***} }\\
& \chg{LLM-direct                  }& \chg{$0.9181$          }& \chg{$0.4057$          }& \chg{$0.0712$          }& \chg{$0.2385$          }& \chg{$0.4650$          }& \chg{---                            }\\
\bottomrule
\end{tabular}
\caption{\textbf{\chg{APPS under }\texttt{\chg{last\_valid}} \chg{selection}}
\chg{($150$ tasks, $15$ runs per cell). Mean score per tier; M$+$H is the
$100$ medium and hard tasks. $\Delta$ is against LLM-direct on all $150$
tasks, with paired Wilcoxon signed-rank, two-sided, on per-task means.
Bold marks the better ZEBRA variant per
column within each $\alpha$ block.}}
\label{tab:last_valid_apps}
\end{table}

\paragraph{\chg{HotpotQA.}} \chg{The same $114$-task set at $\alpha = 0.5$,
$15$ runs per cell.
Results are in Table~\ref{tab:last_valid_hotpotqa}.
}

\begin{table}[h]
\centering
\small
\begin{tabular}{l c c c}
\toprule
\textbf{\chg{Strategy}} & \textbf{\chg{Mean F1}} & \chg{$\boldsymbol{\Delta}$ }\textbf{\chg{vs LLM}} & \chg{$\boldsymbol{p}$ }\\
\midrule
\texttt{\chg{zebra-additive}}     & \chg{$0.4378$          }& \chg{$+0.0408$          }& \chg{$0.0065$\textsuperscript{**} }\\
\texttt{\chg{zebra-mult\_offset}} & \chg{$\mathbf{0.4418}$ }& \chg{$\mathbf{+0.0448}$ }& \chg{$0.043$\textsuperscript{*}   }\\
\midrule
\chg{LLM-direct                  }& \chg{$0.3970$          }& \chg{---                }& \chg{---                          }\\
\bottomrule
\end{tabular}
\caption{\textbf{\chg{HotpotQA under }\texttt{\chg{last\_valid}} \chg{selection}}
\chg{($114$ tasks, $\alpha = 0.5$, $15$ runs per cell). Mean token-level F1.
$p$-values are paired Wilcoxon signed-rank, two-sided, on per-task mean F1
against LLM-direct.}}
\label{tab:last_valid_hotpotqa}
\end{table}

\paragraph{\chg{Reading.}} \chg{Absolute scores fall on both benchmarks, as expected
when no candidate is chosen post hoc, but the comparison is unaffected: both
ZEBRA variants beat LLM-direct in all six cells, significantly in every one,
and the margins ($+0.025$ to $+0.035$ on APPS, $+0.041$ and $+0.045$ F1 on
HotpotQA) are of the same size as under best-candidate scoring. ZEBRA's
advantage is therefore a property of the allocation, not of the selection
rule applied on top of it.
}

\subsection{\chg{End-to-end budgeting: charging the controller to the budget}}
\label{app:ctrl_budget}

\chg{Throughout the paper the budget $B$ covers phase spend, and the controller
call that produces the curves is reported separately as overhead
(App.~\ref{app:overhead}). This is the natural accounting when dealing with these very small budgets.  We
therefore re-run both benchmarks with
}\texttt{\chg{charge\_controller\_from\_budget}}\chg{, in which a reserve is deducted from
$B$ before the phases run and the phase pool is rescaled to
$B - \text{controller spend}$ once the actual cost is known. LLM-direct is
charged the same way for its own split-emitting call, so the comparison stays
paired; because ZEBRA's controller call is the more expensive of the two, this
accounting is strictly adverse to ZEBRA.
}

\paragraph{\chg{APPS ($\alpha = 0.8$).}} \chg{A $150$-task tier-balanced pool
($50$ easy / $50$ medium / $50$ hard), $15$ runs per (task, strategy) cell, with
}\texttt{\chg{gpt-4o-mini}} \chg{for }\texttt{\chg{plan}} \chg{/ }\texttt{\chg{decompose}} \chg{/
}\texttt{\chg{implement}}\chg{, }\texttt{\chg{gpt-4o}} \chg{for }\texttt{\chg{refine}}\chg{, and }\texttt{\chg{gpt-4o}}
\chg{as the controller.
}\begin{table}[h]
\centering
\small
\begin{tabular}{l c c c}
\toprule
\textbf{\chg{Strategy}} & \textbf{\chg{Mean score}} & \chg{$\boldsymbol{\Delta}$ }\textbf{\chg{vs LLM}} & \chg{$\boldsymbol{p}$ }\\
\midrule
\texttt{\chg{zebra-mult\_offset}} & \chg{$\mathbf{0.5441}$ }& \chg{$+0.0144$ }& \chg{$0.014$\textsuperscript{*} }\\
\chg{LLM-direct                  }& \chg{$0.5297$          }& \chg{---       }& \chg{---                        }\\
\bottomrule
\end{tabular}
\caption{\textbf{\chg{APPS with the controller charged to the budget}}
\chg{($150$ tasks, $\alpha = 0.8$, $15$ runs per cell). $p$ is paired Wilcoxon
signed-rank, two-sided, on per-task means.}}
\label{tab:ctrl_budget_apps}
\end{table}

\paragraph{\chg{HotpotQA ($\alpha \in \{0.5, 0.8\}$).}} \chg{The same $114$-task set,
$15$ runs per cell, phase and controller model }\texttt{\chg{gpt-4o-mini}}\chg{. The
reserves deducted upfront are the empirical median controller costs from the
main runs: \$$0.000110$ for ZEBRA and \$$0.000045$ for LLM-direct.
}

\begin{table}[h]
\centering
\small
\setlength{\tabcolsep}{5pt}
\begin{tabular}{l c c c c c c}
\toprule
& \multicolumn{3}{c}{$\alpha = 0.5$} & \multicolumn{3}{c}{$\alpha = 0.8$} \\
\cmidrule(lr){2-4} \cmidrule(lr){5-7}
\textbf{\chg{Strategy}} & \textbf{\chg{F1}} & \chg{$\boldsymbol{\Delta}$ }& \chg{$\boldsymbol{p}$
                  }& \textbf{\chg{F1}} & \chg{$\boldsymbol{\Delta}$ }& \chg{$\boldsymbol{p}$ }\\
\midrule
\texttt{\chg{zebra-additive}}     & \chg{$0.4272$          }& \chg{$+0.0032$          }& \chg{$0.66$\textsuperscript{ns} }& \chg{$\mathbf{0.4843}$ }& \chg{$\mathbf{+0.0531}$ }& \chg{$2\times10^{-4}$\textsuperscript{***} }\\
\texttt{\chg{zebra-mult\_offset}} & \chg{$\mathbf{0.4378}$ }& \chg{$\mathbf{+0.0138}$ }& \chg{$0.15$\textsuperscript{ns} }& \chg{$0.4795$          }& \chg{$+0.0483$          }& \chg{$2\times10^{-4}$\textsuperscript{***} }\\
\midrule
\chg{LLM-direct                  }& \chg{$0.4240$          }& \chg{---                }& \chg{---                        }& \chg{$0.4312$          }& \chg{---                }& \chg{---                                   }\\
\bottomrule
\end{tabular}
\caption{\textbf{\chg{HotpotQA with the controller charged to the budget}}
\chg{($114$ tasks, $15$ runs per cell). Mean token-level F1; $p$ is paired
Wilcoxon signed-rank, two-sided, on per-task mean F1 against LLM-direct.}}
\label{tab:ctrl_budget_hotpotqa}
\end{table}

\paragraph{\chg{Reading.}} \chg{At $\alpha = 0.8$ the stricter accounting costs ZEBRA
nothing decisive: it still wins on APPS ($+0.0144$, $p = 0.014$) and on
HotpotQA both variants gain roughly $5$pp F1 at $p = 2 \times 10^{-4}$.
At $\alpha = 0.5$ on HotpotQA the advantage disappears
($+0.003$ and $+0.014$, both n.s.). The reason is the ratio between the
controller call and the pool it is paid from: the controller costs
$\sim$$30\%$ of the budget at $\alpha = 0.5$ (App.~\ref{app:overhead}), so
charging it internally leaves the phases with a pool small enough that every
allocator is pushed toward the same near-minimal split, and the allocation
decision stops mattering. This is a real boundary on the method rather than a
measurement artifact: ZEBRA buys an allocation, and there has to be enough
budget left to allocate. ZEBRA should therefore be applied only when the
budget is large enough to absorb the controller call and still leave a phase
pool worth optimizing over.
}

\subsection{Sensitivity to curve-estimation noise}
\label{app:sensitivity}

We isolate the effect of curve accuracy by bypassing the live
controller and injecting curves with controlled levels of synthetic
noise. For each task we average the controller's $(a_i, b_i)$
estimates across the $15$ runs of the main benchmark
(\texttt{zebra-additive}, $\alpha = 0.5$); these averaged curves are
treated as a stable per-task ``ground-truth.'' We then perturb each
parameter with multiplicative Gaussian noise
$a_i' = a_i (1 + \mathcal{N}(0, \sigma))$,
$b_i' = b_i (1 + \mathcal{N}(0, \sigma))$, with values clipped to
$a \in [0.01, 1]$ and $b \in [0.01, \infty)$. We test
$\sigma \in \{0\%, 10\%, 50\%\}$. The pipeline, models, and budget
($\alpha = 0.5$) are held fixed; only the injected curves change.
Each $(\text{task}, \sigma)$ cell is repeated $5$ times, yielding
$150 \times 3 \times 5 = 2{,}250$ runs.

Table~\ref{tab:sensitivity} reports per-task mean scores and paired
Wilcoxon tests against the live controller and against the LLM-direct
and Uniform baselines. \textbf{The allocation is robust to estimation
error.} At $\sigma = 50\%$, where parameters can be halved or doubled
(e.g., a true $a = 0.6$ becomes anywhere in roughly $[0.3, 0.9]$),
mean score drops only from $0.5305$ to $0.5165$ -- a $1.4$-point gap
that is not significant against the live controller ($p = 0.08$) but
remains significant against Uniform ($p = 0.03$). At $\sigma \le
10\%$, all baseline comparisons are highly significant
($p < 10^{-3}$). The shape (diminishing returns) of the curves
matters more than the precise parameter values: the optimizer maps
roughly the same $(a, b)$ region to roughly the same allocation, so
the curves only need to be \emph{directionally} correct. \chg{This test perturbs each parameter }\emph{\chg{independently}}\chg{; the correlated-error version, which is the more realistic failure mode, is in Appendix~\ref{app:corr_noise}.
}

The $\sigma = 0\%$ averaged-curve
condition slightly outperforms the live controller
($0.5305$ vs.\ $0.5258$), so per-run estimation \emph{variance} can be a source of allocation suboptimality. 
\begin{table}[h]
\centering
\small
\begin{tabular}{l c c c c}
\toprule
Strategy & Mean & vs.\ live ($p$) & vs.\ LLM ($p$) & vs.\ Uniform ($p$) \\
\midrule
Live controller (additive)   & $0.5258$ & ---     & $0.0013$\textsuperscript{**} & $<10^{-4}$\textsuperscript{***} \\
$\sigma = 0\%$ (averaged)    & $0.5305$ & $0.52$\textsuperscript{ns} & $0.0009$\textsuperscript{***} & $<10^{-4}$\textsuperscript{***} \\
$\sigma = 10\%$              & $0.5384$ & $0.32$\textsuperscript{ns} & $0.0004$\textsuperscript{***} & $<10^{-4}$\textsuperscript{***} \\
$\sigma = 50\%$              & $0.5165$ & $0.08$\textsuperscript{ns} & $0.30$\textsuperscript{ns}    & $0.033$\textsuperscript{*}     \\
LLM-direct                   & $0.5038$ & $0.0013$\textsuperscript{**} & ---       & $0.025$\textsuperscript{*}     \\
Uniform                      & $0.4883$ & $<10^{-4}$\textsuperscript{***} & $0.025$\textsuperscript{*} & ---     \\
\bottomrule
\end{tabular}
\caption{\textbf{Sensitivity to curve-estimation noise} ($n = 150$
tasks, $\alpha = 0.5$). Multiplicative Gaussian noise on $(a, b)$
with $\sigma \in \{0\%, 10\%, 50\%\}$. $p$-values are paired
Wilcoxon signed-rank, two-sided. Even at $\sigma = 50\%$, the
allocation does not significantly underperform the live controller.}
\label{tab:sensitivity}
\end{table}

\subsection{Controller-model variance: gpt-4o vs.\ gpt-4o-mini}
\label{app:controller_compare}

We re-run only the allocate phase in the $\alpha = 0.5$ benchmark
with \texttt{gpt-4o-mini}  replacing \texttt{gpt-4o} in the
curve-estimation step and compare the resulting per-task mean
fractional allocations on \texttt{additive} and
\texttt{mult\_offset}. The two controllers produce highly correlated
allocations (Table~\ref{tab:controller_compare}). On
\texttt{additive} the per-phase-averaged Pearson correlation is
$0.948$ but the mean absolute fractional difference is $5.4$ pp,
driven by the larger-model controller pushing $\sim$$15$ pp from
\texttt{refine} back to \texttt{decompose}$/$\texttt{implement} on
easy tasks. On \texttt{mult\_offset} the controllers agree more
tightly on the actual fractions (MAE $1.75$ pp) with a similar
overall correlation ($0.924$). Within either controller,
run-to-run standard deviation is small ($\le 4.3$ pp on
\texttt{refine}, $\le 1.4$ pp on the upstream phases). The
framework's allocation is therefore not pinned to a particular
controller LLM.

\begin{table}[h]
\centering
\small
\begin{tabular}{l l c c c c}
\toprule
Objective & Tier & Pearson & Spearman & MAE & P90 \\
\midrule
\multirow{4}{*}{\texttt{additive}}     & all    & $0.948$ & $0.863$ & $0.054$ & $0.134$ \\
                                       & easy   & $0.848$ & $0.644$ & $0.083$ & $0.194$ \\
                                       & medium & $0.961$ & $0.869$ & $0.046$ & $0.114$ \\
                                       & hard   & $0.946$ & $0.555$ & $0.034$ & $0.068$ \\
\midrule
\multirow{4}{*}{\texttt{mult\_offset}} & all    & $0.924$ & $0.910$ & $0.018$ & $0.048$ \\
                                       & easy   & $0.789$ & $0.642$ & $0.030$ & $0.088$ \\
                                       & medium & $0.935$ & $0.890$ & $0.016$ & $0.045$ \\
                                       & hard   & $0.964$ & $0.791$ & $0.007$ & $0.017$ \\
\bottomrule
\end{tabular}
\caption{\textbf{Controller-model agreement on per-task fractional
allocations} (\texttt{gpt-4o} vs.\ \texttt{gpt-4o-mini}, $n = 150$,
$\alpha = 0.5$). Pearson/Spearman correlations are computed
\emph{independently per phase} on the per-task mean fractional
allocations, then averaged across the four phases;
\textbf{MAE} is the mean absolute disagreement on a phase fraction
across all $4 \times n$ phase-task pairs; \textbf{P90} is the
$90$th percentile of the same disagreements (in pp of budget). The
two controllers agree closely on \texttt{mult\_offset} and
reasonably on \texttt{additive}; the largest disagreements
concentrate on easy tasks, where any allocation works.}
\label{tab:controller_compare}
\end{table}

\subsection{\chg{Open-weight controllers: does the ordering survive?}}
\label{app:open_weight}

\chg{Every experiment above uses a GPT-4-family model as the controller, which
leaves open whether ZEBRA's advantage is a property of the method or an
artifact of one model family: a sufficiently strong proprietary model might be
the only one able to produce usable curves. We therefore re-run the allocation
step with open-weight controllers, holding the pipeline, the executor models,
the budget, and the prompts fixed, so that only the model that estimates
curves (for ZEBRA) or emits the split directly (for LLM-direct) changes.
Crucially, the controller is swapped for }\emph{\chg{both}} \chg{ZEBRA and LLM-direct in
each column.
}

\paragraph{\chg{APPS.}} \chg{We use the $60$-task tier-balanced subset
($20$ easy / $20$ medium / $20$ hard) at $\alpha = 0.5$, $15$ runs per (task,
strategy) cell, with }\texttt{\chg{gpt-4o-mini}} \chg{as the executor and }\texttt{\chg{gpt-4o}}
\chg{for }\texttt{\chg{refine}}\chg{. The two open-weight controllers are Qwen3.5-9B and
DeepSeek-V4-Pro. The GPT-4o column is the main run restricted to the same $60$ tasks, so all
three columns are paired task-for-task. Results are in Table~\ref{tab:open_weight_apps}.
}

\begin{table}[h]
\centering
\small
\begin{tabular}{l c c c}
\toprule
\chg{Strategy }& \chg{GPT-4o }& \chg{Qwen3.5-9B }& \chg{DeepSeek-V4-Pro }\\
\midrule
\texttt{\chg{zebra-additive}}     & \chg{$0.5427$ }& \chg{$0.5361$ }& \chg{$0.5293$ }\\
\texttt{\chg{zebra-mult\_offset}} & \chg{$0.5229$ }& \chg{$\mathbf{0.5394}$ }& \chg{$\mathbf{0.5355}$ }\\
\midrule
\chg{LLM-direct                  }& \chg{$0.5104$ }& \chg{$0.4841$ }& \chg{$0.5081$ }\\
\bottomrule
\end{tabular}
\caption{\textbf{\chg{APPS: the zebra $>$ llm ordering holds under open-weight
controllers.}} \chg{Mean score on $60$ tier-balanced APPS tasks, $\alpha = 0.5$,
$15$ runs per cell. Executor $=$ }\texttt{\chg{gpt-4o-mini}}\chg{, }\texttt{\chg{refine}} \chg{model
$=$ }\texttt{\chg{gpt-4o}}\chg{; only the controller varies across columns. Both ZEBRA
variants beat LLM-direct in every column.}}
\label{tab:open_weight_apps}
\end{table}

\paragraph{\chg{HotpotQA.}} \chg{To check that this is not specific to the coding
pipeline, we repeat the swap on the $114$-task HotpotQA benchmark at
$\alpha = 0.5$, with }\texttt{\chg{gpt-4o-mini}} \chg{as the phase model. The baseline column uses
}\texttt{\chg{gpt-4o-mini}} \chg{as the controller, as in the main HotpotQA experiment;
the open-weight controller is Qwen3.5-9B. We report absolute token-level F1, with paired Wilcoxon tests on
per-task mean F1 against LLM-direct
(Table~\ref{tab:open_weight_hotpotqa}).
}

\begin{table}[h]
\centering
\small
\setlength{\tabcolsep}{5pt}
\begin{tabular}{l c c c c}
\toprule
& \multicolumn{2}{c}{GPT-4o-mini (baseline)} & \multicolumn{2}{c}{Qwen3.5-9B} \\
\cmidrule(lr){2-3} \cmidrule(lr){4-5}
\chg{Strategy }& \chg{F1 }& \chg{$p$ vs.\ LLM }& \chg{F1 }& \chg{$p$ vs.\ LLM }\\
\midrule
\texttt{\chg{zebra-additive}}     & \chg{$0.4737$ }& \chg{$0.0012$\textsuperscript{**} }& \chg{$0.4514$ }& \chg{$<10^{-4}$\textsuperscript{***} }\\
\texttt{\chg{zebra-mult\_offset}} & \chg{$0.4801$ }& \chg{$0.0036$\textsuperscript{**} }& \chg{$0.4549$ }& \chg{$<10^{-4}$\textsuperscript{***} }\\
\midrule
\chg{LLM-direct                  }& \chg{$0.4302$ }& \chg{---                           }& \chg{$0.4011$ }& \chg{--- }\\
\bottomrule
\end{tabular}
\caption{\textbf{\chg{HotpotQA: the ordering also holds off the coding pipeline.}}
\chg{Mean token-level F1 on $114$ HotpotQA tasks, $\alpha = 0.5$. Phase
model $=$ }\texttt{\chg{gpt-4o-mini}} \chg{throughout; only the controller varies.
$p$-values are paired Wilcoxon signed-rank, two-sided, on per-task mean F1
against LLM-direct ($n = 114$). Runs per cell: $15$.}}
\label{tab:open_weight_hotpotqa}
\end{table}

\paragraph{\chg{Reading.}} \chg{ZEBRA is robust to the controller's model family. In all
five controller$\times$benchmark columns both variants beat LLM-direct. The absolute level barely moves: across the
three APPS controllers }\texttt{\chg{zebra-additive}} \chg{stays within $0.013$
($0.5427$, $0.5361$, $0.5293$) and }\texttt{\chg{mult\_offset}} \chg{within $0.017$
($0.5229$, $0.5394$, $0.5355$). The margin over LLM-direct ranges from
$+0.013$ to $+0.055$ on APPS, and the two largest margins both occur with an
open-weight controller ($+0.052$ for }\texttt{\chg{additive}} \chg{and $+0.055$ for
}\texttt{\chg{mult\_offset}} \chg{under Qwen3.5-9B, against $+0.032$ and $+0.013$ under
GPT-4o); on HotpotQA it likewise widens from $+0.044$ to $+0.050$ F1.
The curve estimation does not depend on a closed-weight model.
}

\subsection{Per-task adaptation: comparison to a fixed average split}
\label{app:fixed_avg}

To test whether ZEBRA's gain comes from \emph{per-task adaptation} or
merely from knowing the right average ratio, we construct a
\textbf{Fixed-Avg} baseline that pre-computes the global mean
per-phase share of \texttt{zebra-additive} across all
$150 \times 15 = 2{,}250$ samples at $\alpha = 0.5$
($11.3/14.0/24.1/50.6\%$ for plan/decompose/implement/refine) and
applies that fixed split to every task. Fixed-Avg has \emph{zero}
controller cost (the ratios are pre-computed) and is otherwise
identical to ZEBRA in pipeline, models, and per-task total budget;
it is the strongest non-adaptive baseline available without
training.

\begin{table}[h]
\centering
\small
\begin{tabular}{l c c c c}
\toprule
                                 & all & easy & medium & hard \\
\midrule
Fixed-Avg (mean score)            & $0.5048$ & $0.9253$ & $0.4866$ & $0.1023$ \\
Fixed-Avg (NB retention)          & $83.4\%$ & $95.4\%$ & $78.7\%$ & $74.8\%$ \\
\midrule
\texttt{zebra-additive} (mean)    & $0.5258$ & $0.9216$ & $0.5324$ & $0.1234$ \\
\texttt{zebra-additive} (NB ret.) & $89.4\%$ & $95.1\%$ & $87.7\%$ & $84.7\%$ \\
$p$ vs.\ Fixed-Avg                & $0.011$\textsuperscript{*}  & $0.48$\textsuperscript{ns} & $0.008$\textsuperscript{**}  & $0.17$\textsuperscript{ns} \\
\midrule
\texttt{mult\_offset} (mean)      & $0.5397$ & $0.9385$ & $0.5506$ & $0.1300$ \\
\texttt{mult\_offset} (NB ret.)   & $94.3\%$ & $97.0\%$ & $90.1\%$ & $96.3\%$ \\
$p$ vs.\ Fixed-Avg                & $<10^{-4}$\textsuperscript{***} & $0.06$\textsuperscript{ns} & $<10^{-3}$\textsuperscript{***} & $0.009$\textsuperscript{**} \\
\bottomrule
\end{tabular}
\caption{\textbf{Dynamic ZEBRA vs.\ Fixed-Avg.} NB retention is
computed as a mean of per-task ratios,
$\mathrm{mean}_i(s_i/s_i^{\text{NB}})$, averaged over tasks with
$s_i^{\text{NB}} \ge 0.01$. Both ZEBRA variants
are statistically indistinguishable from Fixed-Avg on easy tasks --
where any reasonable split works -- and outperform it on medium and
hard tasks. The gap is statistically significant in $3$ of $4$
medium/hard cells: on medium tasks for both variants
(\texttt{additive} $p = 0.008$, \texttt{mult\_offset} $p < 10^{-3}$),
and on hard tasks for \texttt{mult\_offset} ($p = 0.009$) but not
\texttt{additive} ($p = 0.17$). The gain from ZEBRA therefore comes
from per-task adaptation, not from the choice of average ratio.}
\label{tab:fixed_avg}
\end{table}

\paragraph{\chg{A tighter budget ($\alpha = 0.3$).}} \chg{The comparison above is run at
$\alpha = 0.5$. We
repeat it at $\alpha = 0.3$, on $100$ APPS tasks ($50$ easy and $50$ hard;
this run has no medium tier), $15$ runs per (task, strategy) cell, with
}\texttt{\chg{gpt-4o-mini}} \chg{for }\texttt{\chg{plan}} \chg{/ }\texttt{\chg{decompose}} \chg{/
}\texttt{\chg{implement}} \chg{and the controller, }\texttt{\chg{gpt-4o}} \chg{for }\texttt{\chg{refine}}\chg{.
Fixed-Avg uses the same pre-computed split as above. The $50$ easy tasks are
the same set as in Table~\ref{tab:easy_alpha03}, re-run here alongside
Fixed-Avg.
}\begin{table}[h]
\centering
\small
\begin{tabular}{l c c}
\toprule
\textbf{\chg{Strategy}} & \textbf{\chg{Easy}} \chg{($n = 50$) }& \textbf{\chg{Hard}} \chg{($n = 50$) }\\
\midrule
\chg{Fixed-Avg               }& \chg{$0.9217$ }& \chg{$0.1020$          }\\
\texttt{\chg{zebra-additive}} & \chg{$0.9242$          }& \chg{$\mathbf{0.1115}$ }\\
\bottomrule
\end{tabular}
\caption{\textbf{\chg{Dynamic ZEBRA vs.\ Fixed-Avg at a very tight budget}}
\chg{($\alpha = 0.3$, $15$ runs per cell). Mean score per tier. The two are
effectively tied on easy tasks; on hard tasks }\texttt{\chg{zebra-additive}} \chg{leads
by $+0.0095$ ($+9.3\%$ relative).}}
\label{tab:fixed_avg_alpha03}
\end{table}

\chg{The pattern is the same one seen at $\alpha = 0.5$, and it does not soften as
the budget tightens: easy tasks are near saturation, so any reasonable split
reaches them, while on hard tasks -- where the split has to be right --
per-task adaptation still buys the gain. We report means for this run; it was
not powered for a paired significance test at the tier level.
}

\paragraph{Takeaway.}
On easy tasks, dynamic ZEBRA and Fixed-Avg are statistically
indistinguishable -- when the budget is loose, the choice of split
does not matter. On medium and hard tasks (the regimes
Section~\ref{sec:results_by_tier} identifies as where allocation
matters), both ZEBRA variants outperform Fixed-Avg in absolute
retention by $+9$ to $+22$ pp NB-retention. The gap is
statistically significant in $3$ of the $4$ medium/hard cells
($p < 0.01$); the one exception is \texttt{additive} vs.\
Fixed-Avg on hard ($p = 0.17$), where \texttt{mult\_offset} retains
significance ($p = 0.009$). Even so, Fixed-Avg uses ZEBRA's own
\emph{average} split: the benefit ZEBRA delivers is not the
average ratio (any practitioner could hard-code that), but the
per-task variation around it. \chg{The same split of behaviour appears at
$\alpha = 0.3$ (Table~\ref{tab:fixed_avg_alpha03}), so this is a property of
the tier rather than of the budget level.
}

\subsection{\chg{A cost-proportional heuristic}}
\label{app:cost_prop}

\chg{Fixed-Avg above borrows ZEBRA's own average split. A sharper question is
whether a split can be read off the pipeline directly, with no allocator at
all: give each phase the share of the budget it consumes when the pipeline
runs unconstrained. 
}

\chg{We compute the ratios from every no-budget run recorded across the $150$
paper tasks. For each (task, run) we take each phase's share of that run's
total spend, average over the task's runs, then average across the
$150$ tasks, so every task contributes equally regardless of its dollar
spend. This gives $9.0 / 5.4 / 14.8 / 70.8\%$ for
}\texttt{\chg{plan}} \chg{/ }\texttt{\chg{decompose}} \chg{/ }\texttt{\chg{implement}} \chg{/ }\texttt{\chg{refine}} \chg{--
markedly more }\texttt{\chg{refine}}\chg{-heavy than ZEBRA's own average split
($11.3 / 14.0 / 24.1 / 50.6\%$). The baseline applies these fixed ratios to
every task with no controller call. Pipeline, models, task pool and budget are identical to the main results, so the rows below are directly
comparable to Table~\ref{tab:main_overall}.
}

\begin{table}[h]
\centering
\small
\begin{tabular}{l c c c}
\toprule
\textbf{\chg{Strategy}} & \textbf{\chg{Mean score}} & \chg{$\boldsymbol{\Delta}$ }\textbf{\chg{vs Cost-Prop}} & \chg{$\boldsymbol{p}$ }\\
\midrule
\texttt{\chg{zebra-mult\_offset}} & \chg{$\mathbf{0.5397}$ }& \chg{$+0.0105$ }& \chg{$0.049$\textsuperscript{*}     }\\
\chg{Cost-Prop                   }& \chg{$0.5292$          }& \chg{---       }& \chg{---                            }\\
\texttt{\chg{zebra-additive}}     & \chg{$0.5258$          }& \chg{$-0.0034$ }& \chg{$0.45$\textsuperscript{ns}     }\\
\chg{LLM-direct                  }& \chg{$0.5038$          }& \chg{$-0.0254$ }& \chg{$0.0004$\textsuperscript{***}  }\\
\bottomrule
\end{tabular}
\caption{\textbf{\chg{Cost-proportional allocation is a strong baseline}}
\chg{($150$ APPS tasks, $\alpha = 0.5$, $15$ runs per cell). Fixed ratios
$9.0/5.4/14.8/70.8\%$  applied to every
task with no controller call. $p$ is paired Wilcoxon signed-rank, two-sided,
on per-task means against Cost-Prop.}}
\label{tab:cost_prop}
\end{table}

\paragraph{\chg{Reading.}} \chg{This is the strongest baseline we tested, and we report
it as such. Cost-Prop beats LLM-direct by a wide and significant margin
($+0.0254$, $p = 4 \times 10^{-4}$) and is statistically indistinguishable
from }\texttt{\chg{zebra-additive}} \chg{($-0.0034$, $p = 0.45$); only
}\texttt{\chg{mult\_offset}} \chg{is ahead of it ($+0.0105$, $p = 0.049$).
The ratios are
not free: they are measured from unconstrained runs of this
pipeline on these tasks, so Cost-Prop is a profiled baseline, not a zero-shot
one, and it has to be re-profiled for every new pipeline (unlike ZEBRA).  Cost-Prop shows that on a single well-profiled
pipeline a good static split goes a long way; it does not transfer, and
producing it costs more measurement than running the controller.
}

\subsection{Controller overhead vs.\ allocation gain}
\label{app:overhead}

We answer two questions: (a) \emph{how large is the controller
overhead in practice?} and (b) \emph{is the allocation gain large
enough to justify it, even against a comparator with substantially
more budget?}

\paragraph{(a) Controller overhead statistics.}
The controller's curve-estimation call has an essentially fixed
cost per task (it depends on the task description, not on the
budget), so its share of the total spend grows as the budget
shrinks. Table~\ref{tab:overhead_stats} reports overhead summaries
per strategy at the main benchmark budget ($\alpha = 0.5$).

\begin{table}[h]
\centering
\small
\setlength{\tabcolsep}{4pt}
\begin{tabular}{l r r r r r}
\toprule
Strategy & $n$ runs & Ctrl.\ \$ (mean) & Total \$ & Phase \$ &
Ctrl.\ / total spent \\
\midrule
\texttt{zebra-additive}             & $735$ & $\$0.00362$ & $\$0.01244$ & $\$0.00882$ & $29.1\%$ \\
\texttt{zebra-mult\_offset}         & $735$ & $\$0.00362$ & $\$0.01242$ & $\$0.00880$ & $29.2\%$ \\
LLM-direct                          & $735$ & $\$0.00234$ & $\$0.00991$ & $\$0.00757$ & $23.6\%$ \\
\bottomrule
\end{tabular}
\caption{\textbf{Controller overhead statistics.}
Per-strategy summaries at the main benchmark budget ($\alpha = 0.5$),
restricted to the $49$ APPS tasks with valid raw phase-spending data
($49 \times 15 = 735$ runs each); the remaining $101$ tasks are
excluded because of a logging artifact. ZEBRA's controller overhead is
$29.1\%$ of total spend for \texttt{additive} and $29.2\%$ for
\texttt{mult\_offset} ($33.1\%$ of the budget in both cases).}
\label{tab:overhead_stats}
\end{table}

At $\alpha = 0.5$ the controller consumes $29.1\%$ of
\texttt{zebra-additive} total spend and $29.2\%$ of
\texttt{mult\_offset} (about $33\%$ of the per-task budget for
both). Because the controller call has a roughly fixed cost per
task while the phase budget shrinks with $\alpha$, this share grows
at tighter budgets and becomes a non-trivial fraction of the total.

\paragraph{(b) Does the allocation gain justify the overhead?}
We compare ZEBRA at $\alpha = 0.5$ against \emph{Uniform} at
$\alpha = 0.8$, a comparator with $60\%$ \emph{more} total budget
and zero controller overhead. If allocation alone is doing real
work, ZEBRA should still win even at this budget disadvantage.
Table~\ref{tab:overhead} reports the head-to-head.

\begin{table}[h]
\centering
\small
\begin{tabular}{l c c c c}
\toprule
Comparison & Mean diff. & $95\%$ CI & Wilcoxon $p$ & W/L/T \\
\midrule
\texttt{zebra-additive}@$\alpha{=}0.5$ vs.\ Uniform@$\alpha{=}0.8$
   & $+0.0273$ & $[+0.013, +0.042]$ & $1.6\!\times\!10^{-4}$ & $59/33/58$ \\
\texttt{mult\_offset}@$\alpha{=}0.5$ vs.\ Uniform@$\alpha{=}0.8$
   & $+0.0412$ & $[+0.025, +0.057]$ & $<\!10^{-6}$            & $69/20/61$ \\
\midrule
\texttt{zebra-additive}@$\alpha{=}0.5$ vs.\ Uniform@$\alpha{=}0.5$
   & $+0.0375$ & $[+0.023, +0.052]$ & $3.0\!\times\!10^{-6}$  & --- \\
\texttt{mult\_offset}@$\alpha{=}0.5$ vs.\ Uniform@$\alpha{=}0.5$
   & $+0.0514$ & $[+0.035, +0.068]$ & $<\!10^{-6}$            & --- \\
\bottomrule
\end{tabular}
\caption{\textbf{Allocation gain vs.\ controller overhead.}
Top: ZEBRA at $\alpha = 0.5$ vs.\ Uniform at $\alpha = 0.8$ -- a
comparator given $60\%$ more total budget ($1.6\!\times\!$) but
zero controller cost. Both ZEBRA variants still win significantly.
Bottom: same-budget reference ($\alpha = 0.5$ on both sides). All
comparisons are paired Wilcoxon signed-rank on $n = 150$ APPS tasks
($15$ runs per cell). W/L/T sums to $150$ for the top block.}
\label{tab:overhead}
\end{table}

Both ZEBRA variants beat Uniform-$0.8$ significantly, despite the
$60\%$ budget disadvantage. ZEBRA's controller overhead at
$\alpha = 0.5$ -- $\sim$$33\%$ of budget for both \texttt{additive}
and \texttt{mult\_offset} ($\sim$$29\%$ of total spend) -- is
therefore more than compensated by the allocation gain.

\subsection{\chg{Cost normalization and budget utilization}}
\label{app:cost_normalization}

\chg{The comparisons in this paper are iso-}\emph{\chg{budget}}\chg{: every allocator receives
the same per-task $B$. They are not iso-}\emph{\chg{spend}}\chg{, since realized spend is
an emergent consequence of how an allocator distributes $B$. This appendix
reports the cost-normalized view, explains what drives the difference, and
sets out why we do not treat score-per-dollar as the headline metric under a
committed budget.
}

\paragraph{\chg{Performance per dollar.}} \chg{Table~\ref{tab:perf_per_dollar} divides
mean score by mean total realized spend (controller plus all phases).
}\texttt{\chg{mult\_offset}} \chg{beats LLM-direct on this metric at $\alpha = 0.5$; at
$\alpha = 0.8$, and for }\texttt{\chg{additive}} \chg{at both budgets, LLM-direct wins.
}

\begin{table}[h]
\centering
\small
\begin{tabular}{l c c}
\toprule
\chg{Strategy }& \chg{$\alpha = 0.5$ score/total\$ }& \chg{$\alpha = 0.8$ score/total\$ }\\
\midrule
\chg{LLM-direct                  }& \chg{$46.34$          }& \chg{$\mathbf{38.35}$ }\\
\texttt{\chg{zebra-additive}}     & \chg{$39.25$          }& \chg{$29.20$ }\\
\texttt{\chg{zebra-mult\_offset}} & \chg{$\mathbf{49.13}$ }& \chg{$33.48$ }\\
\bottomrule
\end{tabular}
\caption{\textbf{\chg{Performance per dollar.}} \chg{Mean score divided by mean total
realized spend (controller $+$ all phases), $150$ APPS tasks $\times\,15$ runs
per (task, strategy).}}
\label{tab:perf_per_dollar}
\end{table}

\chg{Two things explain these numbers. First, they are what the model predicts:
the phase curves $u(x) = a(1 - e^{-bx})$ are concave and saturating, so
additional budget buys progressively less score per dollar. A strategy that
converts more of the budget into output will therefore show a lower ratio by
construction, and the ratio is not a measure of allocation quality. Second,
and more concretely, LLM-direct's lower spend is not restraint but waste: it
gives some phases more budget than they can use, so the surplus is left
unspent, while other phases are starved and would have converted more budget
into score. The LLM strategy leaves points on the table.
}

\paragraph{\chg{Where the budget is stranded.}} \chg{Every strategy allocates the same
total, what differs is how much of it can actually be spent.
Table~\ref{tab:utilization} shows this in the utilization column
($\mathrm{util} = \mathrm{spend}/\mathrm{alloc}$). LLM-direct allocates
generously to }\texttt{\chg{decompose}} \chg{and }\texttt{\chg{implement}}\chg{, which then run at
$40\%$ and $55\%$ utilization -- budget that is unspendable by construction,
because those phases cannot consume the caps they were given. ZEBRA allocates
}\emph{\chg{less}} \chg{to the same phases, so they run tighter against smaller caps
($57\%$ and $80\%$ for }\texttt{\chg{additive}}\chg{), and moves the reclaimed cap to
}\texttt{\chg{refine}}\chg{, raising its allocation from $6.333$m to $8.101$m -- which
}\texttt{\chg{refine}} \chg{absorbs, still at $82\%$ utilization. In total, LLM-direct
utilizes $70\%$ of its allocation and leaves $30.2\%$ of the budget unspent,
against $80\%$ and $20.1\%$ for }\texttt{\chg{additive}} \chg{and $79\%$ and $20.6\%$ for
}\texttt{\chg{mult\_offset}}\chg{.
}

\begin{table}[h]
\centering
\small
\setlength{\tabcolsep}{5pt}
\begin{tabular}{l l r r c r r}
\toprule
\chg{Strategy }& \chg{Phase }& \chg{alloc \$ }& \chg{spend \$ }& \chg{util }& \chg{leftover \$ }& \chg{\% of budget }\\
\midrule
\chg{LLM-direct }& \texttt{\chg{plan}}      & \chg{$1.376$m }& \chg{$0.921$m }& \chg{$67\%$ }& \chg{$0.455$m }& \chg{$3.7\%$ }\\
           & \texttt{\chg{decompose}} & \chg{$1.438$m }& \chg{$0.575$m }& \chg{$40\%$ }& \chg{$0.863$m }& \chg{$7.1\%$ }\\
           & \texttt{\chg{implement}} & \chg{$3.004$m }& \chg{$1.638$m }& \chg{$55\%$ }& \chg{$1.366$m }& \chg{$11.2\%$ }\\
           & \texttt{\chg{refine}}    & \chg{$6.333$m }& \chg{$5.341$m }& \chg{$84\%$ }& \chg{$0.991$m }& \chg{$8.2\%$ }\\
           & \textbf{\chg{total}}     & \chg{$\mathbf{12.151}$m }& \chg{$\mathbf{8.476}$m }& \chg{$\mathbf{70\%}$ }& \chg{$\mathbf{3.675}$m }& \chg{$\mathbf{30.2\%}$ }\\
\midrule
\texttt{\chg{zebra-additive}}
           & \texttt{\chg{plan}}      & \chg{$0.870$m }& \chg{$0.780$m }& \chg{$90\%$ }& \chg{$0.090$m }& \chg{$0.7\%$ }\\
           & \texttt{\chg{decompose}} & \chg{$1.090$m }& \chg{$0.624$m }& \chg{$57\%$ }& \chg{$0.466$m }& \chg{$3.8\%$ }\\
           & \texttt{\chg{implement}} & \chg{$2.090$m }& \chg{$1.679$m }& \chg{$80\%$ }& \chg{$0.411$m }& \chg{$3.4\%$ }\\
           & \texttt{\chg{refine}}    & \chg{$8.101$m }& \chg{$6.628$m }& \chg{$82\%$ }& \chg{$1.472$m }& \chg{$12.1\%$ }\\
           & \textbf{\chg{total}}     & \chg{$\mathbf{12.151}$m }& \chg{$\mathbf{9.712}$m }& \chg{$\mathbf{80\%}$ }& \chg{$\mathbf{2.439}$m }& \chg{$\mathbf{20.1\%}$ }\\
\midrule
\texttt{\chg{zebra-mult\_offset}}
           & \texttt{\chg{plan}}      & \chg{$1.057$m }& \chg{$0.855$m }& \chg{$81\%$ }& \chg{$0.202$m }& \chg{$1.7\%$ }\\
           & \texttt{\chg{decompose}} & \chg{$0.888$m }& \chg{$0.572$m }& \chg{$64\%$ }& \chg{$0.316$m }& \chg{$2.6\%$ }\\
           & \texttt{\chg{implement}} & \chg{$1.809$m }& \chg{$1.524$m }& \chg{$84\%$ }& \chg{$0.285$m }& \chg{$2.3\%$ }\\
           & \texttt{\chg{refine}}    & \chg{$8.397$m }& \chg{$6.698$m }& \chg{$80\%$ }& \chg{$1.699$m }& \chg{$14.0\%$ }\\
           & \textbf{\chg{total}}     & \chg{$\mathbf{12.151}$m }& \chg{$\mathbf{9.650}$m }& \chg{$\mathbf{79\%}$ }& \chg{$\mathbf{2.501}$m }& \chg{$\mathbf{20.6\%}$ }\\
\bottomrule
\end{tabular}
\caption{\textbf{\chg{Per-phase allocated vs.\ spent (APPS, $\alpha = 0.5$, all
$150$ tasks).}} \chg{$\mathrm{util} = \mathrm{spend}/\mathrm{alloc}$;
$\mathrm{leftover} = \mathrm{alloc} - \mathrm{spend}$;
\% of budget $=$ leftover$/$total alloc. Values in millidollars
($\mathrm{m} = \$10^{-3}$). Total allocation is identical across strategies;
LLM-direct over-allocates to phases that cannot spend their cap and starves
the phase that can.}}
\label{tab:utilization}
\end{table}

\paragraph{\chg{Why raw score is the right headline under a fixed budget.}} \chg{The
setting this paper studies is one where the user has already committed $B$ and
wants the best output within it: the budget is fixed and free at the margin,
and money not spent is not returned. Under that objective, spending less is
not a virtue. Since the phase curves are monotone, $u'(x) > 0$ below the
plateau, so every unspent dollar strictly leaves score unclaimed; a
score-per-dollar ratio, by contrast, is maximized as $x \to 0$ and is
therefore optimized by an allocator that declines to spend at all. Ranking by
that ratio rewards precisely the behaviour a budgeted user does not want.
}

\chg{An analogy makes the asymmetry concrete. Two students each have $20$ hours to
prepare for an exam. The first studies $10$ hours and scores $80$, so
cost-normalized performance is $8.0$; the second studies $18$ hours and scores
$100$, so cost-normalized performance is $5.6$. Given $20$ free hours the
second student has behaved better, despite the lower ratio: the first
student's superior efficiency is bought by leaving $10$ hours -- and $20$
points -- unused. The extra points are exactly where the extra time went, and
the saturating shape of our curves is what captures that phenomenon.
}

\subsection{Empirical validation of pipeline monotonicity}
\label{app:monotonicity}

ZEBRA's solver assumes each phase is non-decreasing in spend
(Section~\ref{sec:cnk}). Recent work on
``overthinking''~\citep{chen2024overthinking,sui2025stop}
shows this assumption can fail for individual chain-of-thought
prompts: very long reasoning traces can hurt accuracy on simple
tasks. We therefore empirically validate the assumption \emph{for
our pipeline} by fitting a within-task fixed-effects regression of
final score on total realized spend, separately for each tier:
\begin{equation*}
\text{score}_{t,r} = \mu_t + \beta\,\text{spend}_{t,r} + \varepsilon_{t,r},
\end{equation*}
where $t$ indexes task, $r$ indexes run, and $\mu_t$ is a per-task
intercept. Pooling $50$ tasks per tier
($n = 6{,}000$ per tier; $18{,}000$ total runs across our APPS
benchmark and ablations) gives the within-task slopes in
Table~\ref{tab:monotonicity}.

\begin{table}[h]
\centering
\small
\begin{tabular}{l c c c c}
\toprule
Tier & $n$ runs & $\hat\beta$ (score per \$) & 95\% CI & $p$ \\
\midrule
Easy   & $6{,}000$ & $9.37$ & $[6.56, 12.19]$ & $<10^{-6}$ \\
Medium & $6{,}000$ & $5.19$ & $[4.15,  6.24]$ & $<10^{-6}$ \\
Hard   & $6{,}000$ & $1.83$ & $[1.19,  2.46]$ & $<10^{-6}$ \\
\bottomrule
\end{tabular}
\caption{\textbf{Pipeline is monotone in total spend on every
tier.} Within-task fixed-effects slopes of score on total realized
spend, fit on $18{,}000$ runs ($150$ tasks $\times 120$ runs).
$\hat\beta > 0$ at $p < 10^{-6}$ in every tier validates the
non-decreasing-utility assumption underlying the
solver.}
\label{tab:monotonicity}
\end{table}

The slope is positive and highly significant in every tier,
confirming that \emph{within} a task, more budget yields higher
score. The slope shrinks with difficulty -- $9.4 \to 5.2 \to 1.8$
score-per-dollar from easy to hard -- consistent with the
diminishing-returns shape that motivates ZEBRA's saturating
exponential. The assumption is therefore appropriate for this
pipeline; the overthinking failure mode reported by
\citet{chen2024overthinking,sui2025stop} would manifest as a
\emph{negative} within-task slope on at least one tier, which we do
not observe\chg{. See Appendix~\ref{app:curve_validation} for how this result fits with the other checks on the estimated curves}.

\section{\chg{Validation of the Estimated Phase Utility Curves}}
\label{app:curve_validation}
\begingroup\chgon

ZEBRA's solver is only as good as the per-phase curves it is handed, so the
central empirical question about the method is whether the controller's estimated
curves carry real, task-specific signal, or merely encode a roughly correct
structural prior (``spend more on \texttt{refine}''). This appendix collects our answer in one place.
Table~\ref{tab:curve_validation_map} maps each sub-question to the experiment
that addresses it.

\begin{table}[h]\chgon 
\centering
\small
\setlength{\tabcolsep}{4pt}
\begin{tabular}{p{0.40\textwidth} p{0.30\textwidth} l}
\toprule
Question about the curves & Experiment & Where \\
\midrule
Is the pipeline actually monotone in spend, as the curve family assumes?
  & Within-task fixed-effects slope of score on spend, per tier
  & App.~\ref{app:monotonicity} \\
Does the predicted \emph{shape} match the measured phase response?
  & Post-hoc calibration of the \texttt{refine} curve against empirical
    $\Delta$score
  & App.~\ref{app:curve_calibration} \\
How much independent estimation error can the allocation absorb?
  & i.i.d.\ multiplicative Gaussian noise on $(a_i, b_i)$
  & App.~\ref{app:sensitivity} \\
How much \emph{correlated} estimation error can it absorb?
  & Shared-latent correlated noise, $\rho = 0.7$
  & App.~\ref{app:corr_noise} \\
Do the curves carry per-task and per-phase signal, or only a structural prior?
  & Cross-task and within-task curve shuffles
  & App.~\ref{app:shuffle} \\
\bottomrule
\end{tabular}
\caption{\textbf{Where each aspect of curve validation is addressed.}
The first three rows are reported elsewhere in the paper; the last two are
the experiments introduced in this appendix.}
\label{tab:curve_validation_map}
\end{table}

\paragraph{Common setup for the two new experiments.} Both use the same
$60$-task tier-balanced APPS subset ($20$ easy / $20$ medium / $20$ hard),
$\alpha = 0.5$, $15$ runs per (task, condition) cell -- the sole exception is
the $\sigma = 0$ cell of the correlated-noise experiment, where the injected
curves are identical across runs and $5$ runs suffice -- with
\texttt{zebra-additive} as the allocator, \texttt{gpt-4o-mini} as the
executor, \texttt{gpt-4o} for \texttt{refine}, and \texttt{gpt-4o} as the
controller. Only the curves handed to the solver are altered; the pipeline,
the models, and the budget are untouched, so any difference is attributable
to curve information alone. The clean condition scores $0.5427$ and
LLM-direct on the same subset scores $0.5104$; these two numbers bracket every
result below.

\subsection{Robustness to correlated curve-estimation error}
\label{app:corr_noise}

The i.i.d.\ noise test of Appendix~\ref{app:sensitivity} perturbs each
parameter independently, which is the mildest form of estimation error:
independent perturbations partially cancel when the solver compares phases
against one another. A controller that systematically misjudges a task --
reading it as harder than it is, and inflating every phase's requirement
together -- produces \emph{correlated} error, which does not cancel and is
the more realistic failure mode. We therefore repeat the noise test with a
shared latent component. For each run we draw one latent perturbation
$z \sim \mathcal{N}(0,1)$ shared across all $(\text{phase},
\text{parameter})$ pairs, plus an independent component
$z_i \sim \mathcal{N}(0,1)$ per parameter, and perturb each parameter
multiplicatively by $x_i' = x_i\,(1 + \sigma\,\varepsilon_i)$ with
$\varepsilon_i = \rho\,z + \sqrt{1 - \rho^{2}}\,z_i$ and $\rho = 0.7$,
clipping $a_i$ to $[0.01, 1]$ and $b_i$ to $[0.01, \infty)$. The base curves
are the per-task mean of the live controller's $15$ \texttt{zebra-additive}
estimates, so the $\sigma = 0$ row is an averaged-curve condition rather than
the live per-run curves used in Appendix~\ref{app:shuffle}. Results are in
Table~\ref{tab:corr_noise}.

\begin{table}[h]\chgon 
\centering
\small
\begin{tabular}{l c c c}
\toprule
Condition & Mean score & $\Delta$ vs.\ $\sigma = 0$ & $p$ vs.\ $\sigma = 0$ \\
\midrule
$\sigma = 0$ (clean curves)  & $0.5379$ & ---       & --- (baseline) \\
$\sigma = 0.10$              & $0.5333$ & $-0.0046$ & $0.34$\textsuperscript{ns} \\
$\sigma = 0.50$              & $0.4883$ & $-0.0496$ & $2 \times 10^{-4}$\textsuperscript{***} \\
\midrule
LLM-direct                   & $0.5104$ & $-0.0275$ & $0.0086$\textsuperscript{**} \\
\bottomrule
\end{tabular}
\caption{\textbf{ZEBRA degrades gracefully under correlated curve noise.}
Shared-latent correlated noise ($\rho = 0.7$) injected into the controller's
curves, on $60$ tier-balanced APPS tasks at $\alpha = 0.5$ ($15$ runs per
cell; $5$ for $\sigma = 0$, whose curves are identical across runs).
$p$-values are paired Wilcoxon signed-rank, two-sided, against the
clean-curve condition. At $\sigma = 0.10$ the degradation is negligible and
ZEBRA still beats LLM-direct significantly ($+0.0229$, $p = 0.012$); only at
$\sigma = 0.50$ -- where correlated error can systematically halve or double
every parameter at once -- does it drop to LLM-direct's level, and even then
the gap is not significant ($-0.0221$, $p = 0.060$).}
\label{tab:corr_noise}
\end{table}

\paragraph{Reading.} The method tolerates realistic correlated error before
it loses its edge. At a modest correlated noise level ($\sigma = 0.10$) the
drop from $0.5379$ to $0.5333$ is negligible and not statistically
significant ($p = 0.34$), and the allocation still beats LLM-direct by
$+0.0229$ ($p = 0.012$). Only under heavy correlated corruption
($\sigma = 0.50$, $0.4883$; $-0.0496$, $p = 2 \times 10^{-4}$) does
performance fall significantly below the clean-curve condition -- and even
there it lands level with LLM-direct rather than clearly under it
($-0.0221$, $p = 0.060$). The curves must be roughly right, but they need
not be accurate.

\subsection{Curve-shuffle controls: do the curves carry task-specific signal?}
\label{app:shuffle}

Noise robustness alone leaves open the opposite worry: if the allocation
survives substantial perturbation, perhaps the per-task curves are doing no
work at all, and the win comes entirely from a correct structural prior about
the pipeline. Noise cannot settle this, because adding noise degrades
information without revealing how much of it was being used. We therefore run
two \emph{shuffle controls}, each of which destroys one specific kind of
curve information while leaving the pipeline, the budget, and the curve
\emph{family} untouched:

\begin{itemize}[leftmargin=1.5em,itemsep=2pt,topsep=2pt]
\item \texttt{cross\_shuf}: tasks are paired, and each task is run with its
  partner's curves (a one-to-one mapping applied per run across the $15$
  runs). The curve shapes remain plausible pipeline curves, but they are no
  longer \emph{this task's} curves. This destroys per-task calibration while
  preserving per-phase structure.
\item \texttt{within\_shuf}: a task keeps its own curves, but the assignment
  of curves to phases is permuted by a fixed reversal
  (\texttt{plan} $\leftrightarrow$ \texttt{implement},
  \texttt{decompose} $\leftrightarrow$ \texttt{refine}). The multiset of
  curves is exactly right; only the knowledge of which curve belongs to
  which phase is destroyed.
\end{itemize}

If the curves were only a structural prior, neither shuffle would cost
anything measurable: the solver would still receive plausible saturating
curves and still concentrate budget in roughly the same place. Both in fact
hurt, and the damage is ordered (Table~\ref{tab:shuffle}).

\begin{table}[h]\chgon 
\centering
\small
\begin{tabular}{l c c c}
\toprule
Condition & Mean (all $60$) & $\Delta$ vs.\ own & $p$ (two-sided) \\
\midrule
\texttt{own} (task's own curves)        & $0.5427$ & ---       & ---     \\
\texttt{cross\_shuf} (partner's curves) & $0.5268$ & $-0.0158$ & $0.018$\textsuperscript{*} \\
\texttt{within\_shuf} (phases permuted) & $0.5166$ & $-0.0261$ & $2 \times 10^{-4}$\textsuperscript{***} \\
\midrule
LLM-direct                              & $0.5104$ & $-0.0323$ & $<10^{-4}$\textsuperscript{***} \\
\bottomrule
\end{tabular}
\caption{\textbf{Curve-shuffle controls: the estimated curves carry
task-specific and phase-specific signal.} Same $60$ tier-balanced APPS tasks,
$\alpha = 0.5$, $15$ runs per cell. $p$-values are paired Wilcoxon
signed-rank against the \texttt{own} condition. Both shuffles cost a
statistically significant amount, and the damage is ordered
\texttt{own} $>$ \texttt{cross\_shuf} $>$ \texttt{within\_shuf} $>$
LLM-direct.}
\label{tab:shuffle}
\end{table}

\paragraph{Reading.} Two facts follow. First, corrupting \emph{per-task}
calibration already costs a significant amount ($-0.0158$, $p = 0.018$), so
the estimated curves are not interchangeable across tasks -- the controller
is reading something about the specific task, not emitting a template.
Second, corrupting \emph{per-phase} structure costs substantially more
($-0.0261$, $p = 2 \times 10^{-4}$). The two shuffles  bracket ZEBRA's margin over LLM-direct. That
margin is highly significant with the task's own curves
($+0.0323$, $p < 10^{-4}$), but survives neither perturbation as a
significant effect: \texttt{cross\_shuf} keeps only a numerical edge
($+0.0164$, $p = 0.23$) and \texttt{within\_shuf} is indistinguishable from
LLM-direct ($+0.0062$, $p = 0.26$). ZEBRA's measured advantage therefore
rests on the task- and phase-specific content of the estimated curves, not
on the structural prior that a saturating-curve knapsack supplies on its
own.

\subsection{What the evidence establishes, and what it does not}
\label{app:curve_validation_summary}

Read together, the two experiments place ZEBRA between the two explanations a
skeptical reader might offer. A noisy signal survives ($\sigma = 0.10$ is not
significantly worse than clean), so the controller need not estimate curves
accurately; but the per-task and per-phase differences do carry real signal,
since removing them measurably hurts. ZEBRA therefore occupies a deliberate
middle ground and degrades gracefully: an imperfect, noisy, zero-shot
estimate still drives better allocation than asking an LLM for the split
directly, which is precisely the design intent -- turn a cheap, imperfect,
task-conditioned signal into an optimization.

We note the limits of this evidence. Quantifying the controller's estimation
\emph{bias} and its cross-phase error \emph{correlation} directly is not
possible in our setting, because phase quality is never observed in
isolation: only the end-to-end score is measurable, so there is no
phase-level ground truth to regress against. The one partial exception is the
\texttt{refine} phase, which the calibration analysis of
Appendix~\ref{app:curve_calibration} does profile -- and even there the
measurement depends on how the other phases were budgeted, so it is evidence
that the chosen curves are sensible rather than an exact target. The shuffle
controls above were designed precisely to work around this limitation: they
establish that the curves carry genuine per-task signal without requiring a
ground-truth curve to compare against.
\endgroup

\section{HotpotQA Generality Experiment}
\label{app:hotpotqa}

\begin{table*}[t]
\centering
\small
\setlength{\tabcolsep}{4pt}
\renewcommand{\arraystretch}{1.1}
\begin{tabularx}{\textwidth}{@{} l l
  >{\centering\arraybackslash}X
  >{\centering\arraybackslash}X
  >{\centering\arraybackslash}X
  >{\centering\arraybackslash}X
  >{\centering\arraybackslash}X @{}}
\toprule
\textbf{Tier} & \textbf{Strategy} &
\textbf{Mean F1} &
\textbf{Cost (\$)} &
\textbf{NB retention} &
\textbf{$p$ vs LLM} &
\textbf{$p$ vs Uniform} \\
\midrule

\multirow{4}{*}{\textbf{Overall} ($n=114$)}
& zebra-additive       & \chg{$0.4737$          }& \chg{$0.000383$ }& \chg{$92.4\%$          }& \chg{$0.0012$}\textsuperscript{\chg{**}} & \chg{$0.28$}\textsuperscript{ns} \\
& zebra-mult\_offset   & \chg{$\mathbf{0.4801}$ }& \chg{$0.000384$ }& \chg{$\mathbf{92.6\%}$ }& \chg{$0.0036$}\textsuperscript{\chg{**}} & \chg{$0.67$}\textsuperscript{ns} \\
& uniform              & \chg{$0.4713$          }& \chg{$0.000263$ }& \chg{$92.3\%$          }& \chg{$0.0023$}\textsuperscript{\chg{**}} & ---                        \\
& llm                  & \chg{$0.4302$          }& \chg{$0.000226$ }& \chg{$82.9\%$          }& ---                          & \chg{$0.0023$}\textsuperscript{\chg{**}} \\
\midrule

\multirow{4}{*}{\textbf{Easy} ($\bar{s}\!\ge\!0.5$, $n=66$)}
& zebra-additive       & \chg{$0.5917$          }& \chg{$0.000378$ }& \chg{$91.6\%$          }& $0.011$\textsuperscript{*}   & \chg{$0.29$}\textsuperscript{ns} \\
& zebra-mult\_offset   & \chg{$\mathbf{0.6067}$ }& \chg{$0.000380$ }& \chg{$\mathbf{93.8\%}$ }& \chg{$0.012$\textsuperscript{*}   }& \chg{$0.29$}\textsuperscript{ns} \\
& uniform              & \chg{$0.5869$          }& \chg{$0.000257$ }& \chg{$91.1\%$          }& \chg{$0.0080$}\textsuperscript{\chg{**}} & ---                        \\
& llm                  & \chg{$0.5480$          }& \chg{$0.000213$ }& \chg{$85.1\%$          }& ---                          & \chg{$0.0080$}\textsuperscript{\chg{**}} \\
\midrule

\multirow{4}{*}{\textbf{Hard} ($\bar{s}\!<\!0.5$, $n=48$)}
& zebra-additive       & \chg{$0.3114$          }& \chg{$0.000390$ }& \chg{$93.5\%$          }& \chg{$0.040$}\textsuperscript{\chg{*}}   & \chg{$0.58$}\textsuperscript{ns} \\
& zebra-mult\_offset   & \chg{$0.3060$          }& \chg{$0.000389$ }& \chg{$91.0\%$          }& \chg{$0.12$\textsuperscript{ns}   }& $0.62$\textsuperscript{ns} \\
& uniform              & \chg{$\mathbf{0.3124}$ }& \chg{$0.000271$ }& \chg{$\mathbf{94.0\%}$ }& \chg{$0.099$}\textsuperscript{\chg{ns}}  & ---                        \\
& llm                  & \chg{$0.2684$          }& \chg{$0.000244$ }& \chg{$79.9\%$          }& ---                          & \chg{$0.099$}\textsuperscript{\chg{ns}} \\

\bottomrule
\end{tabularx}
\caption{\textbf{HotpotQA generality experiment} ($\alpha = 0.5$, $15$
runs per cell, \texttt{gpt-4o-mini}). NB retention is computed as a
mean of per-task ratios, $\mathrm{mean}_i(\mathrm{F1}_i / \mathrm{F1}_i^{\text{NB}})$,
averaged over tasks with $\mathrm{F1}_i^{\text{NB}} \ge 0.01$. Reference no-budget means:
$\overline{\mathrm{F1}}_{\text{overall}} = 0.514$,
\chg{$\overline{\mathrm{F1}}_{\text{easy}} = 0.646$},
\chg{$\overline{\mathrm{F1}}_{\text{hard}} = 0.333$}. Tasks are split at the median
no-budget F1. \textbf{Both ZEBRA variants
significantly beat LLM-direct} (\chg{$p < 0.01$ }overall) and \textbf{tie
Uniform} (\chg{$p > 0.25$}): the optimal split on this benchmark is close to
balanced and ZEBRA recovers that, while LLM-direct does not (cf.\
Table~\ref{tab:hotpotqa_alloc}). Significance markers:
$^{*}p<0.05$, $^{**}p<0.01$, $^{***}p<0.001$, ns~=~not significant.}
\label{tab:hotpotqa_overall}
\end{table*}

This appendix details the HotpotQA generality experiment summarized
in Section~\ref{sec:results_hotpotqa}. The goals are to (i) move
ZEBRA out of the coding domain, (ii) use a different number of
phases, and (iii) use a different orchestration framework -- so that
any positive transfer cannot be attributed to a coding- or
LangGraph-specific artefact.

\subsection{Dataset and task selection}
\label{app:hotpotqa_dataset}

We use the HotpotQA distractor validation split as a source of multi-hop questions with short gold answers and difficulty labels; the pipeline answers from the model's parametric knowledge and does not consume the distractor paragraphs.
We sample \chg{$400$ }questions
deterministically with seed $= 42$ via
\texttt{extract\_hotpotqa\_tasks.py}. Each task is stored as a JSON
file with fields \texttt{\{question, answer, type, level,
supporting\_facts\}}. Only question and answer are used. 

\paragraph{Screening.}
All \chg{$400$ }tasks are screened with the no-budget pipeline ($15$ runs
each, \texttt{gpt-4o-mini}, max-workers $= 8$). Per task we record
the mean, median, and standard deviation of cost (USD) and of F1
score across the $15$ runs.

\paragraph{Filtering.} Tasks are retained if they satisfy both:
\begin{itemize}\itemsep=2pt
\item Non-trivial score range: $0.20 \le \overline{\mathrm{F1}}_{\text{NB}} \le 0.85$.
Below $0.2$ the model essentially cannot answer regardless of
budget; above $0.85$ the model essentially always answers correctly,
leaving no headroom for allocation to matter.
\item Cost stability: $\mathrm{CV}_{\text{cost}} < 0.35$ (where
$\mathrm{CV} = \mathrm{stdev}/\mathrm{median}$). This ensures the
per-task budget reference $\bar{c}$ is meaningful.
\end{itemize}
\chg{$114$ of $400$ }tasks pass both filters. Splitting at the median
no-budget F1 ($0.5$) gives \chg{$66$ }\emph{easy} ($\overline{\mathrm{F1}} \in [0.5, 0.85]$)
and \chg{$48$ }\emph{hard} ($\overline{\mathrm{F1}} \in [0.2, 0.5)$) tasks.

\subsection{Pipeline}
\label{app:hotpotqa_pipeline}

The HotpotQA pipeline is intentionally different from the APPS
pipeline. It uses \emph{three} phases in plain Python (no LangGraph),
chained as \texttt{direct} $\to$ \texttt{research} $\to$
\texttt{synthesize}:

\begin{description}
\itemsep=2pt
\item[direct] One-shot answer to the multi-hop question. Produces a
candidate answer.
\item[research] Decomposes the question into $2$--$4$ sub-questions
and answers each with a specific fact. Produces \emph{evidence
only} -- not a final answer.
\item[synthesize] Combines the \texttt{direct} answer with the
\texttt{research} evidence; explicitly instructed to keep the
\texttt{direct} answer unless the evidence contradicts it. Produces
a final candidate answer.
\end{description}

Each phase supports iterative refinement within its allocated
budget (initial generation; subsequent iterations may refine or
output \texttt{NO\_CHANGE}; the phase terminates on budget
exhaustion, two consecutive no-changes, or a maximum-iteration
cap). Budget enforcement uses the same \texttt{BudgetEnforcedLLM}
interface as APPS. The pipeline keeps the highest-F1 candidate
across phases, ensuring monotonicity in number of phases. \chg{As on APPS,
generation is blind to the answer: the gold answer is never shown to any
phase, F1 is computed only after the run, and the same best-candidate rule is
applied identically to every allocator.
}

\subsection{Strategies}
\label{app:hotpotqa_strategies}

We compare four strategies, all sharing the same pipeline and
controller model (\texttt{gpt-4o-mini}):
\begin{itemize}\itemsep=2pt
\item \textbf{zebra-additive} -- ZEBRA's controller estimates
$(a_i, b_i)$ for each of the three QA phases; allocations are
solved with the additive knapsack of Section~\ref{sec:additive}
(\texttt{solver.py} unchanged from APPS).
\item \textbf{zebra-mult\_offset} -- same controller, the
multiplicative-offset solver of Section~\ref{sec:mult_offset}
(\texttt{solver.py} unchanged).
\item \textbf{llm} -- LLM-direct: the controller, given the same
pipeline description and total budget, outputs per-phase USD
allocations as JSON; the allocations are scaled to the total budget.
\item \textbf{uniform} -- equal split, $B/3$ per phase.
\end{itemize}
A no-budget run (unconstrained pipeline) provides the oracle
ceiling and the per-task budget reference.

\subsection{Budget protocol and metrics}
\label{app:hotpotqa_protocol}

Per task we set $B = \alpha \cdot \tilde{c}_{\text{NB}}$ with
$\alpha = 0.5$ and $\tilde{c}_{\text{NB}}$ the per-task median
no-budget cost. Each (task, strategy) cell is run for $15$
independent runs.

We use the standard HotpotQA~\citep{yang2018hotpotqa} /
SQuAD~\citep{rajpurkar2016squad} evaluation: predictions and
gold answers are normalized (lowercase, strip articles, strip
punctuation, collapse whitespace) and scored by token-level F1
between the resulting tokens. Token F1 is the primary metric
(more granular than exact match for short factoid answers).

\subsection{Results}
\label{app:hotpotqa_results}

The headline results are reported in
Table~\ref{tab:hotpotqa_overall} (overall + per tier) and in
Table~\ref{tab:hotpotqa_alloc} (per-phase allocations).

\begin{table}[h]
\centering
\small
\setlength{\tabcolsep}{5pt}
\renewcommand{\arraystretch}{1.1}
\begin{tabular}{l l c c c c}
\toprule
\textbf{Tier} & \textbf{Strategy} &
\textbf{direct} & \textbf{research} & \textbf{synthesize} &
\textbf{Avg.\ spent (\$)} \\
\midrule
\multirow{4}{*}{Overall ($n=114$)}
& zebra-additive     & \chg{$30.6\%$ }& \chg{$39.1\%$ }& \chg{$30.3\%$ }& \chg{$0.000383$ }\\
& zebra-mult\_offset & \chg{$29.5\%$ }& \chg{$42.4\%$ }& \chg{$28.1\%$ }& \chg{$0.000384$ }\\
& uniform            & $33.3\%$ & $33.3\%$ & $33.3\%$ & \chg{$0.000263$ }\\
& llm                & \chg{$67.4\%$ }& \chg{$21.2\%$ }& \chg{$11.4\%$ }& \chg{$0.000226$ }\\
\midrule
\multirow{4}{*}{Easy ($n=66$)}
& zebra-additive     & \chg{$31.0\%$ }& \chg{$38.7\%$ }& \chg{$30.3\%$ }& \chg{$0.000378$ }\\
& zebra-mult\_offset & \chg{$29.9\%$ }& $42.2\%$ & \chg{$28.0\%$ }& \chg{$0.000380$ }\\
& uniform            & $33.3\%$ & $33.3\%$ & $33.3\%$ & \chg{$0.000257$ }\\
& llm                & \chg{$69.7\%$ }& \chg{$20.3\%$ }& \chg{$9.9\%$  }& \chg{$0.000213$ }\\
\midrule
\multirow{4}{*}{Hard ($n=48$)}
& zebra-additive     & \chg{$30.0\%$ }& \chg{$39.7\%$ }& \chg{$30.2\%$ }& \chg{$0.000390$ }\\
& zebra-mult\_offset & \chg{$29.0\%$ }& \chg{$42.6\%$ }& $28.4\%$ & \chg{$0.000389$ }\\
& uniform            & $33.3\%$ & $33.3\%$ & $33.3\%$ & \chg{$0.000271$ }\\
& llm                & \chg{$64.1\%$ }& \chg{$22.4\%$ }& \chg{$13.5\%$ }& \chg{$0.000244$ }\\
\bottomrule
\end{tabular}
\caption{\textbf{HotpotQA per-phase budget shares} (\%) and
realized spend, averaged across runs and tasks. \emph{Both ZEBRA
variants converge to a near-balanced allocation} ($\sim 30/40/30$),
within a few points of Uniform.}
\label{tab:hotpotqa_alloc}
\end{table}

\paragraph{ZEBRA significantly beats LLM-direct.} Overall,
\texttt{zebra-additive} retains \chg{$92.4\%$ }of the no-budget F1
versus \chg{$82.9\%$ }for LLM-direct -- a \chg{$+9.5$}pp gap on NB retention
(equivalently \chg{$+0.043$ }on absolute mean F1: \chg{$0.474$ }vs.\ \chg{$0.430$;
$p = 0.0012$ }paired Wilcoxon on per-task mean F1,
\chg{$n = 114$}). The gap holds in both tiers: \chg{$+6.5$}pp NB retention on
easy tasks ($p = 0.011$), and widens to \chg{$+13.6$}pp NB retention on
the hard tier (\chg{$p = 0.040$}). Uniform also beats LLM-direct
(\chg{$+9.4$}pp NB retention overall, \chg{$p = 0.0023$}), so the issue is
squarely with the LLM controller's allocation, not with the
pipeline itself.

\paragraph{ZEBRA and Uniform are statistically tied.}
All differences below are NB-retention gaps in percentage points
(Wilcoxon tests are computed on per-task mean F1).
\texttt{zebra-additive} vs.\ Uniform: \chg{$+0.05$}pp overall
(\chg{$p = 0.28$), $+0.5$}pp easy (\chg{$p = 0.29$), $-0.5$}pp hard
(\chg{$p = 0.58$}). \texttt{zebra-mult\_offset} vs.\ Uniform:
\chg{$+0.3$}pp overall (\chg{$p = 0.67$}), and similarly non-significant per
tier. None of the comparisons reach significance at $\alpha = 0.05$.
For reference, the corresponding absolute-F1 differences are
small: \texttt{zebra-additive} vs.\ Uniform is \chg{$+0.002$ overall,
$+0.005$ easy, $-0.001$ }hard.

\paragraph{Why Uniform is so competitive here.} The LLM-direct
allocation is sharply skewed -- \chg{$\sim 67\%$ }on \texttt{direct},
\chg{$\sim 21\%$ }on \texttt{research}, \chg{$\sim 11\%$ }on \texttt{synthesize}
-- and is consistently dominated. The successful strategies all
land near a balanced split: ZEBRA-additive averages \chg{$31/39/30$
}and ZEBRA-mult\_offset averages \chg{$30/42/28$}, both within $\sim 10$pp
of the $33/33/33$ uniform reference (Table~\ref{tab:hotpotqa_alloc}).
On these QA tasks the three phases contribute comparably to
end-to-end F1, so any near-balanced split lands close to the
optimum. ZEBRA does not beat Uniform because there is essentially
no allocation gap to exploit; the gap to LLM-direct, on the other
hand, is large because LLM-direct is qualitatively miscalibrated.

\paragraph{This is the generality result.} On APPS, the same ZEBRA
controllers learn a sharply skewed allocation ($\sim 11/14/24/51\%$
across plan/decompose/implement/refine,
Table~\ref{tab:allocation_distribution}); on HotpotQA they converge
to a balanced one. The framework adapts the \emph{shape} of the
allocation to what the task structure demands. On
benchmarks where balance happens to be optimal, the controller
recovers near-balance and matches Uniform, while still beating an
LLM that allocates by intuition.

\paragraph{ZEBRA's allocation is also consistent across tiers on
HotpotQA, in contrast to APPS.} A second axis of adaptation is
within-benchmark: do per-tier allocations differ? On APPS they do
-- \texttt{refine} alone moves by $\sim$$15$--$30$pp between easy
and medium/hard at $\alpha = 0.5$
(Sec.~\ref{sec:results_allocations},
Table~\ref{tab:allocation_distribution}). On HotpotQA they do not:
\texttt{zebra-additive} allocates \chg{$31.0 / 38.7 / 30.3$ }on easy
versus \chg{$30.0 / 39.7 / 30.2$ }on hard (per-phase shifts of
\chg{$0.9$, $1.0$, $0.1$}~pp respectively); \texttt{zebra-mult\_offset}
behaves the same way (per-tier shifts \chg{$\le 0.9$}~pp on every phase).
The per-task standard deviation of each allocation share is also
small (\chg{$1.5$}--\chg{$4.0$~pp, against $13$--$30$~pp for LLM-direct}), so the
tier-flat average is not hiding bimodal behavior. The reading we draw is: \emph{ZEBRA's per-tier adaptation
is itself task-dependent}. On APPS the difficulty signal moves the
controller's curves and the optimal allocation correspondingly
shifts; on HotpotQA the difficulty signal does not change which
phase is the bottleneck, the controller's curves move
proportionally across tiers, and the allocation stays put. ``Adapt
when adaptation helps; do not adapt when it does not'' is a
property of the controller.

\subsection{What does \emph{not} carry over from APPS}
\label{app:hotpotqa_caveats}

Two parts of the APPS-side story do not reproduce here:
\begin{itemize}\itemsep=2pt
\item \emph{ZEBRA does not beat Uniform on this benchmark.} Uniform
is the right answer here; ZEBRA \emph{is} Uniform, just discovered
zero-shot. We frame this as a feature (the framework adapts) rather
than a strict win, and we state it plainly in the body
(Section~\ref{sec:results_hotpotqa}).
\item \emph{The mult\_offset variant does not lift the hard-tier
mean above additive.} On APPS, \texttt{mult\_offset} is the
strongest variant on hard tasks because the product form rewards
funding the weakest link; at HotpotQA's near-balanced optimum the
two objectives produce nearly identical allocations and similar
scores \chg{on the hard tier ($0.311$ for }\texttt{\chg{additive}} vs.\ \chg{$0.306$ for
}\texttt{\chg{mult\_offset}}\chg{), with }\texttt{\chg{mult\_offset}} \chg{slightly ahead overall
($0.480$ vs.\ $0.474$}).
\end{itemize}
The qualitative story -- ZEBRA dominates LLM-direct, and the
allocation it learns is task-appropriate -- carries over cleanly.

\section{Hybrid (Mid-Pipeline) Re-allocation}
\label{app:hybrid}

ZEBRA as presented in the main paper allocates the budget once,
before execution. A natural extension is to add an
\emph{online} re-allocation step that, partway through the pipeline,
re-estimates the remaining phases' utility curves conditioned on what
has already been produced and re-solves the water-filling problem on
the leftover budget. We implemented and evaluated this hybrid variant
on the same APPS pipeline used in the main paper. The result is
negative: hybrid re-allocation does not improve over one-shot ZEBRA in
this setting. We document the experiment here for completeness and to
inform future work in longer or more uncertain pipelines.

\subsection{Hybrid pipeline}
\label{app:hybrid_pipeline}

The static ZEBRA pipeline,
\[
\texttt{allocate} \to \texttt{plan} \to \texttt{decompose} \to
\texttt{implement} \to \texttt{refine} \to \text{END},
\]
is augmented with two new nodes inserted after \texttt{decompose}:
\[
\texttt{allocate} \to \texttt{plan} \to \texttt{decompose} \to
\texttt{judge} \to \texttt{reallocate} \to \texttt{implement} \to
\texttt{refine} \to \text{END}.
\]

\paragraph{\texttt{judge} node.} The same controller LLM
(\texttt{gpt-4o}) is shown the original task and the partial output
(\texttt{plan} + \texttt{decompose}) and asked to rate the
completeness and correctness of the decomposition on a $1$--$5$ scale
with a one-sentence justification (returned as JSON).

\paragraph{\texttt{reallocate} node.} Given the original task, the
partial output, the already-spent budget
$s = s_{\texttt{plan}} + s_{\texttt{decompose}}$, the total budget
$B$, and (optionally) the judge score and reason, the controller
re-estimates the two operating points
$(\textsf{tokens\_basic},\textsf{tokens\_great})$ for the remaining
phases \texttt{implement} and \texttt{refine}. The same two-point
fitting procedure (Section~\ref{sec:curves}, full derivation in
Appendix~\ref{app:method_twopoint}) converts these into
$(a_i, b_i)$, and the same water-filling solver (Section~\ref{sec:method})
is applied to the remaining budget $B' = B - s$ over
$\{\texttt{implement}, \texttt{refine}\}$, yielding new caps
$(x'_{\texttt{implement}}, x'_{\texttt{refine}})$ that replace the
existing caps in pipeline state.

\subsection{Strategies}
\label{app:hybrid_strategies}

We compared three cells:

\begin{itemize}
\item \textbf{\texttt{zebra-additive} (baseline).} Static, one-shot
  ZEBRA with the additive aggregation, as used in the main paper.
\item \textbf{\texttt{hybrid-no-judge}.} Pre-allocates exactly as
  static ZEBRA, then runs \texttt{reallocate} after \texttt{decompose}
  \emph{without} a judge score.
\item \textbf{\texttt{hybrid-with-judge}.} Same as
  \texttt{hybrid-no-judge}, but the \texttt{reallocate} prompt also
  receives the judge score and reason.
\end{itemize}

All three share the task list, 
the controller and base-phase LLMs, and the evaluation harness. The
only differences are the inserted nodes and the resulting per-phase caps.

\subsection{Setup}
\label{app:hybrid_setup}

We used the same $150$ APPS interview-level tasks ($50$ easy +
$50$ medium + $50$ hard) and the budget level $\alpha = 0.5$. Each
hybrid strategy was run $10$ times per task; the static baseline was
run $15$ times per task (the existing main-paper budget). All other
hyperparameters, prompts, and scoring procedures match the main paper
exactly.

\subsection{Results}
\label{app:hybrid_results}

\begin{table}[h]
\centering
\small
\begin{tabular}{l c c c}
\toprule
Strategy & Mean & Median & Success \% \\
\midrule
\texttt{zebra-additive} (baseline) & $0.526$ & $0.533$ & $52.7\%$ \\
\texttt{hybrid-no-judge}           & $0.501$ & $0.530$ & $49.9\%$ \\
\texttt{hybrid-with-judge}         & $0.525$ & $0.516$ & $52.2\%$ \\
\bottomrule
\end{tabular}
\caption{Hybrid re-allocation versus one-shot ZEBRA on the
$150$-task APPS benchmark at $\alpha = 0.5$. No pairwise comparison
reaches significance 
(all $p_{\text{BH}} > 0.15$).}
\label{tab:hybrid_results}
\end{table}

Neither hybrid variant improves over the static baseline.
\texttt{hybrid-with-judge} matches the baseline within noise;
\texttt{hybrid-no-judge} is numerically slightly worse.

\subsection{Why hybrid does not help here}
\label{app:hybrid_discussion}

In the APPS pipeline, \texttt{allocate} is followed by only two
substantive downstream phases (\texttt{implement} and \texttt{refine})
once \texttt{plan} and \texttt{decompose} have run. With only two
phases left to split a residual budget across, the curve estimates
produced before execution already capture task difficulty well enough. Online re-allocation is more likely to
pay off in pipelines with (i) more downstream phases over which to
redistribute the residual budget, (ii) higher curve-estimation
uncertainty up front, or (iii) intermediate outputs whose quality
varies substantially across runs in ways the controller cannot
anticipate from the task description alone. We did not observe these
conditions in the APPS pipeline, and report the negative result as
evidence that one-shot allocation is sufficient in this regime.

\section{Examples}
\label{app:examples}
\label{app:example}

This appendix walks through a single APPS task end-to-end so the
reader can see exactly what each strategy is shown, what each strategy
outputs, and how the resulting allocation translates into a per-task
quality score. We pick a short, easily-understood task on which both
ZEBRA variants beat every LLM-based allocator by a wide margin, so
that the mechanism behind the gain is visible without distraction.
The appendix is organized as follows:
App.~\ref{app:example:task} introduces the chosen task
(\texttt{apps\_12}).
App.~\ref{app:example:outcome} reports per-strategy scores at
$\alpha = 0.5$ across $15$ seeds;
App.~\ref{app:example:allocs} reports the corresponding mean
per-phase allocations;
App.~\ref{app:example:prompts} reproduces all four allocator prompts
verbatim;
App.~\ref{app:example:execution} traces what happens during execution
seed-by-seed and isolates the mechanism that distinguishes ZEBRA from
the LLM-based allocators.

\subsection{The task: \texttt{apps\_12}}
\label{app:example:task}

\texttt{apps\_12} is a Codeforces-style problem from the APPS interview
set. The generator must implement
\verb|solve(stdin_input: str) -> str| for the following specification:

\begin{quote}\small
Vova has $n$ trophies in a row, each either golden (\texttt{G}) or silver
(\texttt{S}). He may perform \emph{at most one swap} of any two trophies.
Output the maximum length of a contiguous \texttt{G}-segment achievable
after that swap. ($2 \le n \le 10^{5}$.)
\end{quote}

We run all strategies at the main pressure regime $\alpha = 0.5$, which
gives a per-task budget of $B \approx \$0.01795$ (this task has an
unconstrained mean cost of $\bar{c} \approx \$0.0359$). Each cell is
repeated for $n = 15$ seeds. Base phases use
\texttt{gpt-4o-mini-2024-07-18}; refine and the controller use
\texttt{gpt-4o-2024-08-06}, which is $\sim$$16.7\times$ more expensive
per call than the base model. The task has $182$ tests.

\subsection{Per-strategy outcome}
\label{app:example:outcome}

\begin{table}[h]
\centering
\small
\begin{tabular}{lcccc}
\toprule
Strategy & Mean score & Success/15 & Mean phase spend & vs.\ LLM-Direct \\
\midrule
LLM-Direct                  & $0.327$ & $3/15$  & \$0.01339 & ---       \\
LLM-CoT                     & $0.296$ & $3/15$  & \$0.01051 & $-0.031$  \\
ZEBRA-LLM (ablation)        & $0.314$ & $2/15$  & \$0.01120 & $-0.013$  \\
Uniform (reference)         & $0.336$ & $4/15$  & \$0.00884 & $+0.009$  \\
\textbf{ZEBRA-additive}     & \boldmath$0.410$ & \boldmath$6/15$  & \boldmath\$0.01442 & \boldmath$+0.083$ \\
\textbf{ZEBRA-mult\_offset} & \boldmath$0.467$ & \boldmath$9/15$  & \boldmath\$0.01477 & \boldmath$+0.140$ \\
\bottomrule
\end{tabular}
\caption{Per-task summary on \texttt{apps\_12} at $\alpha = 0.5$ over
$15$ seeds. The per-task budget is $B \approx \$0.01795$; the ``Mean
phase spend'' column reports realised phase-only spend, which is bounded
by $B$ (the controller's own \texttt{gpt-4o} allocation call is billed
separately).}
\label{tab:apps12-summary}
\end{table}

\subsection{Mean allocations across 15 seeds}
\label{app:example:allocs}

The six allocators place very different bets given the same budget
$B \approx \$0.01795$. Table~\ref{tab:apps12-alloc} reports the mean
($\pm$ std) per-phase allocation across the $15$ seeds, so the pattern
reflects the strategy's typical behaviour rather than any one
realisation:

\begin{table}[h]
\centering
\footnotesize
\begin{tabular}{lcccccc}
\toprule
& \multicolumn{4}{c}{\textbf{Allocated} ($\times 10^{-3}$)}
& \textbf{Realized} ($\times 10^{-3}$) & \\
\cmidrule(lr){2-5} \cmidrule(lr){6-6}
Strategy & plan & decompose & implement & refine & Phase spend & Score \\
\midrule
Uniform                      & $4.49_{\pm 0.00}$ & $4.49_{\pm 0.00}$ & $4.49_{\pm 0.00}$ & $4.49_{\pm 0.00}$ & $8.84_{\pm 0.39}$  & $0.336$ \\
LLM-Direct                   & $2.00_{\pm 0.00}$ & $2.00_{\pm 0.00}$ & $4.87_{\pm 0.62}$ & $9.08_{\pm 0.62}$  & $13.39_{\pm 0.95}$ & $0.327$ \\
LLM-CoT                      & $1.93_{\pm 0.25}$ & $2.60_{\pm 0.61}$ & $6.73_{\pm 1.61}$ & $6.68_{\pm 1.91}$  & $10.51_{\pm 3.32}$ & $0.296$ \\
ZEBRA-LLM (ablation)         & $2.73_{\pm 0.68}$ & $2.37_{\pm 0.46}$ & $5.57_{\pm 0.96}$ & $7.28_{\pm 1.89}$  & $11.20_{\pm 2.80}$ & $0.314$ \\
\textbf{ZEBRA-additive}      & $1.09_{\pm 0.09}$ & $1.36_{\pm 0.09}$ & $2.47_{\pm 0.12}$ & $\mathbf{13.03}_{\pm 0.24}$ & $14.42_{\pm 1.50}$ & $0.410$ \\
\textbf{ZEBRA-mult\_offset}  & $1.07_{\pm 0.09}$ & $1.34_{\pm 0.08}$ & $2.40_{\pm 0.20}$ & $\mathbf{13.14}_{\pm 0.24}$ & $\mathbf{14.77}_{\pm 1.40}$ & $\mathbf{0.467}$ \\
\bottomrule
\end{tabular}
\caption{Mean per-phase allocation and phase-only realised spend across
$15$ seeds, in units of $10^{-3}$ USD. Each row sums (in expectation)
to $B = \$17.95 \!\times\! 10^{-3}$; ``Phase spend'' is the realised
total and is bounded by $B$. The two ZEBRA solvers concentrate budget
on \texttt{refine} (mean alloc $\$13.03 \!\times\! 10^{-3}$ for additive
and $\$13.14 \!\times\! 10^{-3}$ for \texttt{mult\_offset}, $\sim$$73\%$
of $B$), with the smallest seed-to-seed standard deviation of any
strategy (refine $\pm 0.24$). LLM-Direct allocates $\$9.08
\!\times\! 10^{-3}$, LLM-CoT $\$6.68 \!\times\! 10^{-3}$, and the
ZEBRA-LLM ablation $\$7.28 \!\times\! 10^{-3}$ -- all between
$\frac{1}{2}$ and $\frac{2}{3}$ of ZEBRA's
refine budget. As Section~\ref{app:example:execution} shows, this gap
is exactly what determines whether refine produces a useful revision
or a regression.}
\label{tab:apps12-alloc}
\end{table}

\subsection{Prompts used in the experiments}
\label{app:example:prompts}

We include all four allocator prompts verbatim. They share the same
task description and total budget; they differ only in what they ask
for.

\paragraph{(a) ZEBRA controller prompt (\texttt{token\_twopoint}, used
by ZEBRA-additive, ZEBRA-mult\_offset, and the ZEBRA-LLM ablation).}
The controller is asked for two-point token estimates and a per-phase
criticality $a_i$, never for an allocation:

\begin{quote}\footnotesize\ttfamily\raggedright
You are a budget allocation controller for a multi-agent coding system.\\[2pt]
Task: \{task\_description\}\\
Phase model: gpt-4o-mini-2024-07-18 (input=\$0.15/1M tokens, output=\$0.6/1M tokens)\\
Total budget (USD): 0.01795\\[2pt]
The system has 4 sequential phases. All phases can iterate (refine their output with additional LLM calls) if budget allows.\\
1. plan -- understand the task and create a plan\\
2. decompose -- break the plan into implementable tasks\\
3. implement -- write the solution code\\
4. refine -- review/revise loop using gpt-4o-2024-08-06 ($\sim$17$\times$ more expensive per call than gpt-4o-mini)\\[2pt]
Carefully consider the specific task above and its difficulty. For each phase, estimate the following parameters for THIS particular task:\\[2pt]
- {\bfseries tokens\_basic} (integer, 100--10000): total output tokens this phase needs to produce basically acceptable output ($\sim$50\% of potential quality). Include all iterations.\\
- {\bfseries tokens\_great} (integer, $\geq$ tokens\_basic, up to 20000): total output tokens for near-optimal output ($\sim$90\% of potential quality). Include all iterations.\\
- {\bfseries a} (criticality, 0--1): how critical this phase is for this specific task. When budget is tight, low-a phases are the first to be cut.\\[2pt]
Respond with a JSON object in this exact format:\\
\{\{"plan": \{\{"tokens\_basic": <int>, "tokens\_great": <int>, "a": <0-1>\}\}, "decompose": \{\{...\}\}, "implement": \{\{...\}\}, "refine": \{\{...\}\}\}\}\\[2pt]
Rules:\\
- Output ONLY the JSON. No prose, no markdown, no code fences.\\
- Use numeric values only.\\
- Think carefully about how complex this specific task is and how many tokens each phase needs.\\
- tokens\_basic must be $\geq 100$. tokens\_great must be $\geq$ tokens\_basic.
\end{quote}

\paragraph{(b) LLM-Direct allocator prompt.}
The LLM is asked for a final allocation in one shot:

\begin{quote}\footnotesize\ttfamily\raggedright
You are a budget allocation controller for a multi-agent coding system.\\[2pt]
Task: \{task\_description\}\\[2pt]
Total budget (USD): 0.017951\\
Phases 1-3 model: gpt-4o-mini-2024-07-18 (input=\$0.15/1M tokens, output=\$0.6/1M tokens)\\
Refine model: gpt-4o-2024-08-06 (input=\$2.5/1M tokens, output=\$10.0/1M tokens) -- $\sim$17$\times$ more expensive per call than gpt-4o-mini-2024-07-18.\\[2pt]
The system has 4 sequential phases. All phases can iterate (refine their output with additional LLM calls) if budget allows.\\
1. plan -- understand the task and create a plan\\
2. decompose -- break the plan into implementable tasks\\
3. implement -- write the solution code\\
4. refine -- review$\to$revise loop that catches and fixes real bugs. Each iteration = 2 LLM calls. Uses gpt-4o-2024-08-06 ($\sim$17$\times$ cost per call vs gpt-4o-mini-2024-07-18). Token-heavy.\\[2pt]
Allocate the total budget across these 4 phases.\\
Output ONLY a JSON object mapping each phase to its USD allocation, with no extra text.
\end{quote}

\paragraph{(c) LLM-CoT allocator prompt.}
\label{app:examples_llm_cot}
Identical to LLM-Direct except for the final sentence, which invites
the controller to reason before emitting the JSON. The controller is
given \texttt{max\_tokens} = $800$ instead of $300$:

\begin{quote}\footnotesize\ttfamily\raggedright
You are a budget allocation controller for a multi-agent coding system.\\[2pt]
Task: \{task\_description\}\\[2pt]
Total budget (USD): 0.017951\\
Phases 1-3 model: gpt-4o-mini-2024-07-18 (input=\$0.15/1M tokens, output=\$0.6/1M tokens)\\
Refine model: gpt-4o-2024-08-06 (input=\$2.5/1M tokens, output=\$10.0/1M tokens) -- $\sim$17$\times$ more expensive per call than gpt-4o-mini-2024-07-18.\\[2pt]
The system has 4 sequential phases [\ldots same as above \ldots]\\[2pt]
\textbf{Let's think step by step about how to allocate the budget, then output a JSON object} \{"plan": ..., "decompose": ..., "implement": ..., "refine": ...\} \textbf{with values in USD summing to 0.017951.}
\end{quote}

\paragraph{(d) ZEBRA-LLM ablation allocator prompt.}
The controller call (a) is run first; its fitted utility curves are
then formatted into a prompt and handed to the LLM allocator instead
of to the knapsack solver:

\begin{quote}\footnotesize\ttfamily\raggedright
You are a budget allocation controller for a multi-agent coding system.\\[2pt]
Total budget (USD): 0.017951\\
Model: gpt-4o-mini-2024-07-18 (input=\$0.15/1M tokens, output=\$0.6/1M tokens)\\
Note: The refine phase uses gpt-4o-2024-08-06 which is $\sim$17$\times$ more expensive per call than gpt-4o-mini-2024-07-18.\\[2pt]
The system has 4 sequential phases:\\
1. plan -- understand the task and create a plan\\
2. decompose -- break the plan into implementable tasks\\
3. implement -- write the solution code\\
4. refine -- review/revise loop to fix bugs\\[2pt]
We have estimated the following utility curves for each phase. Each curve models quality $= a \cdot (1 - e^{-b\cdot \text{budget}})$, where:\\
\hspace*{1em}- a (quality ceiling): maximum quality this phase can achieve\\
\hspace*{1em}- b (saturation rate): how quickly returns diminish per dollar spent\\
\hspace*{1em}- Higher b means the phase saturates quickly (needs less budget)\\
\hspace*{1em}- Higher a means the phase is more important for overall quality\\[2pt]
Estimated curves:\\
\hspace*{1em}plan: quality\_ceiling(a)=0.800, saturation\_rate(b)=7444.31\\
\hspace*{1em}decompose: quality\_ceiling(a)=0.700, saturation\_rate(b)=4962.87\\
\hspace*{1em}implement: quality\_ceiling(a)=0.900, saturation\_rate(b)=2977.72\\
\hspace*{1em}refine: quality\_ceiling(a)=0.600, saturation\_rate(b)=1985.15\\[2pt]
Based on these utility curves, allocate the total budget across the 4 phases to maximize overall task quality. Phases are sequential -- upstream quality affects downstream phases.\\
Output ONLY a JSON object mapping each phase to its USD allocation, with no extra text.
\end{quote}

\subsection{Sample raw responses (single seed, illustrative)}
\label{app:example:responses}

The tables above report averages across $15$ seeds; in this subsection
we quote a single seed's raw output verbatim from the canonical seed-1
log of each strategy.
\paragraph{ZEBRA controller raw response.}
The ZEBRA-additive and ZEBRA-mult\_offset strategies share the same
controller call. On seed~1 the controller returns, verbatim:

\begin{quote}\footnotesize\ttfamily
\{\\
\hspace*{1em}"plan": \{"tokens\_basic": 300, "tokens\_great": 600, "a": 0.8\},\\
\hspace*{1em}"decompose": \{"tokens\_basic": 400, "tokens\_great": 800, "a": 0.7\},\\
\hspace*{1em}"implement": \{"tokens\_basic": 800, "tokens\_great": 1500, "a": 0.9\},\\
\hspace*{1em}"refine": \{"tokens\_basic": 600, "tokens\_great": 1200, "a": 0.6\}\\
\}
\end{quote}

\noindent After fitting $b_i$ via the geometric mean of the two-point
constraints (Section~\ref{sec:method}), the per-phase curves are
$(a,b)_\text{plan}=(0.80,\,4963)$,
$(a,b)_\text{decompose}=(0.70,\,3722)$,
$(a,b)_\text{implement}=(0.90,\,1922)$,
$(a,b)_\text{refine}=(0.60,\,2481)$. Refine's $b$ is then divided by
the \texttt{gpt-4o} cost ratio $16.7$, giving
$b_\text{refine} = 148.6$. The same $(a, b)$ vector is then handed to
the additive solver and the \texttt{mult\_offset} solver. 

\paragraph{LLM-Direct raw response.}
The LLM-Direct allocator returns, verbatim:

\begin{quote}\footnotesize\ttfamily
\{"plan": 0.002, "decompose": 0.002, "implement": 0.006, "refine": 0.007951\}
\end{quote}

\noindent A round-numbers split with $\sim$$44\%$ of $B$ on refine.

\paragraph{LLM-CoT raw response.}
The CoT controller (\texttt{gpt-4o}, temperature $0.3$) reasons for
several short paragraphs and emits:

\begin{quote}\footnotesize\ttfamily
\{"plan": 0.002, "decompose": 0.002, "implement": 0.005, "refine": 0.008951\}
\end{quote}

\noindent The CoT prompt rebalances slightly toward refine
($\$8.95 \!\times\! 10^{-3}$ vs.\ LLM-Direct's
$\$7.95 \!\times\! 10^{-3}$) at the cost of implement. 

\paragraph{ZEBRA-LLM ablation raw response.}
Given the curves from the controller, the LLM allocator returns:

\begin{quote}\footnotesize\ttfamily
\{"plan": 0.003, "decompose": 0.002, "implement": 0.006, "refine": 0.006951\}
\end{quote}

\noindent The LLM has been told the curves (and that refine is the
expensive phase) yet still under-funds refine relative to the
algorithmic solver: refine alloc $\$6.95 \!\times\! 10^{-3}$ vs.\
ZEBRA-mult\_offset's $\$12.92 \!\times\! 10^{-3}$. 

\paragraph{ZEBRA-additive raw output.}
The additive solver maximises $\sum_i u_i(x_i)$ subject to
$\sum_i x_i = B$. With refine's $b$ scaled down by the
\texttt{gpt-4o} cost ratio, refine has the largest marginal utility at
the binding budget and absorbs the bulk of the allocation:

\begin{quote}\footnotesize\ttfamily
\{"plan": 0.001150, "decompose": 0.001420, "implement": 0.002537, "refine": 0.012844\}
\end{quote}

\paragraph{ZEBRA-mult\_offset raw output.}
The multiplicative-with-offset solver maximises
$\prod_i (u_i(x_i) + \epsilon)$ subject to $\sum_i x_i = B$ (see
Section~\ref{sec:mult_offset}); the optimum keeps every phase
non-trivially funded and pushes the slowest-saturating phase (refine)
to its maximum:

\begin{quote}\footnotesize\ttfamily
\{"plan": 0.001134, "decompose": 0.001399, "implement": 0.002499, "refine": 0.012919\}
\end{quote}

\subsection{What happens during execution}
\label{app:example:execution}

For all strategies, each phase budget is handed to a
\texttt{BudgetEnforcedLLM} wrapper that runs up to $10$ iterations and
stops when (i) the phase output stops changing or (ii) the phase
budget is exhausted. Refine specifically implements a review--revise
pair (two \texttt{gpt-4o-2024-08-06} calls per round) and a candidate
selection step that compares the implement output against any
revisions on the $182$-test internal grader; the highest-scoring
candidate is shipped.

\emph{For \texttt{apps\_12} at $\alpha = 0.5$, refine \emph{starts}
on ${\ge}14/15$ seeds for every strategy -- but only the two ZEBRA
solvers reliably let the revise call \emph{finish}.} Across the four
non-ZEBRA strategies (LLM-Direct, LLM-CoT, ZEBRA-LLM ablation,
Uniform), the revise call hits the wrapper's per-call output-token
budget mid-write and either logs
\texttt{[revise] Output missing END\_OF\_OUTPUT token \ldots Truncated
output to parseable prefix} or earlier
\texttt{Budget below preferred length} / \texttt{Budget too low for
any output tokens} warnings. The truncated parseable prefix is a
syntactically partial function that scores worse than the implement
output, so refine reverts. The two ZEBRA solvers, with $\sim$$\$13
\!\times\! 10^{-3}$ on refine, log neither marker on seed~1: the
revise call completes within budget, the revision is well-formed,
and refine selects it.

\medskip

\noindent The mechanism behind the gap is a three-step chain visible
seed-by-seed:
\begin{enumerate}\itemsep=2pt
\item Implement frequently produces a near-correct but
edge-case-buggy solution (the always-add-$+1$ shortcut, the missing
\texttt{`S' not in trophies} branch, etc.). All five strategies'
implement phases share a similar bug rate -- mean implement-only
score is $0.27$--$0.31$ across LLM, LLM-CoT, ZEBRA-LLM, and
\texttt{mult\_offset}; \texttt{additive} is slightly higher at
$0.41$ thanks to a longer implement window, but still below the
pass threshold.

\item Refine \emph{can} fix these bugs -- review correctly
identifies the edge case on every refine-fire seed we audited --
but only if the revise call has enough budget to actually
\emph{finish} writing the corrected function. With
$\$7$--$9 \!\times\! 10^{-3}$ on refine, gpt-4o exhausts its
per-call output-token budget mid-revision and the wrapper logs
\texttt{[revise] Output missing END\_OF\_OUTPUT token \ldots
Truncated output to parseable prefix} on the canonical seed-1 logs
of LLM-Direct, LLM-CoT, and the ZEBRA-LLM ablation. The truncated
prefixes are syntactically incomplete (no return path, no closing
\texttt{max} reduction) and score $7$--$19/182$, well below the
implement output. With $\$13 \!\times\! 10^{-3}$, ZEBRA's revise
calls complete cleanly (no truncation marker, no
\texttt{Budget below preferred length} warning), produce longer
revisions ($\sim$$35$--$45$ lines vs.\ $\sim$$22$--$32$ lines for
the truncated cases), introduce explicit invariants (the
\texttt{total\_g} check on this task), and dominate the implement
output on the $182$-test grader.

\item ZEBRA's allocators sit in the second regime on every seed
(\texttt{additive} at $\$13.03 \pm 0.24$, \texttt{mult\_offset} at
$\$13.14 \pm 0.24$, both essentially deterministic across seeds).
LLM-based allocators sit in the first regime
(LLM-Direct $\$9.08 \pm 0.62$; LLM-CoT $\$6.68 \pm 1.91$;
ZEBRA-LLM ablation $\$7.28 \pm 1.89$). The ZEBRA-LLM ablation in
particular shows that the win is from the \emph{algorithmic} solver,
not the curves -- the same curves under an LLM allocator collapse
into a refine-starved split that puts the revision in the
regression regime.
\end{enumerate}

\end{document}